\pdfoutput=1

\documentclass[11pt]{article}

\usepackage[final]{acl}

\usepackage{times}
\usepackage{latexsym}
\usepackage{lipsum}  

\usepackage[T1]{fontenc}

\usepackage[utf8]{inputenc}

\usepackage{microtype}

\usepackage{inconsolata}

\usepackage{url}            
\usepackage{hyperref}       
\usepackage{booktabs}       
\usepackage{amsfonts}       
\usepackage{nicefrac}       
\usepackage{microtype}      
\usepackage{xcolor}         
\usepackage{enumitem}
\usepackage{graphicx}

\usepackage{amsmath}
\usepackage{amssymb}
\usepackage{mathtools}
\usepackage{amsthm}
\usepackage{soul}
\usepackage{enumitem}
\usepackage{subcaption}
\usepackage{pifont}

\usepackage{titletoc}

\title{Calibration Across Layers: Understanding Calibration Evolution in LLMs}

\author{
  Abhinav Joshi \qquad Areeb Ahmad \qquad Ashutosh Modi\\
  Department of Computer Science and Engineering\\
  Indian Institute of Technology Kanpur (IIT Kanpur)\\
  \texttt{\{ajoshi,areeb,ashutoshm\}@cse.iitk.ac.in} \\
}

\begin{document}
\maketitle
\begin{abstract}
Large Language Models (LLMs) have demonstrated inherent calibration capabilities, where predicted probabilities align well with correctness, despite prior findings that deep neural networks are often overconfident. Recent studies have linked this behavior to specific components in the final layer, such as entropy neurons and the unembedding matrix’s null space. In this work, we provide a complementary perspective by investigating how calibration evolves throughout the network's depth. Analyzing multiple open-weight models on the MMLU benchmark, we uncover a distinct \textit{confidence correction phase} in the upper/later layers, where model confidence is actively recalibrated after decision certainty has been reached. Furthermore, we identify a low-dimensional \textit{calibration direction} in the residual stream whose perturbation significantly improves calibration metrics (ECE and MCE) without harming accuracy. Our findings suggest that calibration is a distributed phenomenon, shaped throughout the network’s forward pass, not just in its final projection, providing new insights into how confidence-regulating mechanisms operate within LLMs. 
\end{abstract}


\section{Introduction} \label{sec:intro}

\begin{figure}[t]
\centering
 \includegraphics[width=0.77\linewidth]{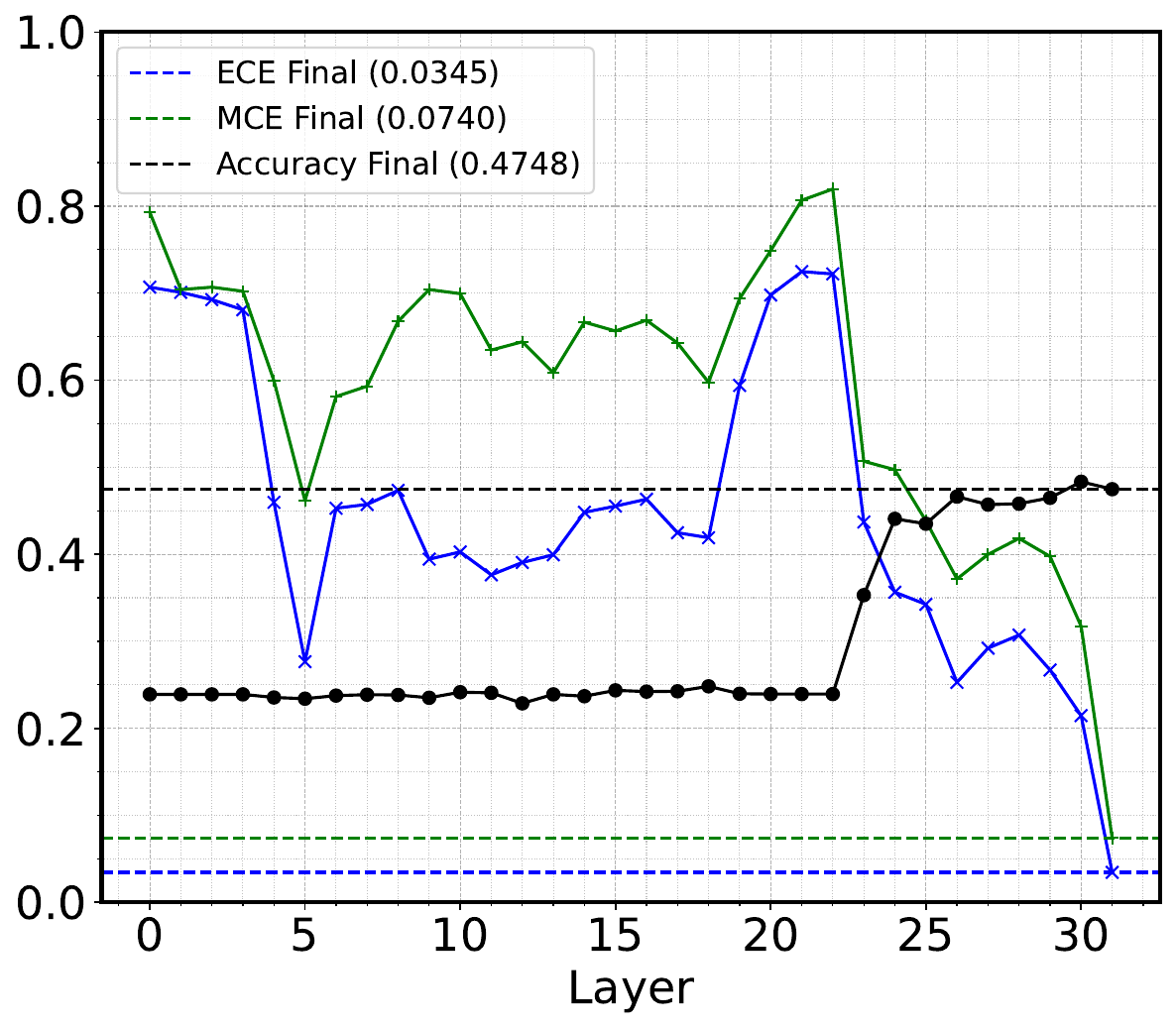}
 \caption{{
 The figure shows performance (Accuracy) along with model calibration scores (ECE and MCE) of the phi-2 model on the MMLU Humanities dataset. We observe that the model performance remains near-random (25\%, 4-options) for initial layers and starts to rise from layer 22 and saturates at layer 26, with minor changes in the 26-31 layers. However, the ECE and MCE scores first rise (layers 25-28) and then decline (layers 28-31), highlighting the model calibration changing in the later layers. }}
  \label{fig:mmlu_humanities_calibration_metrics_hook_resid_post_phi-2}
\end{figure}


\begin{figure*}[t]
\centering
 \includegraphics[width=0.92\linewidth]{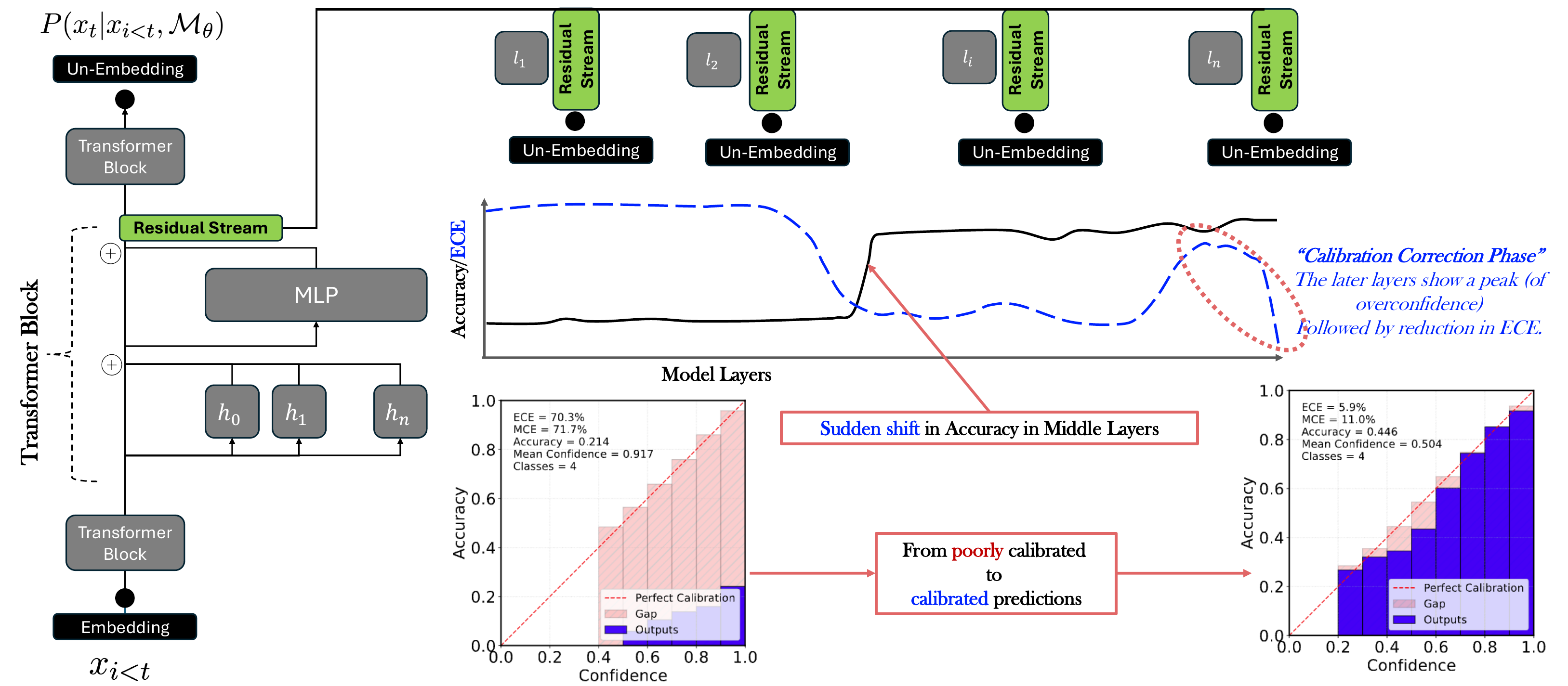}
  \caption{{The figure provides an overview of the performed study. The Residual stream signals from each of the layers are projected back to the vocabulary space using the unembedding matrix. The obtained predictions are inspected for accuracy and model calibration scores (ECE/MCE). The models show a sudden peak arising in the middle layers, after which the performance remains saturated. Interestingly, the model goes into a calibration correction phase where the ECE first rises and then reduces, while maintaining the same accuracy, i.e. going from a poorly calibrated predictions to calibrated predictions (shown as reliability diagrams, also see Figure \ref{fig:all_layers_calibration_relialibility_diagrams}).}}
  \label{fig:calibration_image}
\end{figure*}

Large Language Models (LLMs) have demonstrated strong generalization across a wide variety of tasks \citep{devlin-etal-2019-bert, Radford2019LanguageMA, incontextfewshotlearners}, yet it is challenging to understand how they manage and express uncertainty. Understanding the internal mechanisms by which LLMs regulate confidence is becoming increasingly important, especially as these models are deployed in settings where overconfidence can be costly. \textit{Model calibration}, the alignment between a model’s confidence and its accuracy, has emerged as a key axis for interpreting model behavior. Although deep neural networks have historically been found to be miscalibrated or overconfident \citep{on-calibration-guo-2017}, recent empirical studies suggest that LLMs exhibit surprisingly well-calibrated behavior across multiple tasks \citep{kadavath2022languagemodelsmostlyknow, achiam2023gpt, calib-prob-mcqa}. This has sparked growing interest in uncovering architectural or representational mechanisms that support/cause calibration in LLMs.

\noindent Recent investigations have made notable progress in this direction. For instance, entropy neurons in the final layers have been shown to adjust the uncertainty of model predictions while minimally affecting the output distribution \citep{stolfo2024confidence-regulation-neurons, gurnee2024universal}. These neurons modulate the entropy of the output distribution by operating in the null space of the unembedding matrix, effectively influencing confidence without altering accuracy. Such findings provide compelling evidence that calibration is an active, structured process, and that LLMs contain specialized components for regulating confidence.

\noindent In this work, we build on this line of research by exploring whether such mechanisms are also present \textit{within the intermediate layers} of the model. Specifically, we investigate how calibration evolves across the full depth of transformer-based language models. While prior work has illuminated the role of final-layer structures, the calibration dynamics of earlier layers remain less well understood.
We examine multiple popular open-weight models (Phi-2 \citep{javaheripi2023phi}, LLaMA-3 \citep{grattafiori2024llama3herdmodels}, LlaMa-2 \cite{touvron2023llama2openfoundation}, Mistral-7B \cite{jiang2023mistral7b}), on multiple real-world benchmarks (see App. \ref{app-sec:dataset-details}) with a special focus on the MMLU benchmark \citep{hendrycks2021measuringmassivemultitasklanguage}, inspecting the confidence and prediction behavior across layers via the residual stream \citep{elhage2021mathematical}. Our analysis reveals that calibration is not restricted to the model’s final stages. Instead, we find that LLMs undergo a clear \textit{confidence correction phase} in their later layers, where confidence is actively adjusted, even after prediction accuracy has stabilized. Further, we identify a low-dimensional direction in the residual stream that is consistently aligned with changes in model confidence. Perturbing this direction improves calibration metrics (ECE and MCE) without degrading accuracy, weakly suggesting the existence of a meaningful calibration subspace distributed across layers. In a nutshell, we make the following contributions: 
\begin{itemize}[nosep,noitemsep,leftmargin=*]
    \item We provide a layerwise analysis of calibration dynamics in transformer-based LLMs, showing that confidence is not simply correlated with accuracy but evolves through a distinct \textit{confidence correction phase}, where models become temporarily overconfident before self-adjusting in later layers. (see Figure \ref{fig:calibration_image} for an overview and Figure \ref{fig:mmlu_humanities_calibration_metrics_hook_resid_post_phi-2} for calibration changing across later layers.)
    \item We identify a “calibration direction” in the residual stream that governs confidence modulation, and demonstrate that small perturbations along this direction improve calibration metrics (ECE and MCE) without sacrificing accuracy. (see Figure \ref{fig:all_layers_calibration_intervention_main} and App. Figure \ref{fig:all_layers_calibration_phi-2_mmlu_all_truthfulqa_intervention} for generalization across datasets) 
    \item We provide a complementary perspective to existing work on final-layer calibration mechanisms by revealing distributed calibration behavior across the network’s depth, especially in intermediate layers that have received limited attention in prior studies.
\end{itemize}

\noindent In summary, our findings contribute to a more complete picture of how calibration is implemented within LLMs. Rather than being an isolated property of the final output layer, calibration appears to be a dynamic and distributed process. We hope this perspective encourages further work toward interpretable and controllable confidence modulation in language models. Our code is publicly accessible at \url{https://github.com/Exploration-Lab/LLM-Calibration-Mechanism}.
\section{Related Work} \label{sec:related_work}

Understanding and regulating uncertainty in neural networks has long been a foundational challenge in machine learning. Early studies revealed that modern deep networks tend to be overconfident and poorly calibrated \citep{on-calibration-guo-2017}, prompting the development of theoretical frameworks for uncertainty estimation, including Bayesian approximations and dropout-based techniques \citep{Gal2016UncertaintyID}. While much of this work focused on vision models or shallow classifiers, recent attention has shifted toward calibration in LLMs, where the stakes of miscalibrated predictions can be higher. Several recent studies have revealed that LLMs, despite their size and complexity, often exhibit surprisingly strong calibration properties. \citet{kadavath2022languagemodelsmostlyknow} and \citet{achiam2023gpt} showed that LLM token-level probabilities correlate well with accuracy, suggesting an emergent form of self-knowledge. This idea has been extended by \citet{yin-etal-2023-large} and \citet{xiong2024can}, who further investigated how well LLMs can express and act on their uncertainty in downstream tasks. However, \citet{kapoor2024large} cautioned that such behavior may not generalize without explicit training signals, raising questions about when and how such calibration arises. A complementary line of work has sought to uncover the mechanisms underlying these behaviors. Notably, \citet{stolfo2024confidence-regulation-neurons} identified final-layer ``confidence regulation neurons'' that influence the entropy of the model’s output distribution without significantly changing its predictions. Similarly, \citet{cancedda2024spectralfiltersdarksignals} explored how the spectral properties of the unembedding matrix contribute to calibration, emphasizing the importance of low-energy directions previously overlooked. Other studies like \citet{sharma2024the} demonstrated that a large fraction of the unembedding space is redundant and can be compressed without performance loss, which may affect how uncertainty is encoded. These findings suggest that LLMs use sophisticated mechanisms at their output layers to regulate confidence. However, most prior analysis have centered exclusively on final-layer phenomena, entropy neurons, attention patterns, and projection bottlenecks, while neglecting how confidence emerges and evolves across the depth of the network. Tools like the Logit Lens \citep{logit-lens} and residual stream analyses \citep{elhage2021mathematical} have made it possible to study intermediate representations, but their use in the context of calibration remains limited.


\noindent In parallel, some recent work has explored prompting strategies and fine-tuning methods for improving confidence estimation in LLMs \citep{tian-etal-2023-just}, while broader surveys \citep{geng-etal-2024-survey-calibration, survey-uncertainty-in-dnn-2023} have documented a wide variety of calibration techniques, from temperature scaling to Bayesian ensembling. Yet, these methods often treat the model as a black box, providing little insight into the internal computations shaping confidence.

\noindent In contrast, our work provides a mechanistic, layer-wise perspective on calibration, complementing prior studies by tracking how uncertainty evolves throughout the forward pass. We identify a \textit{confidence correction phase} in the later layers and a \textit{calibration direction} in the residual stream, demonstrating that confidence is explicitly modulated across depth, not just at the output. 
\section{Background} \label{sec:background}

\noindent In this section, we review essential background on transformer-based language modeling and model calibration. We focus on aspects most relevant to our study, which include token-level prediction in decoder-only transformers and how calibration metrics quantify model uncertainty.

\noindent\textbf{Transformer-based Language Modeling:} Transformer-based language models (LMs) are typically trained to predict the next token in a sequence, modeling the conditional probability distribution \( P(x_t \mid x_1, \ldots, x_{t-1}) \) over a vocabulary \( \mathcal{V} \) (Modern language models commonly use vocabularies of size \( |\mathcal{V}| \geq 50{,}000 \) \citep{Radford2019LanguageMA, liu2019robertarobustlyoptimizedbert}.) These models are implemented as deep neural networks parameterized by \( \theta \), denoted \( \mathcal{M}_\theta \), and trained in an autoregressive fashion. Given a token sequence \( x = [x_1, \ldots, x_{t-1}] \in \mathcal{V}^{t-1} \), the model outputs a vector of logits \( \mathbf{z}_t \in \mathbb{R}^{|\mathcal{V}|} \), where each component corresponds to the unnormalized log-probability of a vocabulary token. Applying the softmax function to \( \mathbf{z}_t \) yields the probability distribution over the next token. Internally, decoder-only models consist of a stack of transformer blocks \( f_{\theta_1}, f_{\theta_2}, \ldots, f_{\theta_L} \), which process the input sequence via self-attention and feedforward layers. These blocks operate on and update a shared residual stream, with skip connections facilitating gradient flow and information propagation \citep{elhage2021mathematical}. At each layer, token representations are refined until the final hidden state is projected to the vocabulary space. We focus on the representation corresponding to the last input token \( x_{t-1} \), as this token is typically responsible for generating the next-token prediction. The final residual vector for this token is first normalized by a LayerNorm module, then projected to the vocabulary space via a learned weight matrix \( \mathbf{W}_U \in \mathbb{R}^{|\mathcal{V}| \times d_{\text{model}}} \), often referred to as the unembedding matrix \citep{elhage2021mathematical}. 

\noindent\textbf{Layer Normalization (LayerNorm)} \citep{ba2016layernormalization} plays a critical role in stabilizing training and enhancing convergence in transformer models. Given an input vector \( \mathbf{z}_t \in \mathbb{R}^{d_{\text{model}}} \), LayerNorm transforms it as: 
\[
\mathtt{LayerNorm}(\mathbf{z}_t) = \frac{\mathbf{z}_t - \mu_{\mathbf{z}_t}}{\sqrt{\mathtt{Var}(\mathbf{z}_t) + \epsilon}} \odot \boldsymbol{\gamma} + \boldsymbol{\beta}
\] 
Here, \( \mu_{\mathbf{z}_t} \) and \( \mathtt{Var}(\mathbf{z}_t) \) denote the mean and variance of the vector components, and \( \boldsymbol{\gamma}, \boldsymbol{\beta} \in \mathbb{R}^{d_{\text{model}}} \) are learned scale and shift parameters. This operation standardizes the input and enables better gradient flow across layers.

\noindent\textbf{Model Calibration:}
In machine learning models \emph{calibration} refers to the alignment between a model's predicted confidence and the actual likelihood of being correct. A model is said to be well-calibrated, if across many predictions, tokens predicted with a given probability \( p \) are correct approximately \( p \) fraction of the time.
Formally, for a given input prompt \( x = [x_1, \ldots, x_{t-1}] \), let \( y = x_t \) denote the true next token, and let \( \hat{y} = \arg\max_{v \in \mathcal{V}} P_{\mathcal{M}_\theta}(v \mid x) \) be the model's predicted token. The model's \emph{confidence} is given by \( p = P_{\mathcal{M}_\theta}(\hat{y} \mid x) \), while its \emph{accuracy} is defined as \( a = \mathbb{I}(\hat{y} = y) \), where \( \mathbb{I} \) is the indicator function. To assess calibration over a dataset \( \mathcal{D} = \{(x^{(i)}, y^{(i)})\}_{i=1}^N \), predictions are grouped into \( M \) bins \( \{B_m\}_{m=1}^M \) based on their confidence scores (e.g., into intervals such as [0.0, 0.1), [0.1, 0.2), etc.). For each bin \( B_m \), we define:
\begin{align*}
\text{conf}(B_m) &= \frac{1}{|B_m|} \sum_{i \in B_m} p^{(i)}, \\
\text{acc}(B_m)  &= \frac{1}{|B_m|} \sum_{i \in B_m} a^{(i)},
\end{align*}
\noindent where \( p^{(i)} \) and \( a^{(i)} \) are the confidence and accuracy for the \( i \)-th prediction. The \emph{Expected Calibration Error} (ECE) aggregates the absolute difference between confidence and accuracy over bins, weighted by the number of samples per bin:
\[
\text{ECE} = \sum_{m=1}^{M} \frac{|B_m|}{N} \left| \text{acc}(B_m) - \text{conf}(B_m) \right|.
\]

\noindent A related metric, the \emph{Maximum Calibration Error} (MCE), captures the worst-case bin-level deviation between accuracy and confidence:
\[
\text{MCE} = \max_{m \in \{1, \ldots, M\}} \left| \text{acc}(B_m) - \text{conf}(B_m) \right|.
\]

\noindent Both metrics are minimized (i.e., equal to zero) when the model is perfectly calibrated. For a comprehensive treatment of calibration techniques in language models, please refer to  \citet{pavlovic2025understandingmodelcalibration}.

\noindent\textbf{Reliability Diagrams:}
To visualize model calibration, we use \emph{reliability diagrams} \cite{on-calibration-guo-2017} (see App. Fig \ref{fig:all_layers_calibration_relialibility_diagrams}), which plot predicted confidence against empirical accuracy for different confidence intervals. In a perfectly calibrated model, the points lie on the diagonal \( y = x \), indicating that predicted probabilities align with observed correctness. Deviations below the diagonal suggest overconfidence, while points above the diagonal indicate underconfidence. Reliability diagrams provide an intuitive, qualitative assessment of how model confidence corresponds to actual performance. 
\section{Experimental Setup} \label{sec:experiments}

In this work, we study the calibration behavior of transformer-based language models (LLMs) by analyzing their performance on multiple-choice question answering (MCQA) tasks. The model predicts the next token conditioned on the context (input query). Our experimental setup focuses on assessing how the input structure, query framing, and model components impact calibration.

\noindent\textbf{Task Setup:}
We evaluate LLMs on a task where a query is presented in the form of a multiple-choice question answering (MCQA) prompt. The input consists of two primary components:  
1) \emph{Query Information} (\(x_{\text{query}}\)): This contains the specific question or context associated with a dataset instance.  
2) \emph{Choice Set} (\(x_{\text{options}}\)): This includes the set of answer options provided for each instance. The number of options depends on the dataset (e.g., four in the case of MMLU). 
Specifically, given the input structure as:
$$
P(x_t|x_{i<t}, \mathcal{M}_{\theta}) = P(x_t|x_{\text{query}}, x_{\text{options}}, x_{\epsilon}, \mathcal{M}_{\theta})
$$
where, $x_{\text{query}}$ represents the query information (specific question or context), $x_{\text{options}}$ denotes the set of available answer choices (e.g., {A, B, C, D}), $x_{\epsilon}$ represents the set of prompt templates used for MCQA,
$\mathcal{M}_\theta = \{ f_{\theta_1}, f_{\theta_2}, \dots, f_{\theta_L} \}$ represents the language model with parameters $\theta$ across $L$ layers.
The model is expected to generate the correct answer choice token as the next token in the output sequence, which we evaluate for performance.
Additionally, to ensure diversity and mitigate potential position biases in the answer choices, we randomize the order of the answer options, where:
$$
x_{\text{options}} \leftarrow \{ \text{A. } o_{\text{correct}}, \text{B. } o_{\text{wrong}} \}
$$
This formulation allows us to test the model's ability to handle different question structures while monitoring its confidence across various layers of the transformer. In this work, we stick to reasoning captured using multiple-choice question answering (MCQA)-style prompts \citep{robinson2023leveraging, joshi2024cold}. The MCQA setup provides a principled and constrained setting for investigating the internal decision-making processes of LLMs \cite{wiegreffe2025answer, joshi-etal-2025-towards}. Unlike open-ended or cloze-style generation, MCQA structures the task as a selection among discrete alternatives, thereby reducing confounding factors related to token frequency, length bias, and linguistic fluency \protect\citep{incontextfewshotlearners}. This format enables precise analysis of the transition from contextual representation to decision, making it well-suited for quantifying/measuring calibration changing across intermediate representations. 


\noindent The use of structured MCQA format helps in consistent evaluation of model calibration across layers, as it requires the model to make a discrete decision among a fixed set of alternatives. Unlike open-ended generation or cloze-style completion, MCQA also has less ambiguity in output interpretation by constraining the prediction space, allowing us to more directly isolate and measure model confidence. Moreover, the use of standardized evaluation using metrics such as Expected Calibration Error (ECE) and Maximum Calibration Error (MCE) can be established in a straightforward fashion, which is more difficult to apply in generative settings. Additionally, because the model's output is evaluated on a fixed set of tokens, MCQA avoids the stochasticity introduced by sampling strategies (e.g., temperature sampling or top-$k$ decoding), which often confound confidence analysis in generation tasks. In contrast, open-ended generation introduces several challenges for layer-wise calibration analysis, i.e., the ambiguity of token-level correctness, the absence of well-defined calibration metrics for full sequences, and the non-determinism inherent in decoding strategies. For these reasons, we specifically adopt MCQA as a controlled and interpretable framework for understanding internal confidence dynamics and calibration behavior in large language models.

\noindent\textbf{Datasets and Prompt Templates:}
For our experiments, we use multiple real-world datasets (see App. \ref{app-sec:dataset-details} for details) with a primary focus on the Massive Multitask Language Understanding (MMLU) benchmark \cite{hendrycks2021measuringmassivemultitasklanguage}, which spans 57 diverse subjects across STEM, humanities, social sciences, and other fields. This benchmark is designed to evaluate general knowledge acquired during pretraining of language models. Each question is paired with four answer choices, and model performance is evaluated based on the correctly predicted choice.
Notably, the nature of the answer choices varies across MMLU categories. In some categories (e.g., logical reasoning or high school computer science), the choices are relatively formulaic and repeat across multiple instances (e.g., “True”/“False”). In contrast, other categories (e.g., medical or legal domains) present unique, context-dependent options for each question. This variation introduces different levels of reasoning complexity and lexical diversity, making calibration analysis more nuanced. We include a representative prompt template in App. Figure \ref{app-fig:mmlu_question_prompt}.
This diversity introduces varying degrees of lexical and semantic complexity, which we believe provides an overall generalization of the experimental findings regarding calibration that we further explore in our experiments.

\noindent\textbf{Monitoring Layer Performance:}
To understand how calibration varies across layers, we take inspiration from the approach by Logit Lens \cite{logit-lens}, computing accuracy at different layers of the transformer. After each transformer block, we extract the residual stream representation \(z_t\) and project it onto the vocabulary space using the unembedding matrix \(\mathbf{W}_U\) as follows:
\[
\text{logits}(z_t) = \mathbf{W}_U \, \text{LayerNorm}(z_t)
\]
We then compute the accuracy of the logits at each layer and track the changes in performance as the information propagates through the layers. This helps us pinpoint which layer's representations are most decisive for the model’s predictions. Additionally, we calculate the Expected Calibration Error (ECE) and Maximum Calibration Error (MCE) at each layer to quantify the calibration at different stages of the model.
These measurements provide insight into both representational quality and the internal emergence of confidence across depth (model layer internals).
Also see App. \ref{app-sec:residual_stream_computations} for details on the residual stream computations.


\noindent\textbf{Analyzing the Prediction (Unembedding and Confidence Dynamics): }
To further analyze the role of the unembedding matrix, we apply Singular Value Decomposition (SVD) to it. The unembedding matrix \(\mathbf{W}_U\) is decomposed as:
\[
\mathbf{W}_U = \mathbf{U}_U \Sigma_U \mathbf{V}_U^T
\]
This decomposition helps us separate the projection of the residual stream onto the prediction space, allowing us to study the significance of different components of the matrix. We find that the singular values exhibit a consistent pattern, where the initial values are large, followed by a long tail with decreasing values. Recent research \cite{sharma2024the} suggests that this decomposition can be used to improve model performance by pruning less significant components, but the last few singular values, especially those in the tail, play a crucial role in the model's predictions and calibration \cite{cancedda2024spectralfiltersdarksignals}. (also see App. Figure \ref{fig:eigen_values_unemembedding_matrix} for singular values of Unembedding Matrix in Phi-2 and Llama-3-8B models showing null space)

\noindent\textbf{Notes from Prior Work:}
Recent studies have revealed intriguing structural properties of transformer models, particularly in their final layers. For instance, the unembedding matrix $\mathbf{W}_U$ often exhibits a characteristic spectral pattern when decomposed via Singular Value Decomposition (SVD), i.e., a handful of large singular values followed by a long tail and a sharp drop in the final 5\% of the spectrum. \citet{sharma2024the} shows that substantial portions of these component matrices can be pruned (via SVD) without hurting, sometimes even improving, model performance. In contrast, \citet{cancedda2024spectralfiltersdarksignals} argue that the final singular modes, often dismissed as unimportant, in fact carry signals critical to prediction accuracy. Complementing this, \citet{stolfo2024confidence-regulation-neurons} propose that the model deliberately shapes this low-rank null space to regulate prediction confidence, effectively influencing model calibration.

\noindent While these findings highlight how late-stage components influence model confidence and calibration, less is known about the evolution of calibration within the model, especially across intermediate layers. Our study addresses this gap by analyzing how uncertainty and confidence emerge and evolve throughout the model's depth. Specifically, we measure both predictive performance and calibration metrics (ECE, MCE) layer-wise to localize where in the transformer stack the model becomes ``confident'' and how reliably that confidence reflects correctness.
This layerwise perspective allows us to identify where in the model, confidence stabilizes and to what extent it is calibrated across depth.

\section{Results and Analysis} \label{sec:results}

\begin{figure*}[t]
\centering
\captionsetup[subfigure]{labelformat=parens}

\newcommand{\scalefactor}{0.24} 

\begin{subfigure}[b]{\scalefactor\linewidth}
    \includegraphics[width=\linewidth]{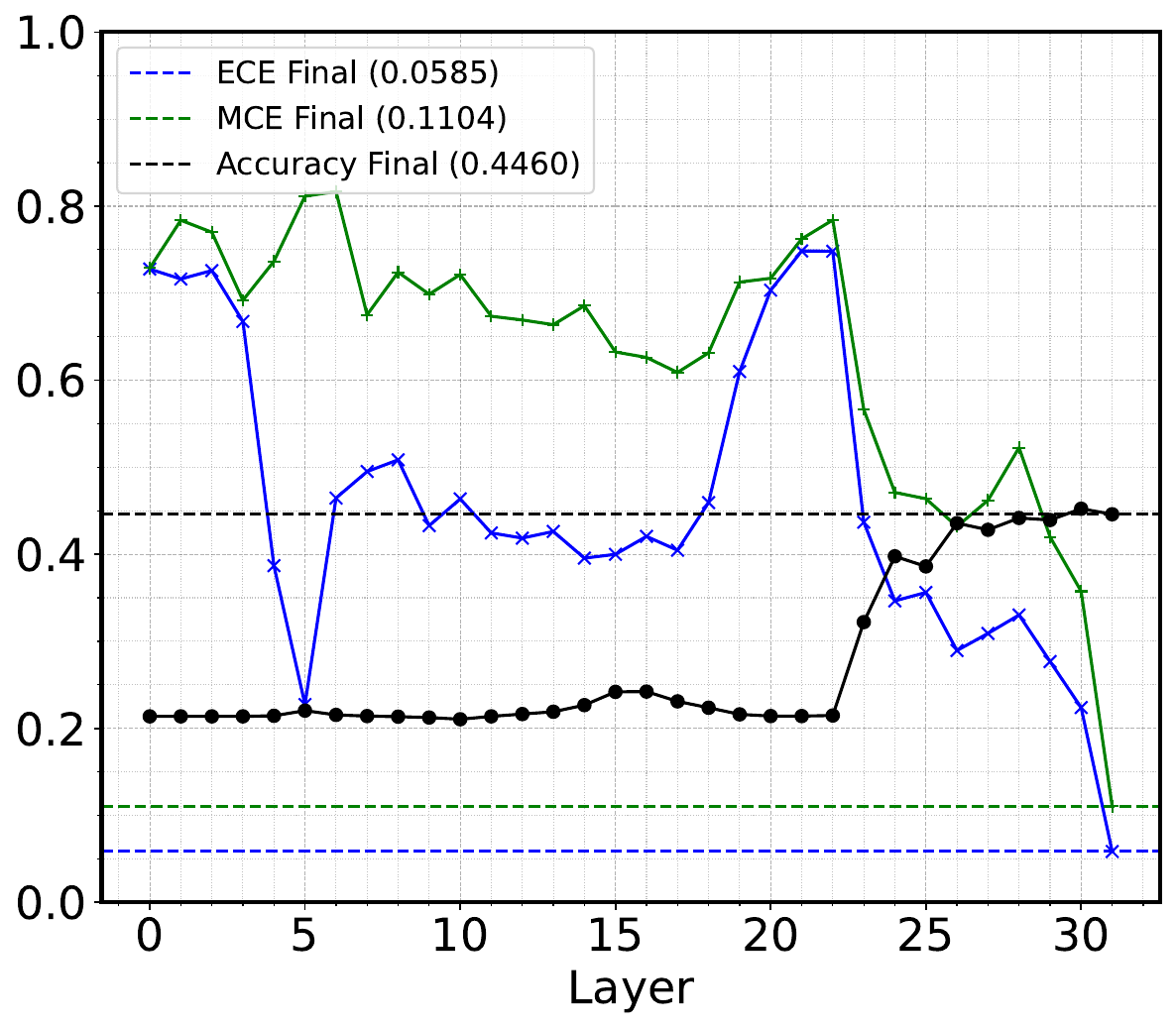}
    \caption{MMLU STEM}
\end{subfigure}
\hfill
\begin{subfigure}[b]{\scalefactor\linewidth}
    \includegraphics[width=\linewidth]{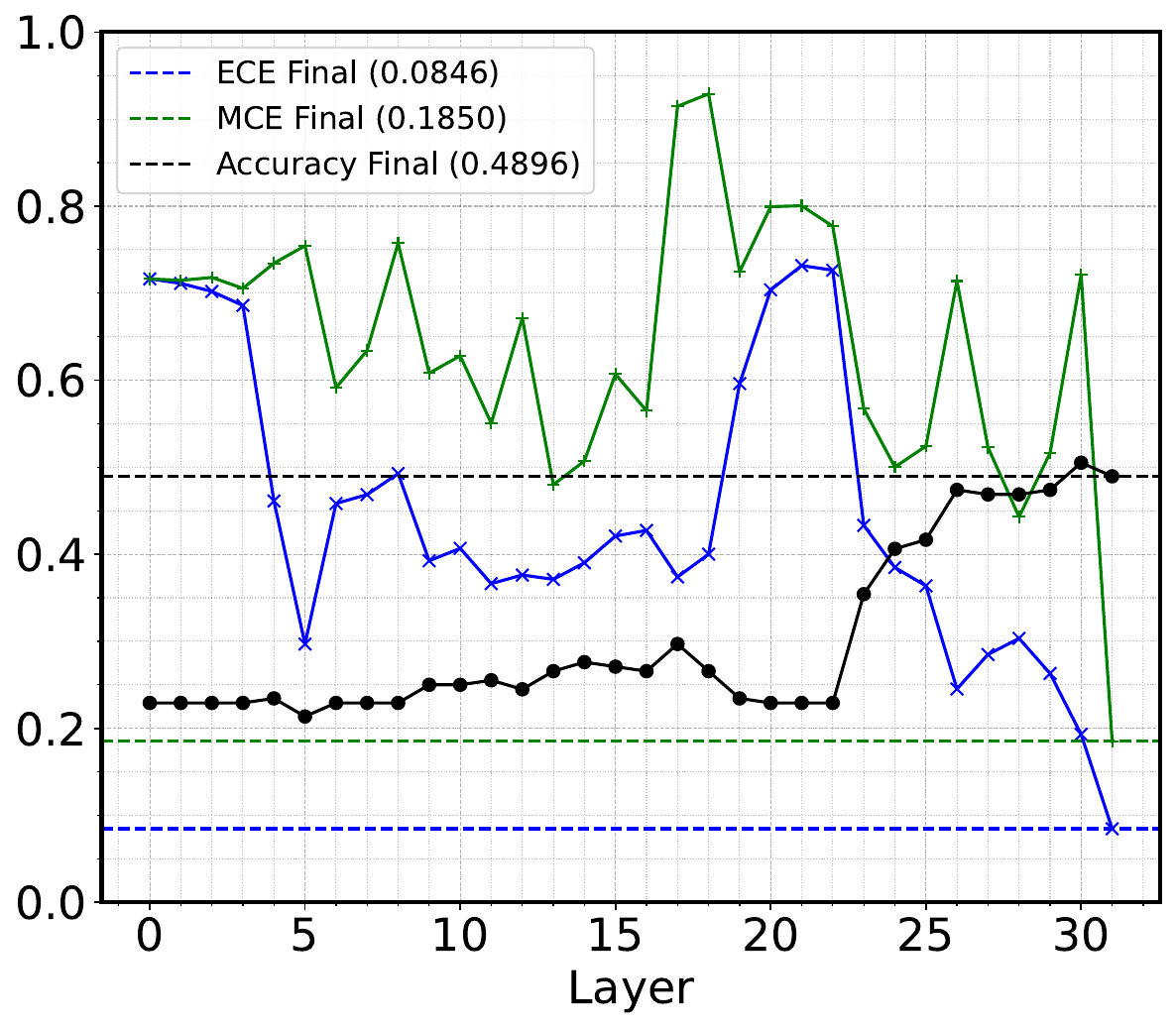}
    \caption{MMLU Humanities}
\end{subfigure}
\hfill
\begin{subfigure}[b]{\scalefactor\linewidth}
    \includegraphics[width=\linewidth]{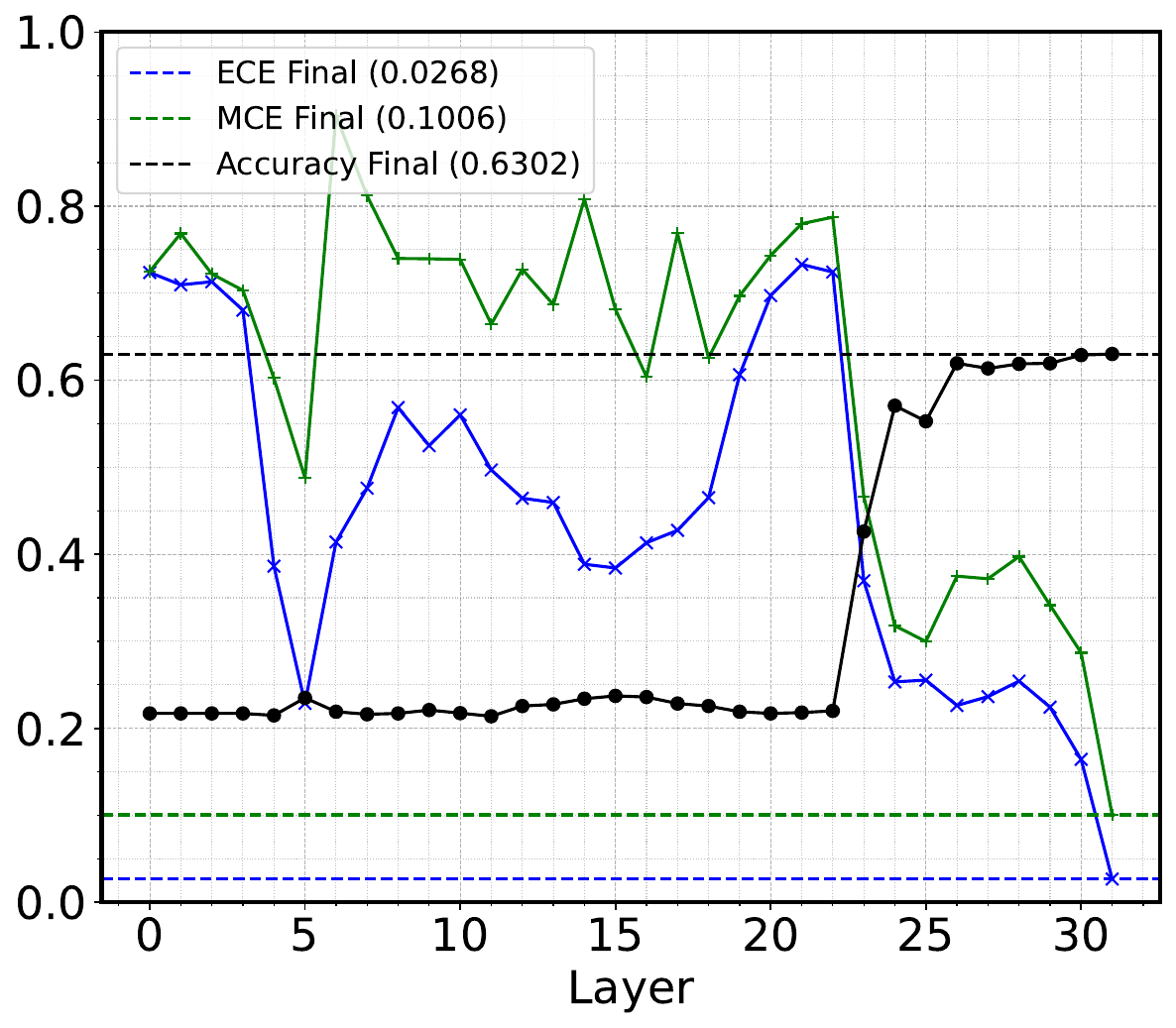}
    \caption{MMLU Social Science}
\end{subfigure}
\hfill
\begin{subfigure}[b]{\scalefactor\linewidth}
    \includegraphics[width=\linewidth]{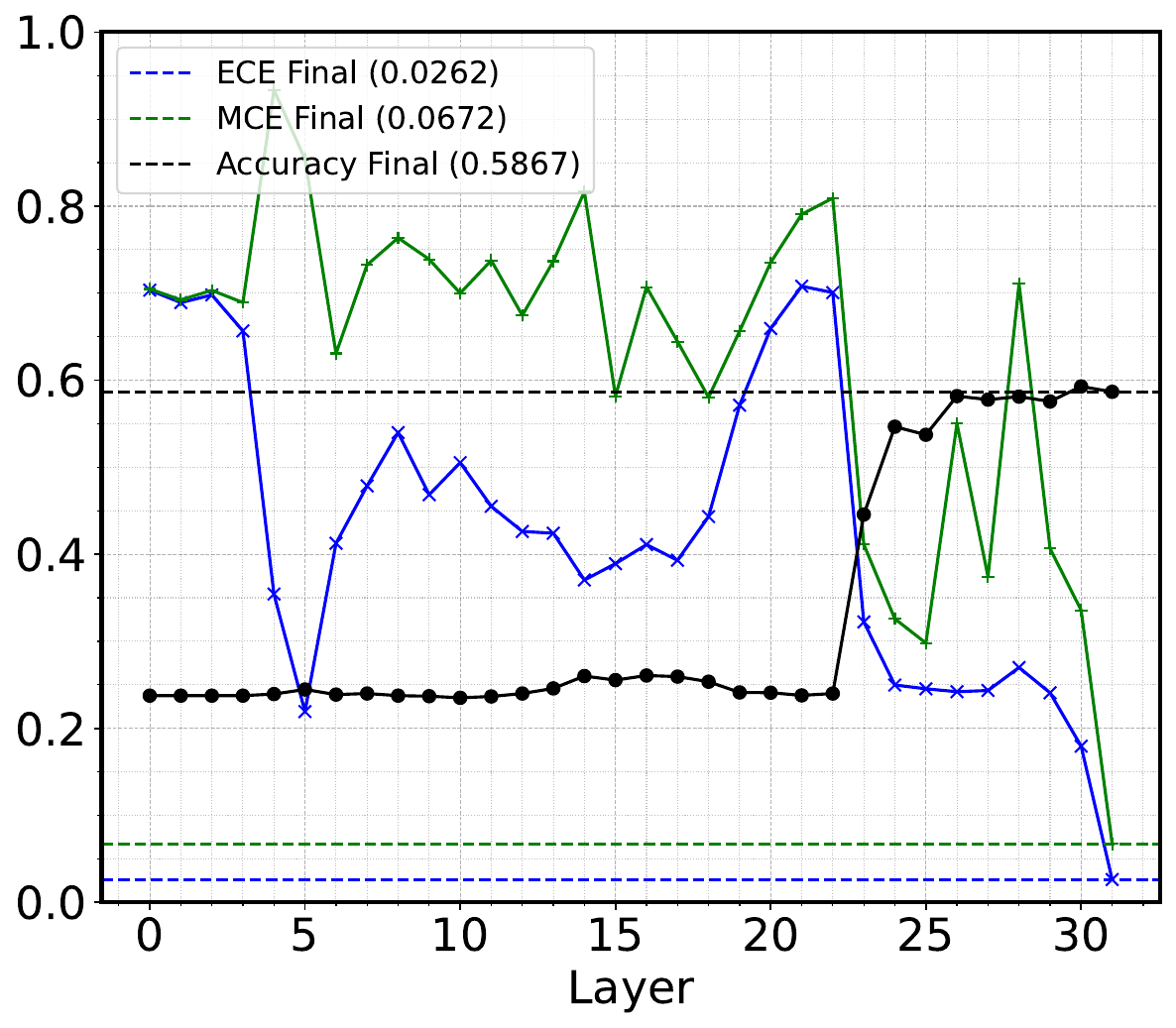}
    \caption{MMLU Others}
\end{subfigure}
\hfill
\caption{The figure shows performance (Accuracy) along with model calibration scores (ECE and MCE) of the Phi-2 model on the different datasets. We observe that the model performance starts to rise from layer 22 and saturates at layer 25/26, with minor changes in the 26-31 layers. However, the ECE and MCE scores first rise (layers 26-28) and then decline (layers 29-31), highlighting calibration changing in the later layers, with meager changes in the model performance. 
The upper/later layers show the presence of \textbf{\textit{calibration correction phase}}. 
Similar trends are found for other models (Llama-3-8B Figure \ref{fig:all_layers_calibration_Llama-3-8B_mmlu_all}, Mistral-7B Figure \ref{fig:all_layers_calibration_mistralai_Mistral-7B-v0.1}, and Llama-2-7B Figure \ref{fig:all_layers_calibration_meta-llama_Llama-2-7b-hf_mmlu_all}).
}
\label{fig:all_layers_calibration_phi-2_mmlu_all}
\end{figure*}

\begin{figure*}[t]
\centering
\captionsetup[subfigure]{labelformat=parens}

\newcommand{\scalefactor}{0.33} 

\begin{subfigure}[b]{\scalefactor\linewidth}
    \includegraphics[width=\linewidth]{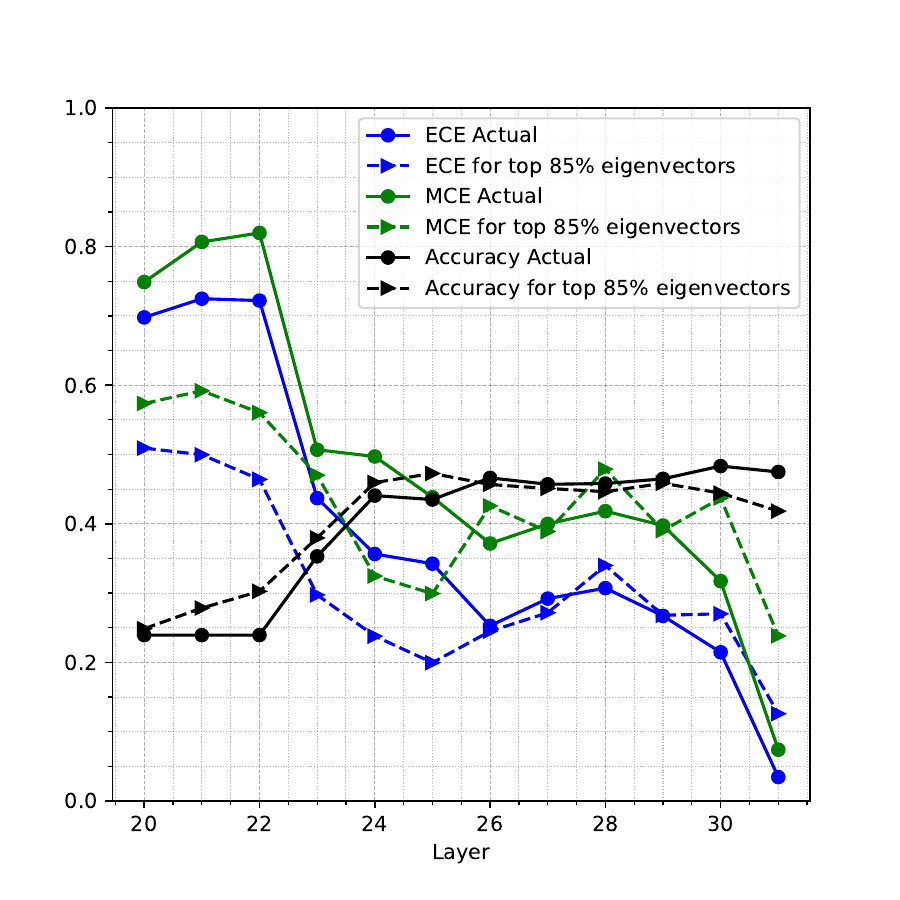}
    \caption{top $85\%$}
\end{subfigure}\hfill
\begin{subfigure}[b]{\scalefactor\linewidth}
    \includegraphics[width=\linewidth]{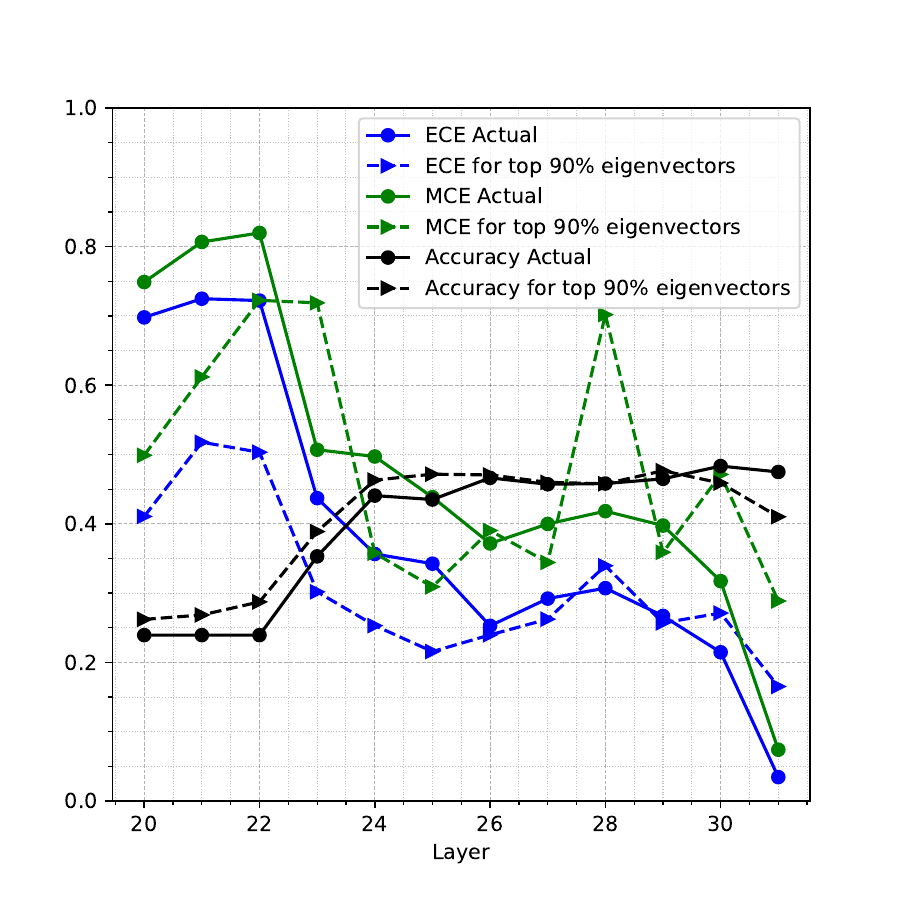}
    \caption{top $90\%$}
\end{subfigure}\hfill
\begin{subfigure}[b]{\scalefactor\linewidth}
    \includegraphics[width=\linewidth]{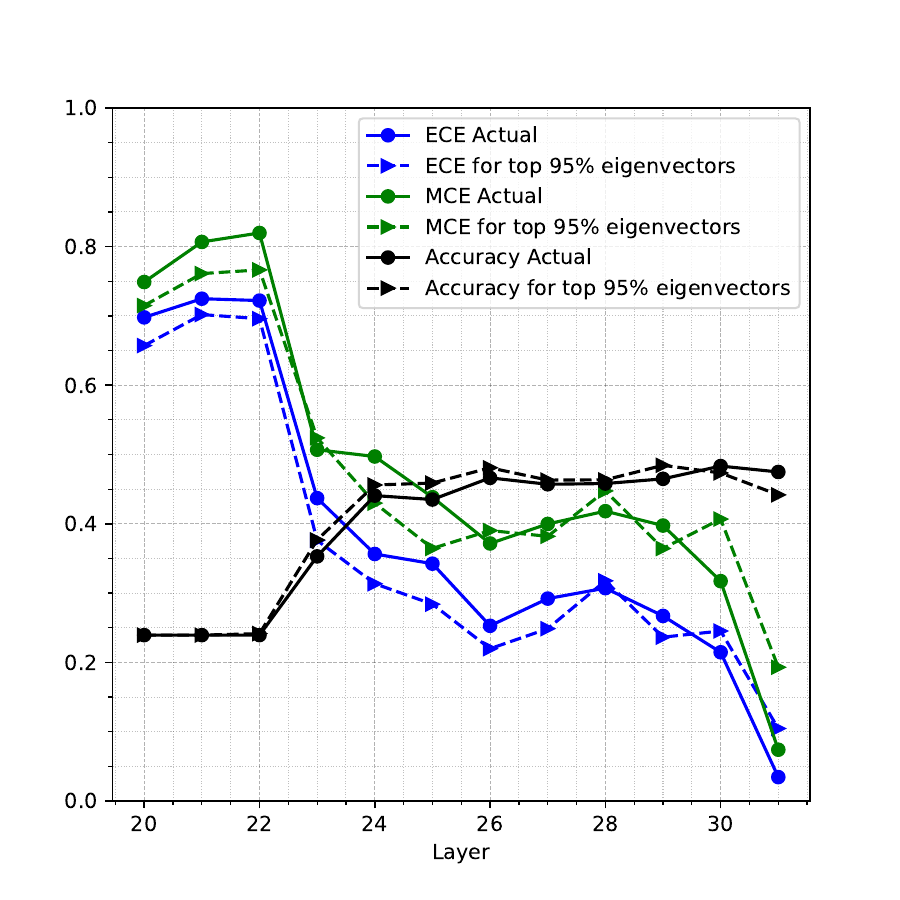}
    \caption{top $95\%$}
\end{subfigure}
\caption{The figure shows performance (Accuracy) along with model calibration scores (ECE and MCE) of the phi-2 model computed by reconstructing the unembedding matrix, using only top-85\%, top-90\% and top-95\% singular values in (a), (b), and (c), respectively. Overall, we observe the ECE scores with minor fluctuations, pointing towards a small contribution of lower singular values in model calibration. 
}
\label{fig:all_layers_reconstructed_unembedding}
\end{figure*}

We present our findings in three stages: \textbf{1) How calibration evolves} across layers of a transformer, \textbf{2) The role of the unembedding} matrix's low-rank components, and \textbf{3) The discovery of a direction} in activation space that appears to regulate model calibration.

\begin{figure*}[t]
\centering
\captionsetup[subfigure]{labelformat=parens}

\newcommand{\scalefactor}{0.24} 

\begin{subfigure}[b]{\scalefactor\linewidth}
    \includegraphics[width=\linewidth]{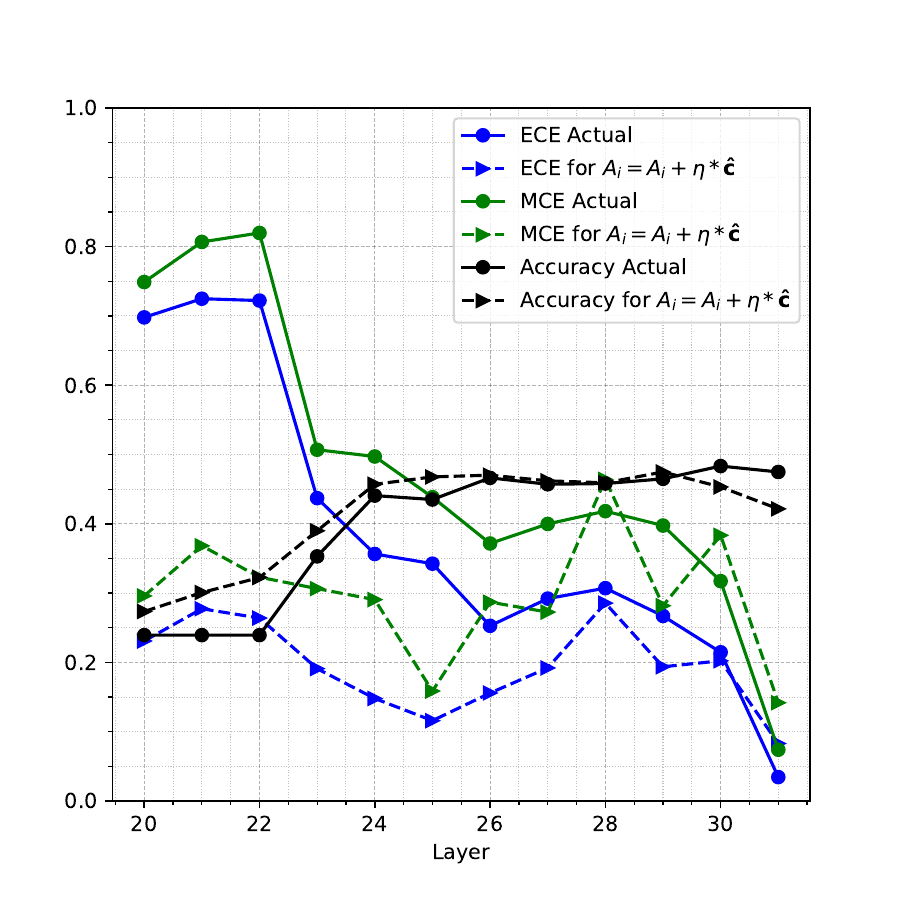}
    \caption{MMLU Humanities}
\end{subfigure}
\hfill
\begin{subfigure}[b]{\scalefactor\linewidth}
    \includegraphics[width=\linewidth]{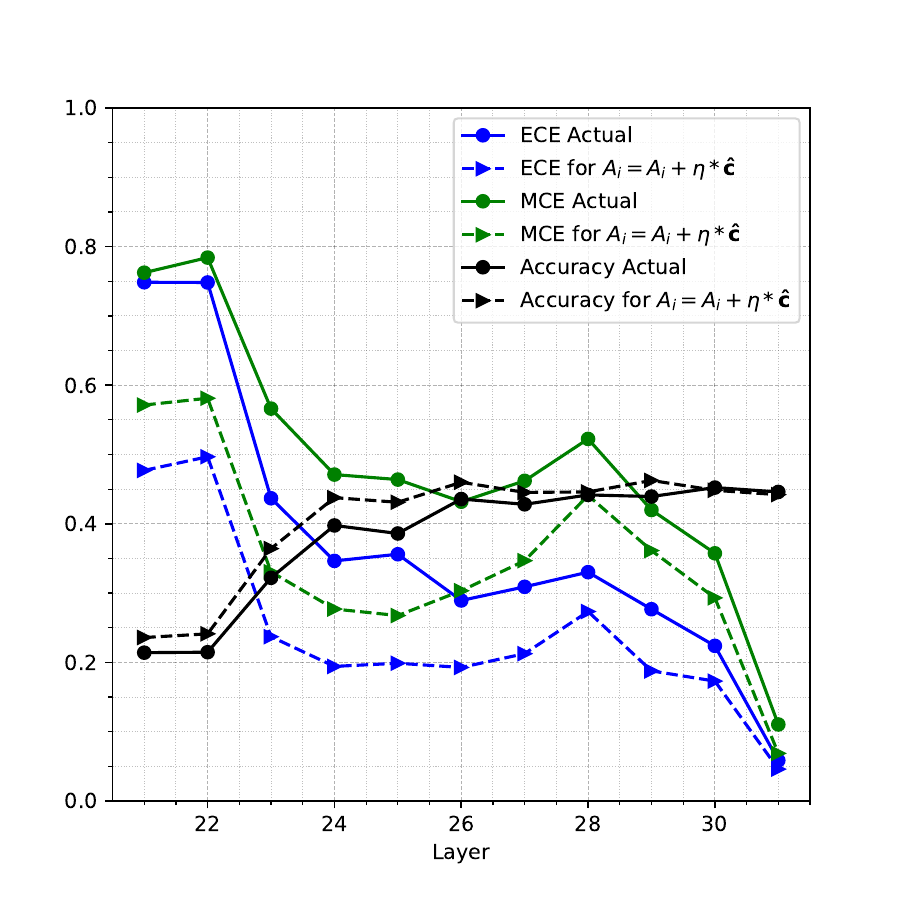}
    \caption{MMLU STEM}
\end{subfigure}\hfill
\begin{subfigure}[b]{\scalefactor\linewidth}
    \includegraphics[width=\linewidth]{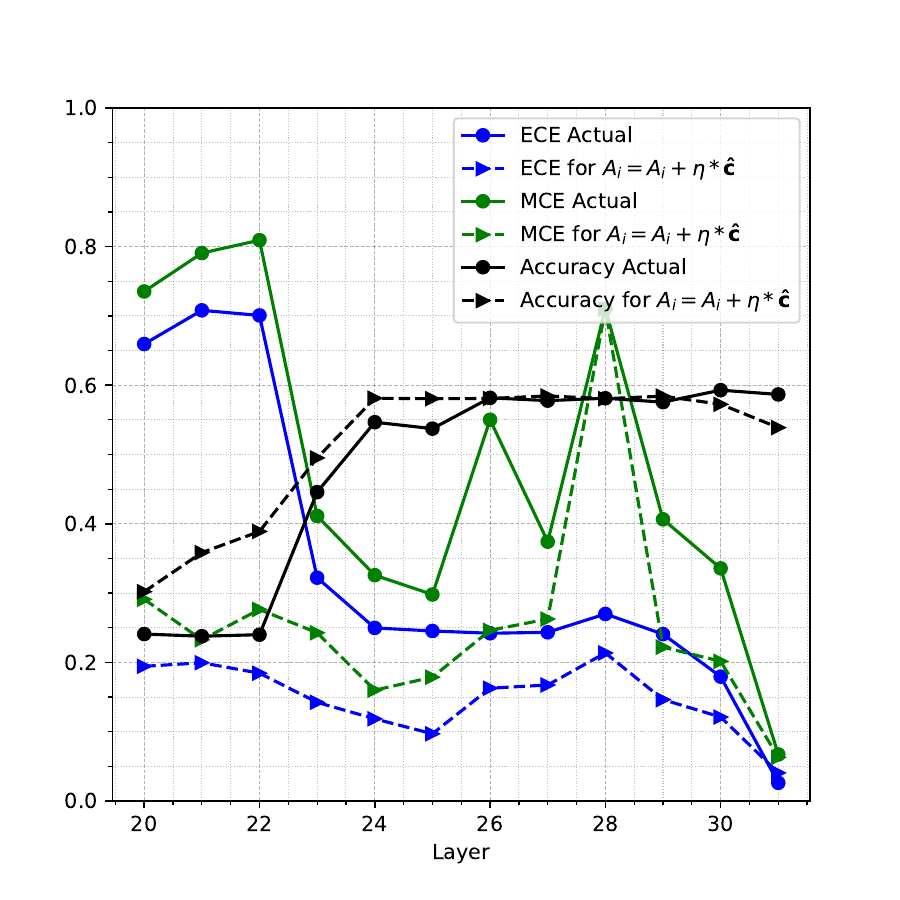}
    \caption{MMLU Others}
\end{subfigure}
\hfill
\begin{subfigure}[b]{\scalefactor\linewidth}
    \includegraphics[width=\linewidth]{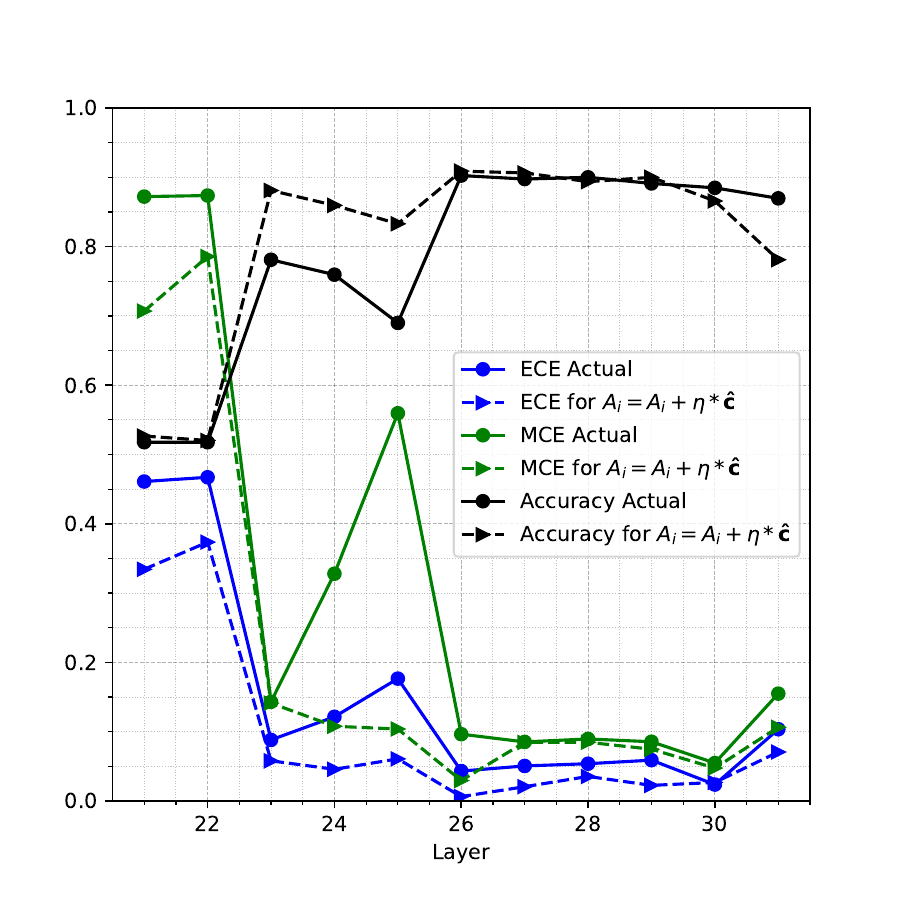}
    \caption{TruthfulQA}
\end{subfigure}
\caption{The figure shows performance (Accuracy) along with model calibration scores (ECE and MCE) of the phi-2 model on the different datasets when adding the calibration direction to the residual stream. The added calibration direction to the residual stream helps shift the calibration scores to lower values, validating the impact of the calibration direction. Interestingly, the direction found using MMLU Humanities works well for other datasets like MMLU Others. (Due to space constraints, we move the results on other datasets to the App. Figure \ref{fig:all_layers_calibration_phi-2_mmlu_all_truthfulqa_intervention})
}
\label{fig:all_layers_calibration_intervention_main}
\end{figure*}

\noindent\textbf{Layerwise Calibration Dynamics:}
We begin by analyzing how calibration and prediction performance vary across transformer layers in the phi-2 model. Each transformer block modifies the residual stream, which we project into the vocabulary space using the unembedding matrix $\mathbf{W}_U$ (see \S\ref{sec:experiments}). At each layer, we compute predictive performance (via Accuracy), Expected Calibration Error (ECE), and Maximum Calibration Error (MCE).

\noindent Across multiple datasets, a consistent trend emerges: accuracy begins to rise significantly from layer 22 and stabilizes by layer 26. However, the calibration behavior follows a different trajectory, ECE and MCE scores increase after layer 25, peaking around layer 28, before declining toward the final layers. This suggests that even after the model has become sufficiently accurate, it undergoes a phase of overconfidence before recalibrating its predictions. We refer to this as a “\textit{confidence correction phase}” in the final layers.

\noindent To visualize this phenomenon, Figure \ref{fig:all_layers_calibration_phi-2_mmlu_all} shows how reliability improves across the final layers. Reliability diagrams (App. Figure \ref{fig:all_layers_calibration_relialibility_diagrams}) further confirm this trend, revealing a widening and then narrowing gap between model confidence and accuracy. 
This denotes that the residual stream in the later layers is affected/modified in such a way that modulates the model calibration with no/minor change in the model performance (black line). 
The upper/later layers show the presence of \textbf{\textit{calibration correction phase}}. 
Similar trends are found for other models (Llama-3-8B Figure \ref{fig:all_layers_calibration_Llama-3-8B_mmlu_all}, Mistral-7B Figure \ref{fig:all_layers_calibration_mistralai_Mistral-7B-v0.1}, and Llama-2-7B Figure \ref{fig:all_layers_calibration_meta-llama_Llama-2-7b-hf_mmlu_all}).

\noindent\textbf{Effect of Unembedding Null Space:}
Prior work \cite{cancedda2024spectralfiltersdarksignals, stolfo2024confidence-regulation-neurons} suggests that the lower-rank (small singular value) components of the unembedding matrix may be involved in calibration, particularly via “entropy neurons” writing into its null space. To test this, we decompose the unembedding matrix $\mathbf{W}_U = \mathbf{U}_U \Sigma_U \mathbf{V}_U^T$ and reconstruct it by discarding the smallest 5\%, 10\%, and 15\% of singular values:

\begin{equation*}
\hat{\mathbf{W}}_{U} = \mathbf{U}_{U} \, \Sigma_{U}^{[:k]} \, \mathbf{V}^{T}_{{U}_{[:k]}}    
\end{equation*}

\noindent $\text{where, } k \in \{85\%, 90\%, 95\%\}$. 
This intervention limits the influence of the null space on model outputs. Figure~\ref{fig:all_layers_reconstructed_unembedding} shows the results using these truncated matrices. Accuracy remains largely unchanged, indicating that most predictive capacity lies in the dominant singular vectors. However, we observe fluctuations in calibration metrics, especially MCE, supporting the hypothesis that the null space plays a supporting role in calibration.

\noindent Interestingly, we find that these effects are in both directions (increasing and decreasing calibration) for middle layers, pointing towards no clear indication of ECE/MCE being increased when null space is removed. Null-space sensitivity appears across the upper layers of the network, suggesting that calibration is mediated by distributed subspaces throughout the model.

\begin{figure}[t]
\centering
 \includegraphics[width=0.95\linewidth]{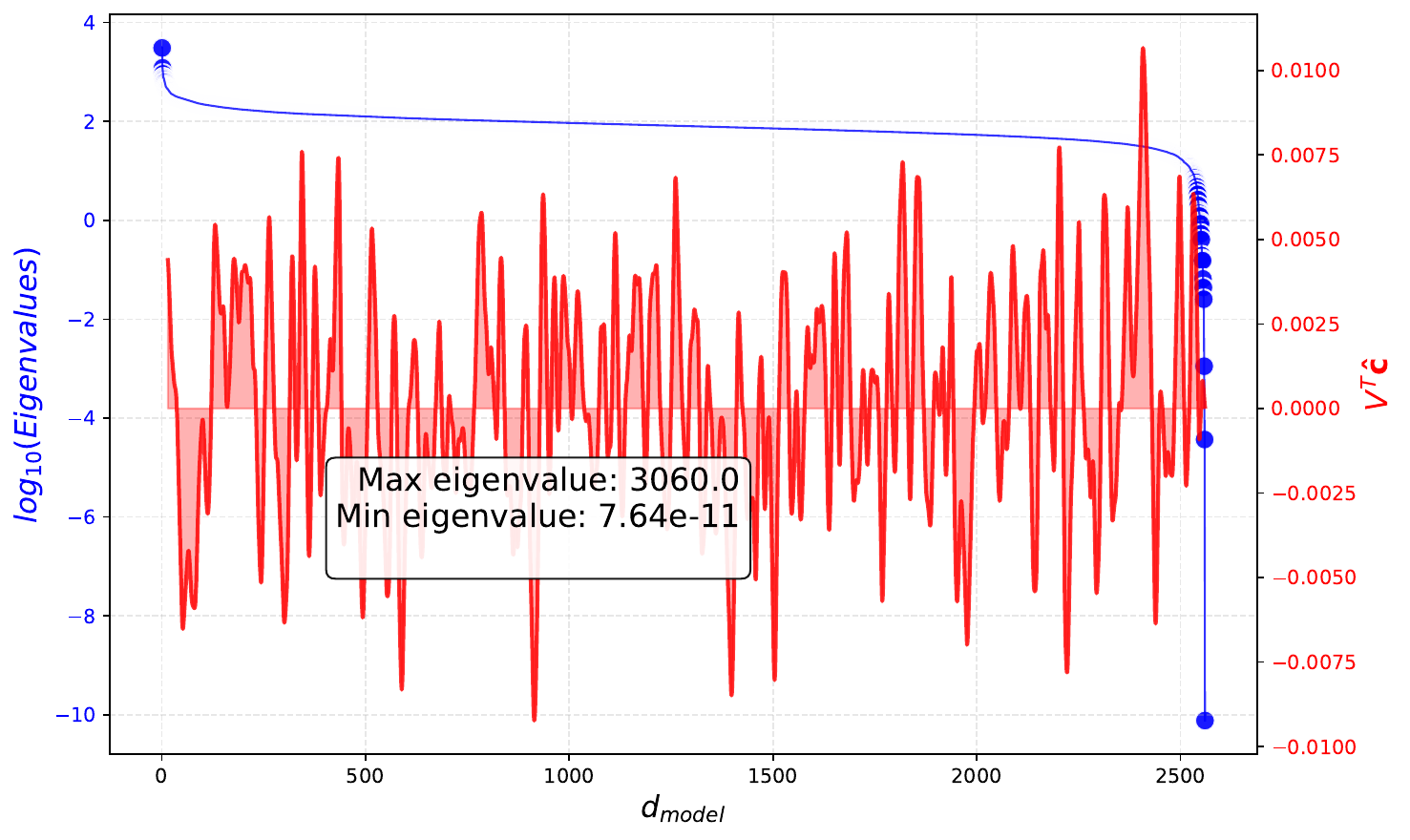}
 \caption{{The figure shows the log of eigenvalues for each principal direction and its alignment with calibration direction $\mathbf{\hat{c}}$, denoting the writing not only in the null space (tail towards the right), but throughout.
  }}
  \label{fig:phi-2_eigenvalue}
\end{figure}

\noindent\textbf{Discovery of a Calibration Subspace:}
Given the observations above, we ask: \textbf{Is there a specific direction in representation space that the model uses to recalibrate predictions?} From layer-wise activation traces, we identify significant representation changes starting from layer 28, precisely when calibration begins to improve while accuracy plateaus (also see App. Figure \ref{fig:phi-2_layer_difference}). We use a simple strategy and define the \textit{calibration direction} $\mathbf{\hat{c}}$ as the mean of the normalized differences between successive layer outputs in the final three layers:
$$
\mathbf{\hat{c}} = \frac{1}{3}(\vec{c}_{29} + \vec{c}_{30} + \vec{c}_{31}), \quad \vec{c}_i = \frac{A_i - A_{i-1}}{\|A_i - A_{i-1}\|}
$$
here, \(A_i\) denotes the residual stream output after layer \(i\). This direction captures the internal shift the model undergoes to improve calibration, without affecting prediction correctness.
We also verify that this calibration direction is \textit{not} aligned with the null space (low singular values) of $\mathbf{W}_U$, as shown in Figure~\ref{fig:phi-2_eigenvalue}, indicating it arises from a distinct mechanism.

\noindent\textbf{Modulating Calibration via Subspace Intervention:}
To test the functional role of the calibration direction, we perturb the residual stream along $\mathbf{\hat{c}}$ during inference:
$$
A_i' = A_i + \eta \mathbf{\hat{c}}, \quad \eta > 0
$$
Figure \ref{fig:all_layers_calibration_intervention_main} shows that this small intervention leads to lower ECE and MCE scores without harming classification accuracy. Remarkably, the effect generalizes across datasets: the direction $\mathbf{\hat{c}}$ computed on the MMLU-Humanities split improves calibration on other MMLU subsets as well (including other datasets like TruthfulQA, Figure \ref{fig:all_layers_calibration_phi-2_mmlu_all_truthfulqa_intervention} (d)).
This suggests the existence of a task-agnostic calibration subspace, i.e., distinct from the prediction subspace, that the model uses to regulate confidence.

\section{Discussion}
Our findings suggest that model calibration is not merely a byproduct of prediction accuracy but a distinct representational property shaped by specific components within the network. The emergence of calibration improvements in the final transformer layers, despite minimal accuracy gains, points to a dedicated phase in the model’s forward pass where confidence is explicitly regulated. The fact that interventions in the residual stream using the found direction $\mathbf{\hat{c}}$ can improve calibration without affecting accuracy further supports the hypothesis that calibration resides in a separate, manipulable subspace. 
While the identified calibration directions show promising results within individual models and some datasets, they are not directly found across different architectures, and more investigations would be needed on similar lines (see \S \ref{sec:limitations} for more details). This lack of generalization (of the found calibration direction) suggests that the directions are partly model- and domain-specific, and motivates future work to identify more universal confidence-modulating features in these layers. We see this work as an initial step toward uncovering/understanding the mechanisms of confidence regulation in LLMs, with future research needed to evaluate generalization across generative tasks, domains, and training regimes.
%
Additionally, the limited but non-negligible role of the unembedding null space reinforces insights from prior work \cite{cancedda2024spectralfiltersdarksignals, stolfo2024confidence-regulation-neurons}, but our layerwise analysis shows that this effect is not isolated to the final projection step. Rather, it is distributed, suggesting a broader calibration mechanism involving intermediate representations.


Taken together, these results open up promising avenues for controllable calibration in LLMs through geometric interventions and motivate deeper exploration into how representations across layers encode not just “what” the model predicts but “how confident” it should be.


\noindent Several parallel lines of research have also investigated calibration dynamics across intermediate representations in deep neural networks, particularly in vision tasks using architectures like ResNets \cite{He2015DeepRL} and VGG \cite{simonyan2015deepconvolutionalnetworkslargescale}. A recent work by \citet{wang2024calibration} identifies a calibration bottleneck in the middle layers of vision models, using linear probes trained on hidden representations. Their analysis reveals a U-shaped trend across layers, i.e., model predictions are more calibrated in the middle layers, with miscalibration increasing again toward the final layers, which they attribute to overcompression of information in the later layers.
In contrast, we observe a different trend in the open-weight transformer-based language models that we experimented with, where the model's calibration is improved/regulated in the later/upper layers before the final predictions are made.
A noteworthy difference between the other studies and our experimental setup is the use of logit lens-style probing, i.e., we directly project residual stream representations into the unembedding space to analyze the model’s own prediction distribution, without training any additional classifiers, which we believe avoids introducing supervision and better reflects the model’s internal confidence dynamics. 
These differences, in both architecture (transformers \cite{NIPS2017_3f5ee243}  vs. CNNs \cite{He2015DeepRL, simonyan2015deepconvolutionalnetworkslargescale}) and methodology (unsupervised probing \cite{logit-lens} vs. trained classifiers \cite{belinkov-2022-probing}), highlight the need for more domain- and architecture-specific analyses when studying calibration behavior in deep models. We believe our findings contribute to this growing literature by presenting an unsupervised, layerwise view of calibration dynamics in large-scale language models, which needs further investigation.

\section{Conclusion and Future Directions} \label{sec:conclusion}

In this work, we conduct an investigation into how large transformer-based language models regulate their confidence across layers. Our analysis uncovers a structured three-phase calibration pattern: an initial \textit{decision formation} phase, a subsequent phase of \textit{overconfidence}, and a final \textit{confidence correction phase} in the upper/later layers. Specifically, in the Phi-2 model, we observe that while accuracy plateaus beyond layer 24, calibration metrics such as ECE and MCE continue to fluctuate, first worsening, then sharply improving, revealing that model confidence is actively corrected even after predictions have stabilized.
This phenomenon points to a dynamic internal mechanism modulating uncertainty across depth. We identify a low-dimensional \textit{calibration direction} in the residual stream that weakly appears to underlie this correction phase. Perturbing this direction improves calibration across layers without degrading accuracy, suggesting that confidence regulation is not confined to the output layer but is instead distributed and tunable throughout the model's forward pass.

These findings extend prior work on entropy neurons by showing that confidence correction is a deliberate, multi-layer process, and that calibration can emerge progressively rather than being finalized at the prediction head. Our work provides some insights for probing this behavior.
Practically, our results raise caution in relying on intermediate layers for downstream decision-making, as they may exhibit high accuracy but poor calibration. However, the ability to adjust confidence post-hoc via the calibration direction suggests new opportunities that need further investigations for early exiting and efficient inference that maintain reliability.
Looking ahead, future work may explore how these layerwise calibration mechanisms arise during pretraining, whether similar correction dynamics generalize across model families and sizes, and how these phenomena can be explicitly modeled or optimized for applications requiring reliable confidence estimates. 
\section*{Limitations} \label{sec:limitations}
One of the primary limitations of this study is that this work restricts its analysis to a single-token classification setting, focusing on multiple-choice question answering tasks. While this setup may not reflect the full complexity of generative language modeling, it allows for a clean and controlled examination of model confidence and calibration using well-established metrics such as Expected Calibration Error (ECE) and Maximum Calibration Error (MCE). This framing avoids the confounding effects introduced by autoregressive decoding, ensuring the interpretability of the results. Extending this analysis to multi-token generation remains an exciting direction for future work, where a deeper understanding of calibration over token sequences and temporal dynamics could be developed.

\noindent Our method for identifying the calibration direction in the residual stream is currently model- and dataset-specific. Although the discovered direction in Phi-2 leads to meaningful calibration improvements without degrading accuracy, it does not generalize to other models such as Mistral or LLaMA-2. This limitation highlights interesting differences in how confidence is regulated across architectures and invites further investigation into whether model-specific inductive biases or training schemes influence the emergence of such calibration structures. Similarly, the confidence correction phase we report is most evident in knowledge acquisition tasks like MMLU, where performance saturates in mid-to-late layers. In contrast, reasoning-based datasets exhibit gradually increasing accuracy, making it harder to isolate calibration behavior independently from prediction quality. We view this as an opportunity to refine analysis tools that can disentangle calibration from competence in such settings.

\noindent Finally, while our approach to defining the calibration direction is based on a simple difference between layerwise residuals, it lays the groundwork for richer strategies. More principled methods, such as those based on optimization, gradient sensitivity to ECE loss, or attribution techniques, could uncover more robust and generalizable directions. We see our current results as a strong proof of concept that invites further methodological development and broader application across tasks and architectures. We believe this line of investigation opens up a promising path toward mechanistically understanding calibration in the coming future.



\section*{Ethical Considerations}



This paper aims to advance the field of machine learning, focusing specifically on model calibration and confidence regulation in transformer architectures. While we do not foresee any immediate ethical concerns arising from the research presented, it is essential to recognize that the broader implications of developing more reliable models include both positive and potentially negative societal consequences. Future applications of these techniques could affect areas such as fairness, bias mitigation, and decision-making in systems built upon LLMs, and it will be critical to assess and address these issues in subsequent research and applications.

\section*{Acknowledgments}

We would like to thank the anonymous reviewers and the meta-reviewer for their insightful comments and suggestions. 
This research work was partially supported by the Research-I Foundation of the Department of CSE at IIT Kanpur. 

\bibliography{references}

\clearpage
\newpage

\appendix

\section*{Appendix}

\appendix


\titlecontents{section}[18pt]{\vspace{0.05em}}{\contentslabel{1.5em}}{}
{\titlerule*[0.5pc]{.}\contentspage} 


\titlecontents{table}[0pt]{\vspace{0.05em}}{\contentslabel{1em}}{}
{\titlerule*[0.5pc]{.}\contentspage} 

\startcontents[appendix] 
\section*{Table of Contents} 
\printcontents[appendix]{section}{0}{\setcounter{tocdepth}{4}} 


\startlist[appendix]{lof} 
\section*{List of Figures} 
\printlist[appendix]{lof}{}{\setcounter{tocdepth}{1}} 

\clearpage
\newpage


\section{Additional Computation Details}
\label{app:additional_computation_details}

\subsection{Residual Stream Computations} \label{app-sec:residual_stream_computations}
The transformer architecture operates by reading from and writing to a residual stream across different layers \cite{elhage2021mathematical}. Each layer applies various transformations (e.g., LayerNorm, Multi-Head Attention, FeedForward) to the residual stream. Mathematically, the operation at each transformer layer can be described as:
\[
z_i = f_i\left(z_0 + \sum_{j=1}^{i-1} z_j\right)
\]
where \(z_0\) is the embedding vector from the embedding matrix \(\mathbf{W}_E\), and \(f_i\) represents the function applied by the \(i\)-th transformer block. These operations modify the residual stream, which ultimately affects the prediction of the model.
This residual stream formulation is central to mechanistic interpretability approaches, providing a lens into how information is incrementally composed between layers \cite{elhage2021mathematical}.

\section{Prompt Templates} \label{app:prompt-templates}

\begin{figure*}[t]
\begin{center}
\scalebox{0.85}{
    \begin{tabular}{p{1.1\linewidth}}
      \toprule
      \texttt{Following are some multiple choice questions. You should directly answer the question by choosing the correct option.}\\
      \texttt{\textcolor{blue}{[ in-context examples (if few-shot/in-context learning experiment) ]}} \\
      \texttt{Question: \textcolor{teal}{A generalised statement pertaining to the task} -:  \textcolor{orange}{question/statement} 
      }\\
      \texttt{A. \textcolor{blue}{\textcolor{orange}{choice1}}} \\
      \texttt{B. \textcolor{blue}{\textcolor{orange}{choice2}}} \\
      \texttt{Answer:\textcolor{red}{ \underline{ A}}} \\
      \bottomrule
    \end{tabular}
    }
\caption[Input Prompt Template]{Input prompt formats for the MCQA-based evaluation of autoregressive open-weight models , (e.g., \texttt{llama(-3)}, \texttt{Phi-2}, etc.).
The \texttt{black text} is the templated input for all datasets. The \texttt{\textcolor{orange}{orange text}} is the input from the datasets which contains either a review or a statement or a question. 
The \texttt{\textcolor{teal}{teal text}} is a template comment describing the task, which changes according to the dataset
 The next-token prediction probabilities of the option IDs at the \textcolor{red}{\underline{\texttt{red text}}} is used as the observed prediction distribution. 
}
\label{app-fig:question-prompt}
\end{center}
\end{figure*}

\begin{figure*}[ht]
\begin{center}
\scalebox{0.85}{
    \begin{tabular}{p{1.1\linewidth}}
      \toprule
      \texttt{Following are some multiple choice questions. You should directly answer the question by choosing the correct option.}\\
      \texttt{\textcolor{blue}{[ in-context examples (if few-shot/in-context learning experiment) ]}} \\
      \texttt{Question: \textcolor{orange}{Mars has an atmosphere that is almost entirely carbon dioxide. Why isn't there a strong greenhouse effect keeping the planet warm?} 
      }\\
      \texttt{A: \textcolor{blue}{\textcolor{orange}{the atmosphere on Mars is too thin to trap a significant amount of heat}}} \\
      \texttt{B: \textcolor{blue}{\textcolor{orange}{There actually is a strong greenhouse effect and Mars would be 35oC colder than it is now without it.}}} \\
      \texttt{C: \textcolor{blue}{\textcolor{orange}{Mars does not have enough internal heat to drive the greenhouse effect}}} \\
      \texttt{D: \textcolor{blue}{\textcolor{orange}{the greenhouse effect requires an ozone layer which Mars does not have}}} \\
      \texttt{Answer:\textcolor{red}{ \underline{ A}}} \\
      \bottomrule
    \end{tabular}
    }
\caption[Input Prompt Example]{Input prompt formats for the MCQA-based evaluation of autoregressive open-weight models (e.g., \texttt{llama(-2)}, \texttt{Phi-2}, etc.).
The \texttt{black text} is the templated input for all datasets. The \texttt{\textcolor{orange}{orange text}} is the input from the \textbf{MMLU dataset}.
 The next-token prediction probabilities of the option IDs at the \textcolor{red}{\underline{\texttt{red text}}} are used as the observed prediction distribution. 
\label{app-fig:mmlu_question_prompt}
}
\end{center}
\end{figure*}

\begin{figure*}[ht]
\begin{center}
\scalebox{0.85}{
    \begin{tabular}{p{1.1\linewidth}}
      \toprule
      \texttt{Following are some multiple choice questions. You should directly answer the question by choosing the correct option.}\\
      \texttt{\textcolor{blue}{[ in-context examples (if few-shot/in-context learning experiment) ]}} \\
      \texttt{Question: \textcolor{teal}{Select the suitable option for the following statement} -: \textcolor{orange}{enchanted with low-life tragedy and liberally seasoned with emotional outbursts . . . what is sorely missing, however, is the edge of wild, lunatic invention that we associate with cage's best acting .} 
      }\\
      \texttt{A: \textcolor{blue}{\textcolor{orange}{Negative}}} \\
      \texttt{B: \textcolor{blue}{\textcolor{orange}{Positive}}} \\
      \texttt{Answer:\textcolor{red}{ \underline{ A}}} \\
      \bottomrule
    \end{tabular}
    }
\caption[Rotten Tomatoes dataset Prompt Template]{Input prompt formats for the MCQA-based evaluation of autoregressive open-weight models , (e.g., \texttt{llama(-3)}, \texttt{Phi-2}, etc.).
The \texttt{black text} is the templated input for all datasets. The \texttt{\textcolor{orange}{orange text}} is the input from the \textbf{Rotten Tomatoes dataset}.
The \texttt{\textcolor{teal}{teal text}} is a template comment describing the task.
 The next-token prediction probabilities of the option IDs at the \textcolor{red}{\underline{\texttt{red text}}} are used as the observed prediction distribution. }
\label{app-fig:rottentomatoes_question_prompt}
\end{center}
\end{figure*}

\begin{figure*}[ht]
\begin{center}
\scalebox{0.85}{
    \begin{tabular}{p{1.1\linewidth}}
      \toprule
      \texttt{Following are some multiple choice questions. You should directly answer the question by choosing the correct option.}\\
      \texttt{\textcolor{blue}{[ in-context examples (if few-shot/in-context learning experiment) ]}} \\
      \texttt{Question: \textcolor{teal}{ Which of the following events (given as options A or B) is a more plausible} \textcolor{orange}{effect} \textcolor{teal}{of the event} -: \textcolor{orange}{'The woman betrayed her friend.'?} 
      }\\
      \texttt{A: \textcolor{blue}{\textcolor{orange}{Her friend sent her a greeting card.}}} \\
      \texttt{B: \textcolor{blue}{\textcolor{orange}{Her friend cut off contact with her.}}} \\
      \texttt{Answer:\textcolor{red}{ \underline{ B}}} \\
      \bottomrule
    \end{tabular}
    }
\caption[COPA dataset Prompt Template]{Input prompt formats for the MCQA-based evaluation of autoregressive open-weight models , (e.g., \texttt{llama(-3)}, \texttt{Phi-2}, etc.).
The \texttt{black text} is the templated input for all datasets. The \texttt{\textcolor{orange}{orange text}} is the input from the \textbf{COPA dataset}.
The \texttt{\textcolor{teal}{teal text}} is a template comment describing the task.
 The next-token prediction probabilities of the option IDs at the \textcolor{red}{\underline{\texttt{red text}}} are used as the observed prediction distribution. }
\label{app-fig:copa_question_prompt}
\end{center}
\end{figure*}

\begin{figure*}[ht]
\begin{center}
\scalebox{0.85}{
    \begin{tabular}{p{1.1\linewidth}}
      \toprule
      \texttt{Following are some multiple choice questions. You should directly answer the question by choosing the correct option.}\\
      \texttt{\textcolor{blue}{[ in-context examples (if few-shot/in-context learning experiment) ]}} \\
      \texttt{Question: \textcolor{teal}{Select the suitable option for the following statement} -: \textcolor{orange}{The cat was bitten the mouse.} 
      }\\
      \texttt{A: \textcolor{blue}{\textcolor{orange}{Unacceptable}}} \\
      \texttt{B: \textcolor{blue}{\textcolor{orange}{Acceptable}}} \\
      \texttt{Answer:\textcolor{red}{ \underline{ A}}} \\
      \bottomrule
    \end{tabular}
    }
\caption[CoLA dataset Prompt Template]{Input prompt formats for the MCQA-based evaluation of autoregressive open-weight models , (e.g., \texttt{llama(-3)}, \texttt{Phi-2}, etc.).
The \texttt{black text} is the templated input for all datasets. The \texttt{\textcolor{orange}{orange text}} is the input from the \textbf{CoLA dataset}.
The \texttt{\textcolor{teal}{teal text}} is a template comment describing the task.
 The next-token prediction probabilities of the option IDs at the \textcolor{red}{\underline{\texttt{red text}}} are used as the observed prediction distribution. }
\label{app-fig:cola_question_prompt}
\end{center}
\end{figure*}

 For all our experiments, we follow a standard prompt template. This section provides the details of the prompt templates used in our multiple-choice question answering (MCQA) evaluations of autoregressive open-weight language models (e.g., \texttt{LLaMA(-2)}, \texttt{Phi-2}). All prompts follow a unified format to ensure consistency across tasks and models.
Figure \ref{app-fig:question-prompt} presents the general template used. Each prompt begins with an instruction directing the model to select the correct answer from a set of multiple-choice options. In few-shot or in-context learning settings, this instruction is optionally followed by a set of in-context examples. The question is prefaced by a task-specific description, and followed by a list of labeled answer choices (A, B, C, etc.).
The final line of the prompt contains the prefix \texttt{Answer:}, which serves as the model’s response cue. For evaluation, we extract the model’s next-token prediction probabilities at this position over the answer option tokens (e.g., \texttt{A}, \texttt{B}), which we treat as the model’s predicted distribution over choices.

Figure \ref{app-fig:mmlu_question_prompt} illustrates a fully instantiated example using a question from the MMLU dataset. The correct answer is provided at the end of the prompt and underlined. The colored annotations in the figure denote fixed template components (in black) and variable elements drawn from the dataset (in orange and teal).

This templated format is applied uniformly across datasets and experimental configurations, enabling controlled comparisons of model behavior across domains and prompting setups.

\section{Dataset Details}
\label{app-sec:dataset-details}

We evaluate model calibration and confidence dynamics across a diverse set of NLP benchmarks, selected to cover a range of reasoning, linguistic, and factual understanding capabilities. Below, we detail each dataset used in our experiments:

\noindent\textbf{MMLU (Massive Multitask Language Understanding) \citep{hendrycks2021measuringmassivemultitasklanguage}.}
MMLU is a comprehensive benchmark designed to test knowledge and problem-solving ability across 57 diverse subjects, spanning STEM, humanities, social sciences, law, medicine, and more. Each instance consists of a multiple-choice question with four answer options. Unlike many standard MCQA datasets, MMLU introduces lexical and semantic complexity by using dynamically varying answer choices, which increases the challenge for models to generalize and calibrate effectively across categories. This benchmark is widely adopted for evaluating pretraining quality in LLMs.

\noindent\textbf{CoLA (Corpus of Linguistic Acceptability) \citep{warstadt2019neural}: }
CoLA is a binary classification task that requires models to judge whether a given English sentence is grammatically acceptable. The dataset is drawn from linguistic publications and includes a broad spectrum of syntactic phenomena, making it a strong test of a model’s grasp of grammatical rules. Performance is typically measured using Matthews Correlation Coefficient (MCC) and accuracy, providing a nuanced view of linguistic acceptability modeling.

\noindent\textbf{COPA (Choice of Plausible Alternatives) \citep{gordon-etal-2012-semeval}: }
COPA is a commonsense reasoning benchmark where the task is to select the most plausible cause or effect given a premise. Each instance presents two alternatives, and the model must determine which one best explains or results from the premise. The task challenges causal inference and contextual reasoning and is part of the SuperGLUE benchmark suite \cite{wang2020supergluestickierbenchmarkgeneralpurpose}. Accuracy is used as the primary evaluation metric.

\noindent\textbf{Rotten Tomatoes \citep{rottentomatoes}: }
This is a sentiment classification benchmark consisting of short movie reviews labeled as positive or negative. Reviews are often limited to a sentence or short paragraph, focusing on fine-grained lexical and compositional sentiment cues. This dataset is widely used to benchmark sentiment understanding and general text classification performance in LLMs.

\noindent\textbf{TruthfulQA \citep{lin2022truthfulqameasuringmodelsmimic}: }
TruthfulQA evaluates a model’s ability to provide factually correct and non-deceptive answers. It includes questions across multiple domains, such as science, health, and politics, carefully crafted to elicit plausible but incorrect responses from language models trained on internet-scale data. The benchmark serves as a diagnostic tool for hallucination and misinformation. Evaluations are conducted using truthfulness and informativeness scores, often involving human or model-based judgment. In this work, we specifically use a binary version of the dataset, framing an MCQA query for quantifying calibration.

Details of all the prompt templates and the MCQA formulations, for all the datasets used, are provided in our codebase.

\begin{table*}[t]
\centering
\small
\setlength\tabcolsep{6pt}
\renewcommand{\arraystretch}{1.1}
\label{app-tab:dataset-desc}
\begin{tabular}{lcccc}
\toprule
\textbf{Dataset} & \textbf{Task Type} & \textbf{\# Samples} & \textbf{Avg. Prompt Length (tokens)} & \textbf{\# Choices} \\
\midrule
\textbf{MMLU (STEM)}            & Subject Knowledge (STEM)         & 3{,}018 & 149.09 & 4 \\
\textbf{MMLU (Humanities)}      & Subject Knowledge (Humanities)   & 4{,}705 & 535.10 & 4 \\
\textbf{MMLU (Social Sciences)} & Subject Knowledge (Soc. Sci.)    & 3{,}077 & 116.35 & 4 \\
\textbf{MMLU (Other)}           & Subject Knowledge (Misc.)        & 3{,}242 & 163.32 & 4 \\
\textbf{CoLA}                   & Grammatical Acceptability        & 1{,}043 & 41.83  & 2 \\
\textbf{COPA}                   & Causal Commonsense Reasoning     & 1{,}000 & 34.89  & 2 \\
\textbf{Rotten Tomatoes}        & Sentiment Classification         & 1{,}066 & 115.52 & 2 \\
\textbf{TruthfulQA (Binary)}        &   Factual Knowledge      & 790 & 159.21  & 2 \\
\bottomrule
\end{tabular}
\caption{\textbf{Dataset Overview.} Summary of datasets used in our evaluation. We report the type of reasoning or knowledge tested, number of samples used, average prompt length (in tokens), and number of answer choices. MMLU categories are grouped based on domain.}
\end{table*}



\section{Extended Model Evaluations}

\subsection{Calibration Dynamics in Mistral and LLaMA-2 7B}

To assess the generality of our findings, we evaluate calibration behavior in two additional open-weight transformer models: Mistral-7B and LLaMA-2-7B, along with LLaMA-3-8B. Across all three models, we observe an interesting pattern consistent with Phi-2: a \textit{confidence correction phase} emerges in the later layers, characterized by stabilization of accuracy and a sharp improvement in calibration metrics such as Expected Calibration Error (ECE) and Maximum Calibration Error (MCE).

This suggests that the confidence correction behavior is not unique to Phi-2 but may reflect a broader inductive bias of transformer-based language models, where model confidence is actively adjusted after the prediction has converged. We observe this phenomenon in both Mistral and LLaMA variants, although the sharpness and layerwise extent of the correction vary slightly across models and tasks.

However, our interventional experiments using a learned \textit{calibration direction}, a low-dimensional vector in the residual stream found to modulate confidence in Phi-2, did not generalize well to other models. Attempts to extract and apply similar directions in Mistral-7B, LLaMA-2-7B, and LLaMA-3-8B yielded inconsistent results and failed to produce consistent improvements in calibration metrics. This suggests that while the calibration phase itself may be general, the specific encoding of confidence control within the residual stream may vary significantly across architectures and training regimes.

These findings point toward a promising avenue for future research: understanding how architectural or training factors give rise to shared calibration dynamics, and why certain models encode more “steerable” calibration subspaces than others.

\section{Reliability Diagrams Across Layers}

\begin{figure*}[t]
\centering
\captionsetup[subfigure]{labelformat=parens}

\newcommand{\scalefactor}{0.235} 

\begin{subfigure}[b]{\scalefactor\linewidth}
    \includegraphics[width=\linewidth]{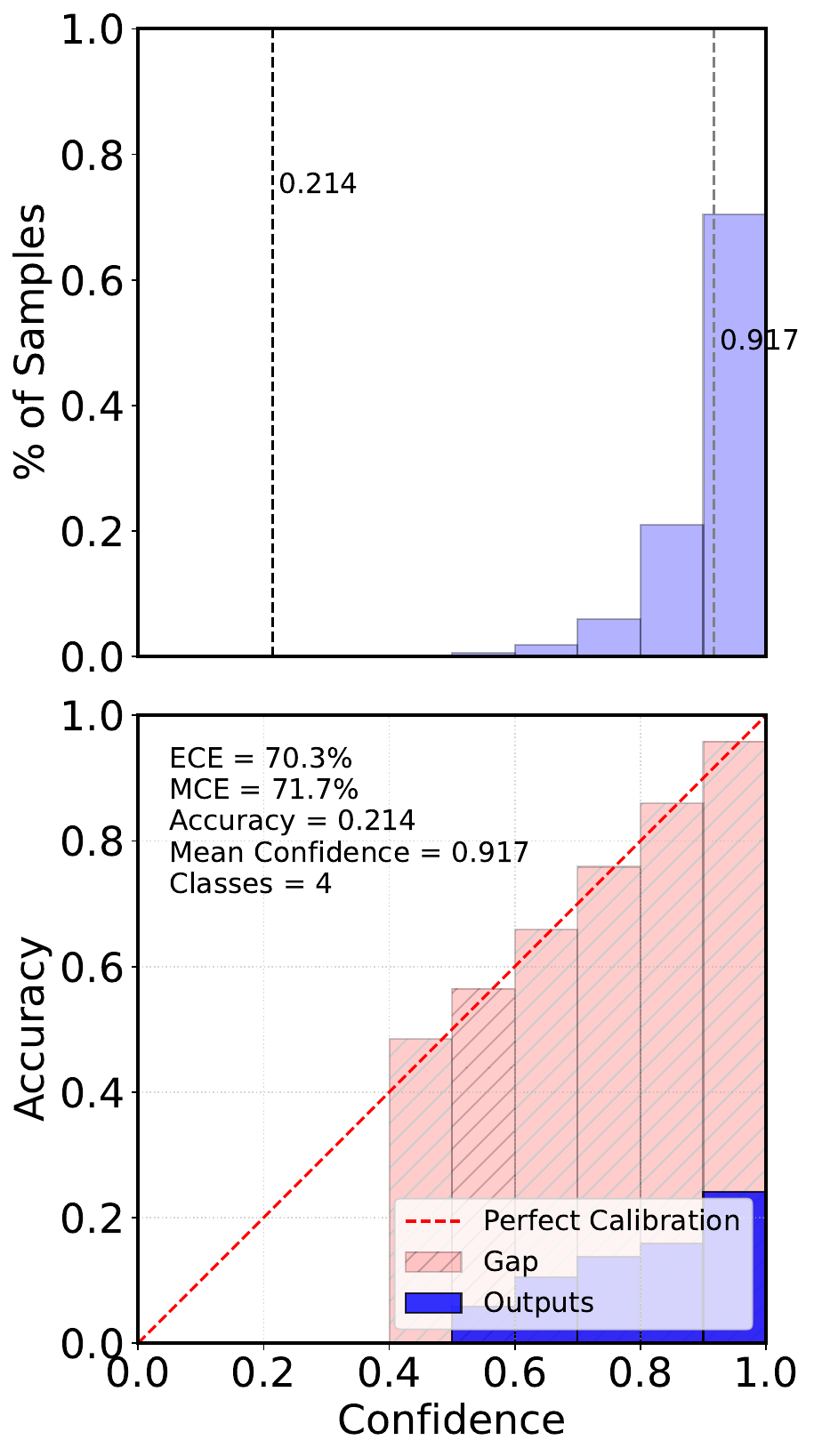}
    \caption{Layer 20}
\end{subfigure}\hfill
\begin{subfigure}[b]{\scalefactor\linewidth}
    \includegraphics[width=\linewidth]{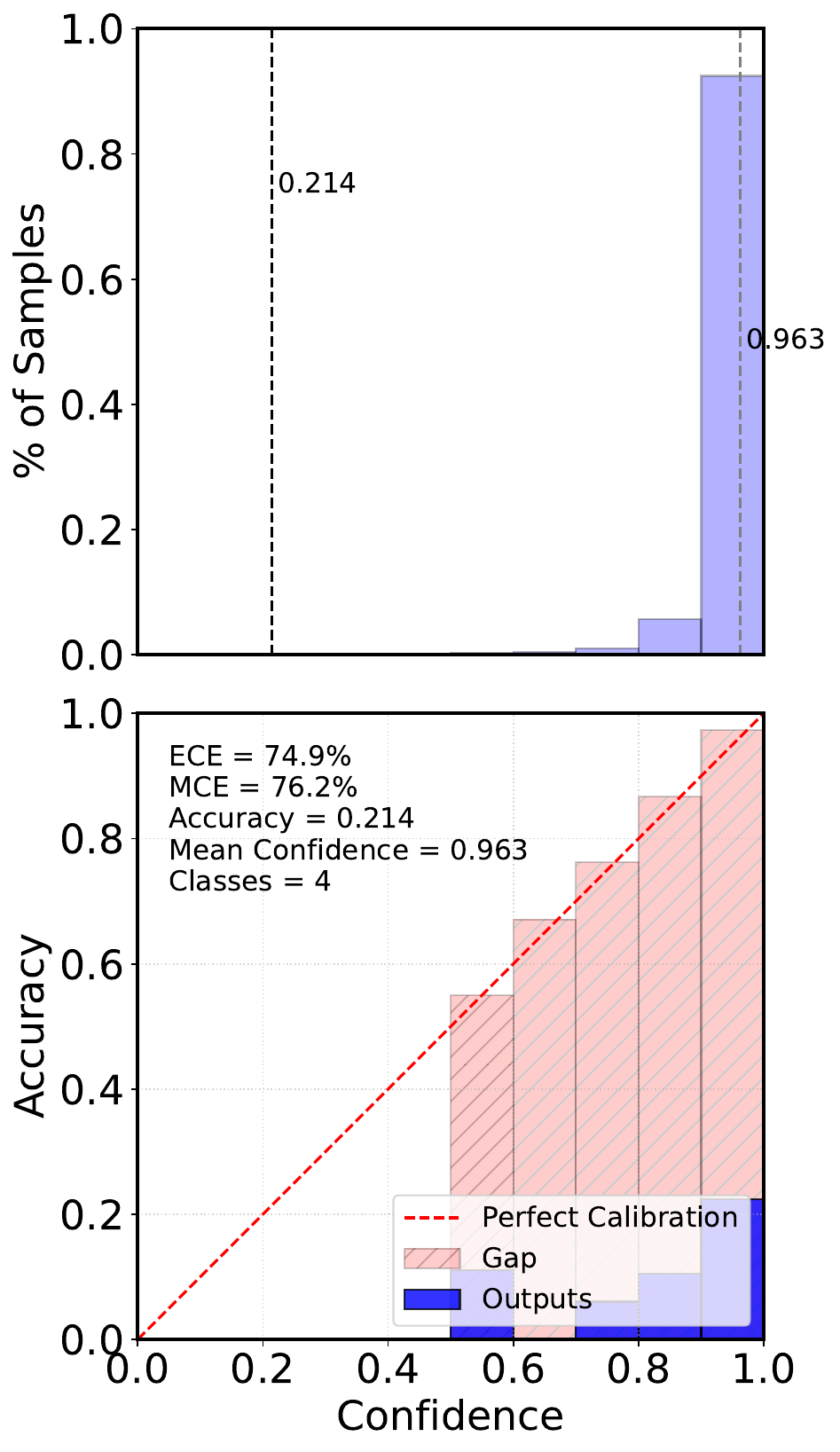}
    \caption{Layer 21}
\end{subfigure}\hfill
\begin{subfigure}[b]{\scalefactor\linewidth}
    \includegraphics[width=\linewidth]{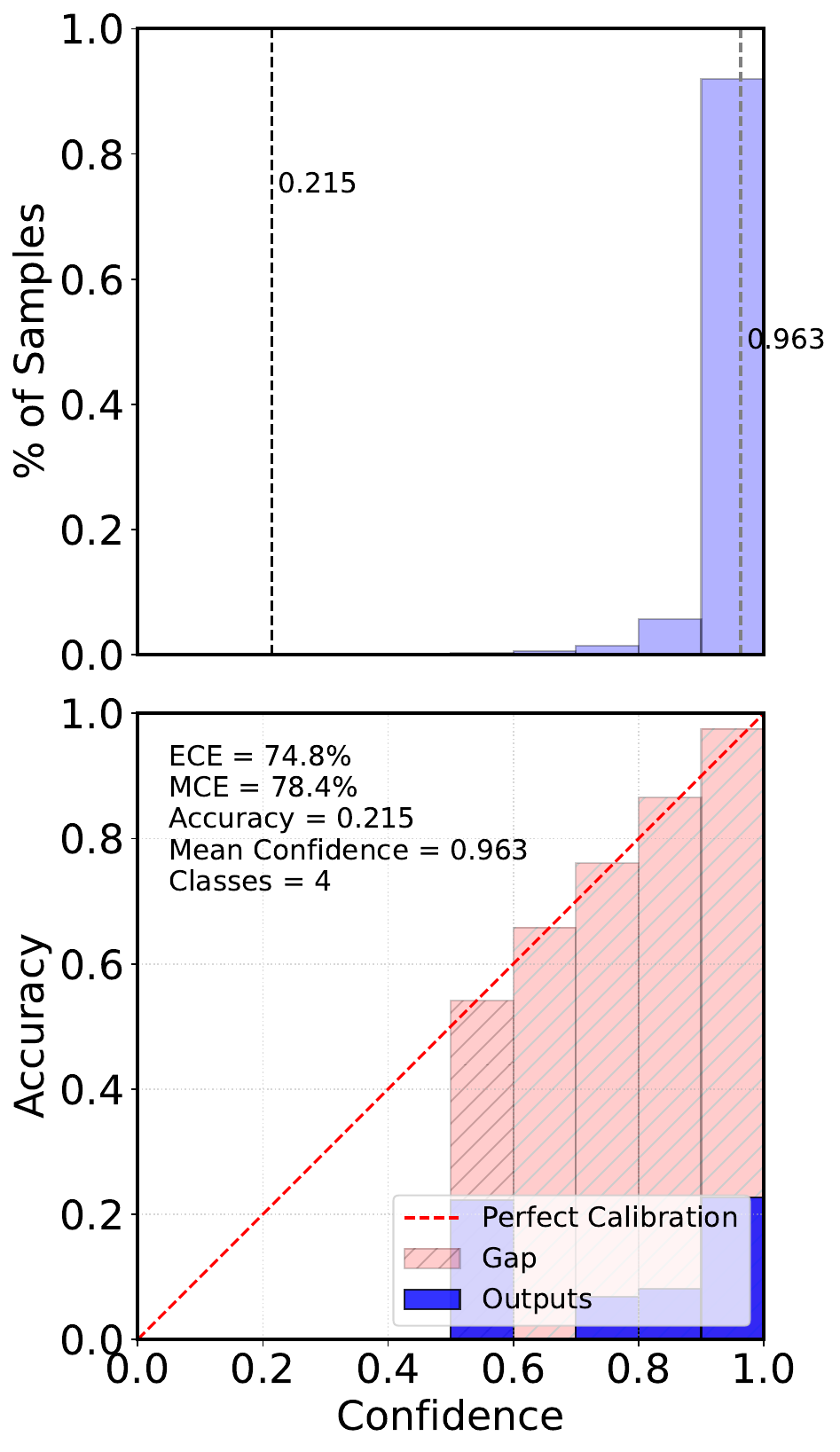}
    \caption{Layer 22}
\end{subfigure}
\hfill
\begin{subfigure}[b]{\scalefactor\linewidth}
    \includegraphics[width=\linewidth]{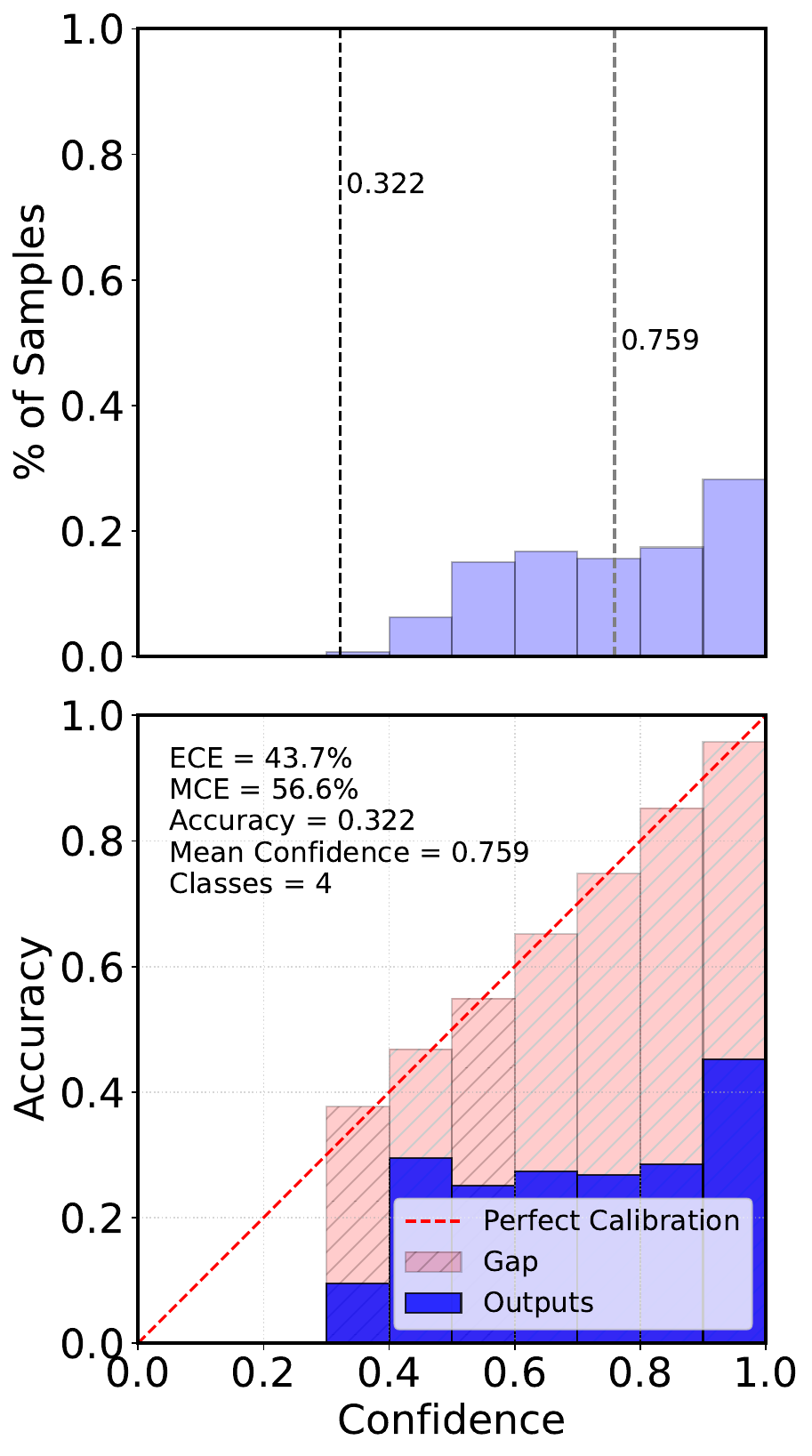}
    \caption{Layer 23}
\end{subfigure}

\begin{subfigure}[b]{\scalefactor\linewidth}
    \includegraphics[width=\linewidth]{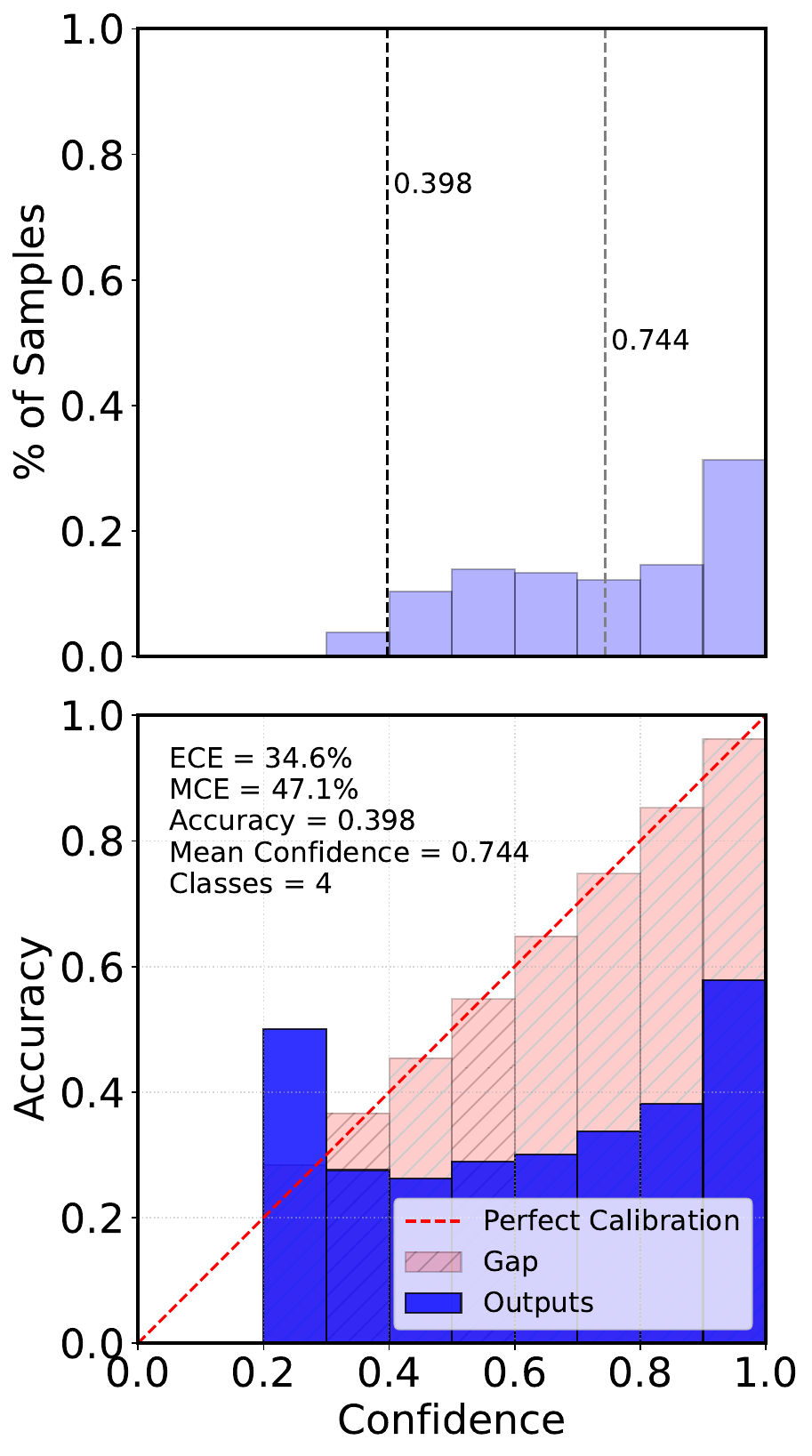}
    \caption{Layer 24}
\end{subfigure}\hfill
\begin{subfigure}[b]{\scalefactor\linewidth}
    \includegraphics[width=\linewidth]{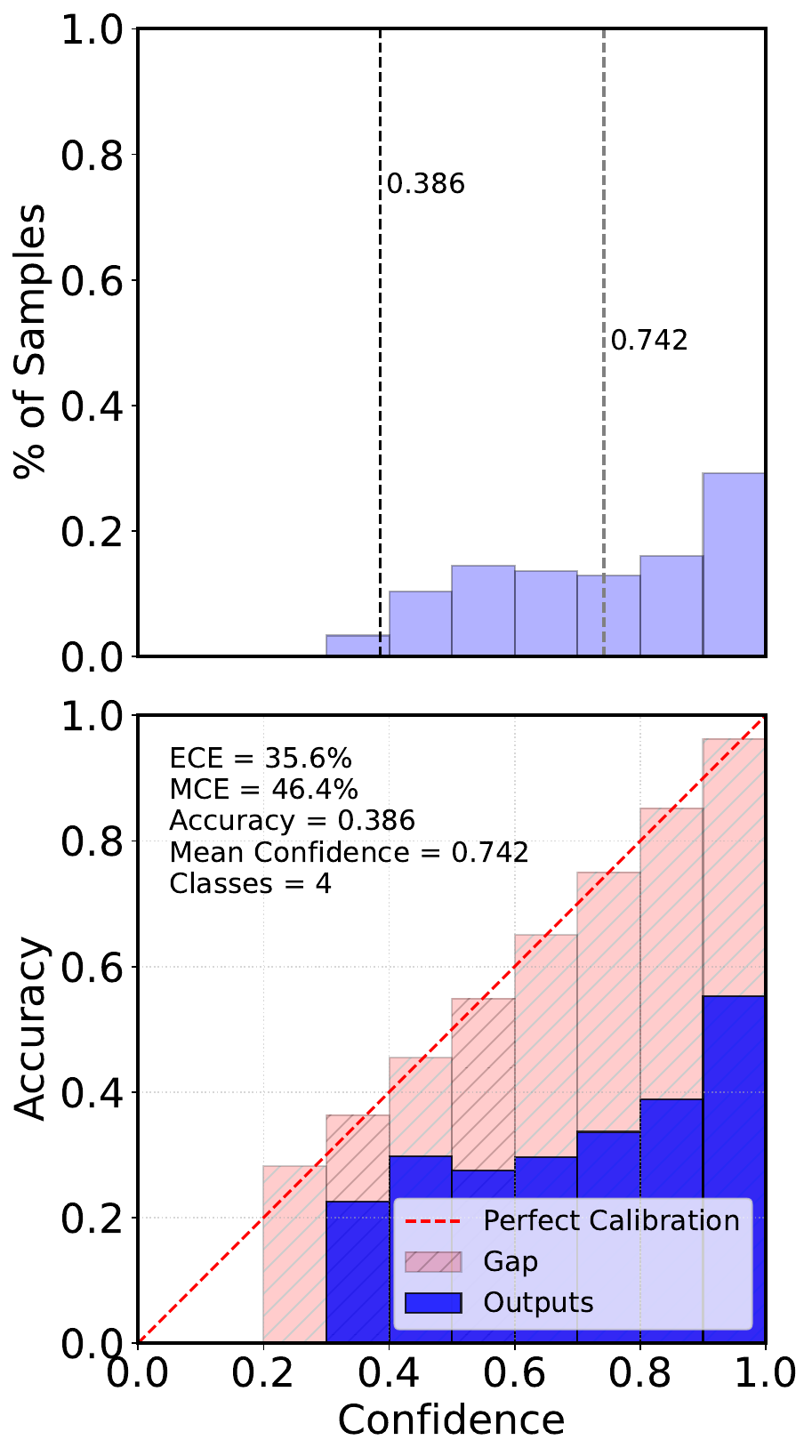}
    \caption{Layer 25}
\end{subfigure}\hfill
\begin{subfigure}[b]{\scalefactor\linewidth}
    \includegraphics[width=\linewidth]{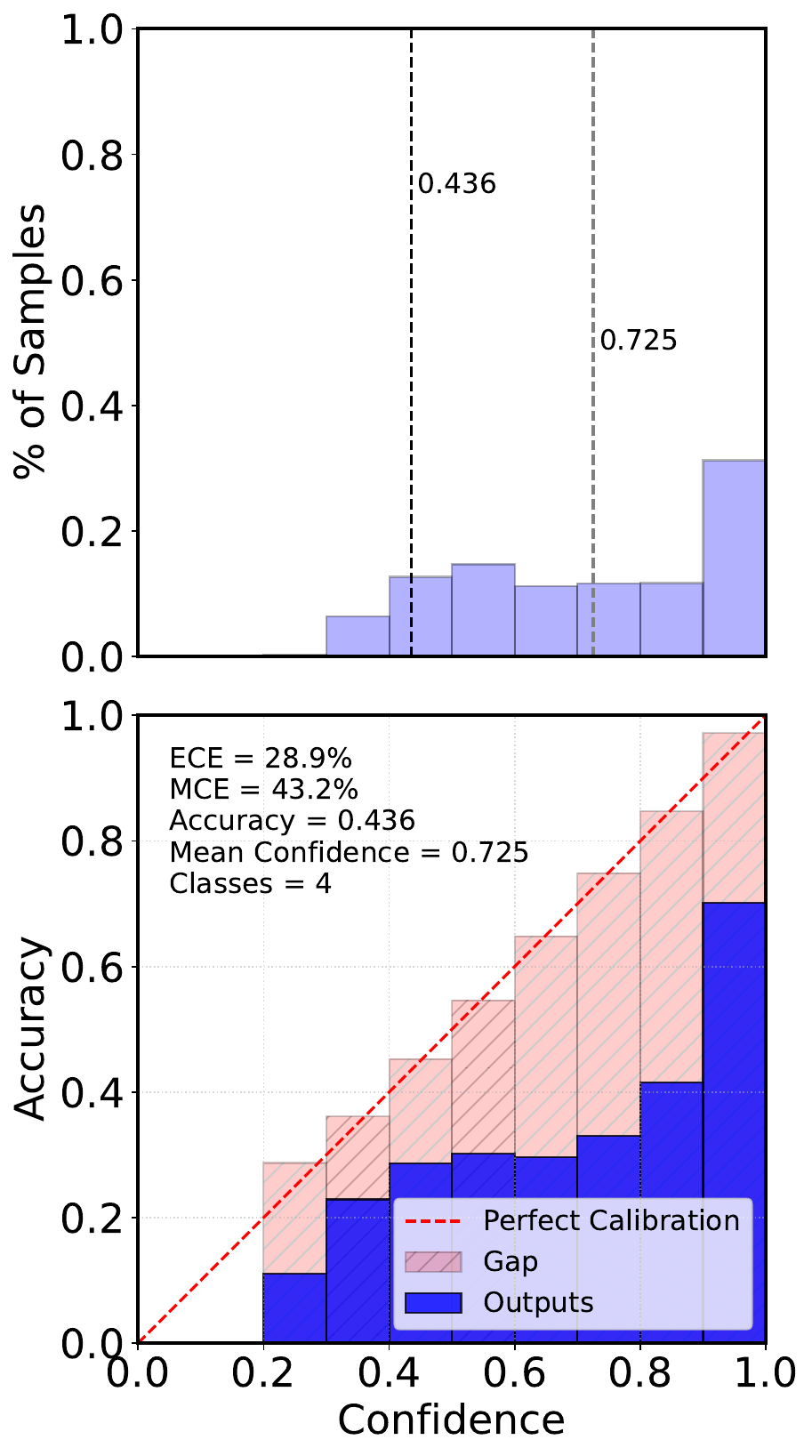}
    \caption{Layer 26}
\end{subfigure}\hfill
\begin{subfigure}[b]{\scalefactor\linewidth}
    \includegraphics[width=\linewidth]{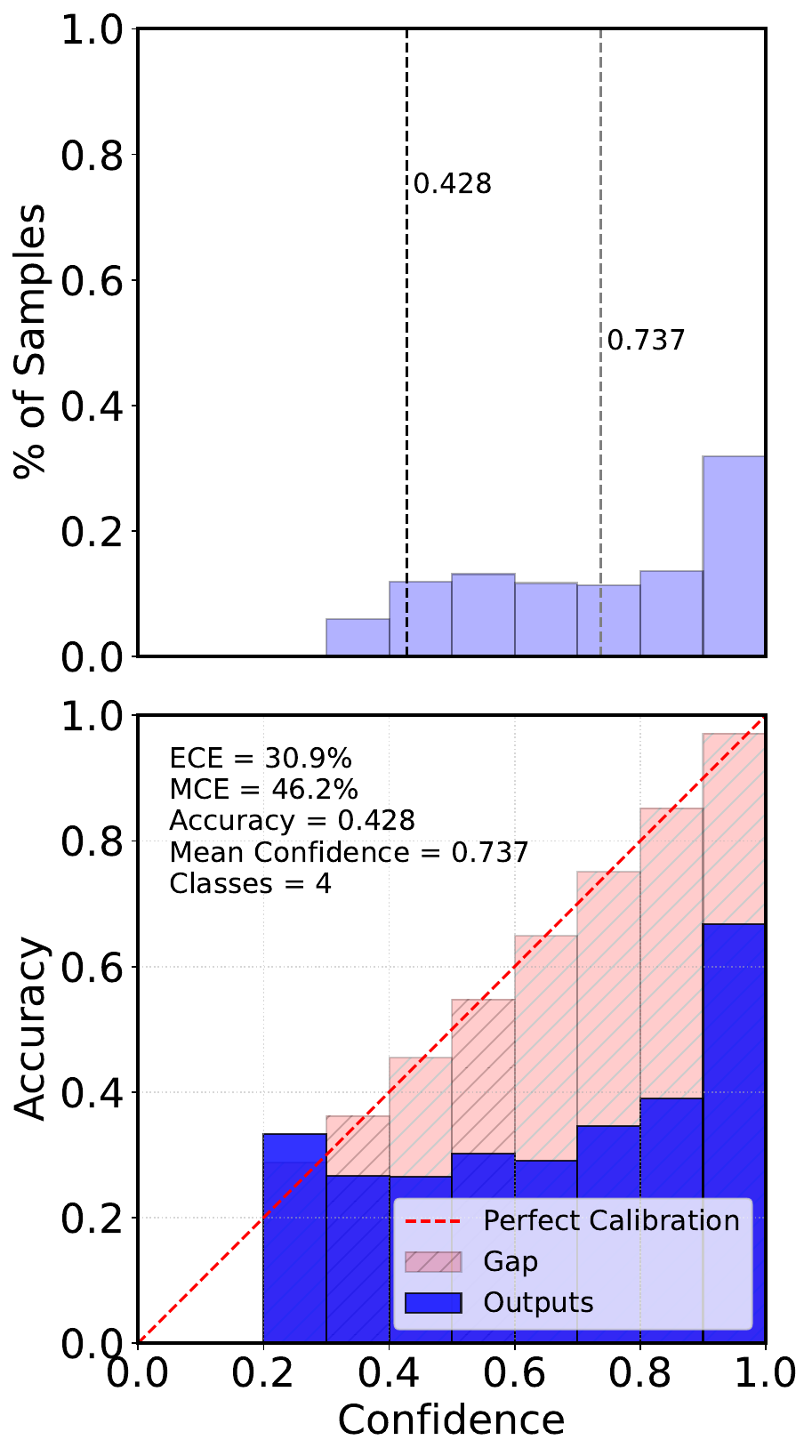}
    \caption{Layer 27}
\end{subfigure}


\begin{subfigure}[b]{\scalefactor\linewidth}
    \includegraphics[width=\linewidth]{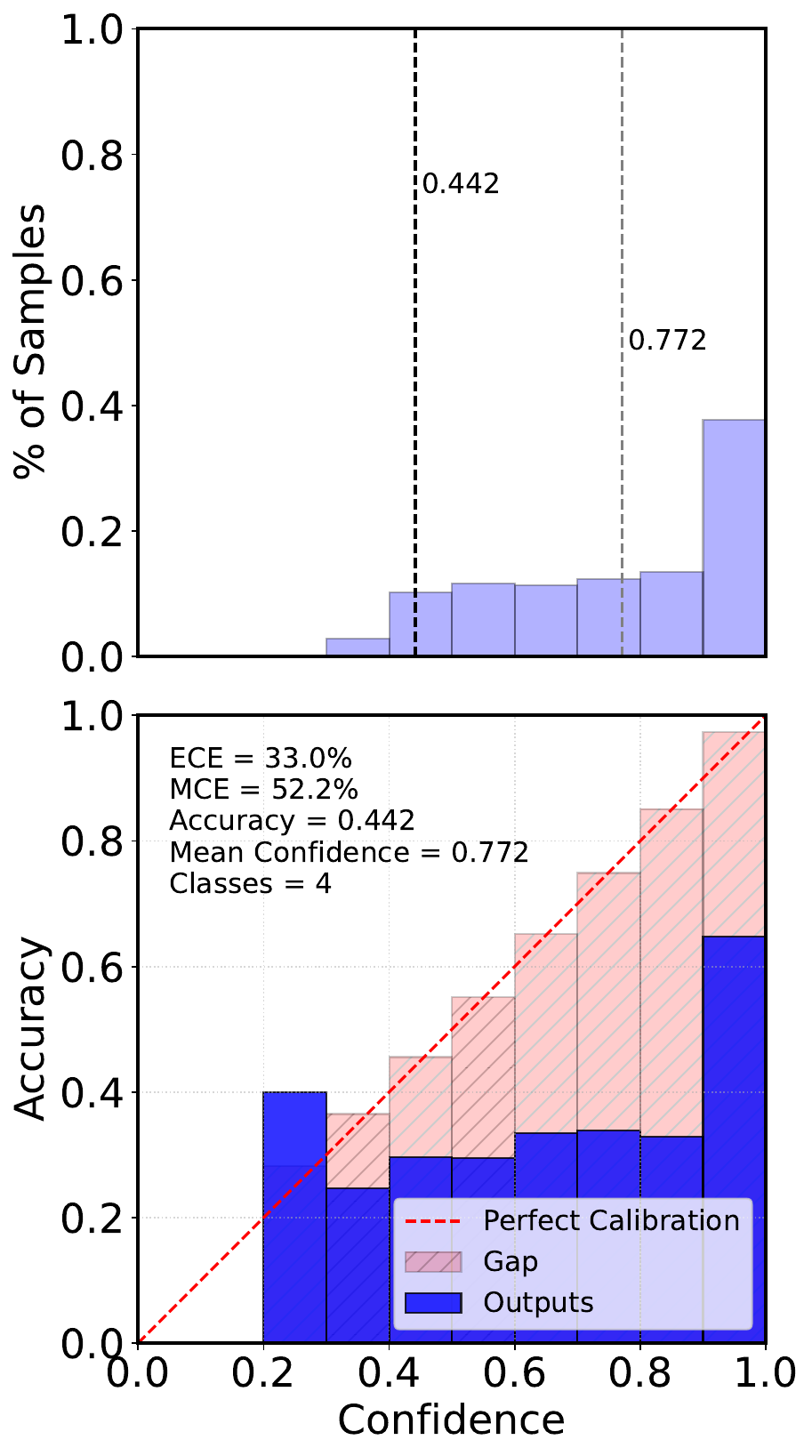}
    \caption{Layer 28}
\end{subfigure}\hfill
\begin{subfigure}[b]{\scalefactor\linewidth}
    \includegraphics[width=\linewidth]{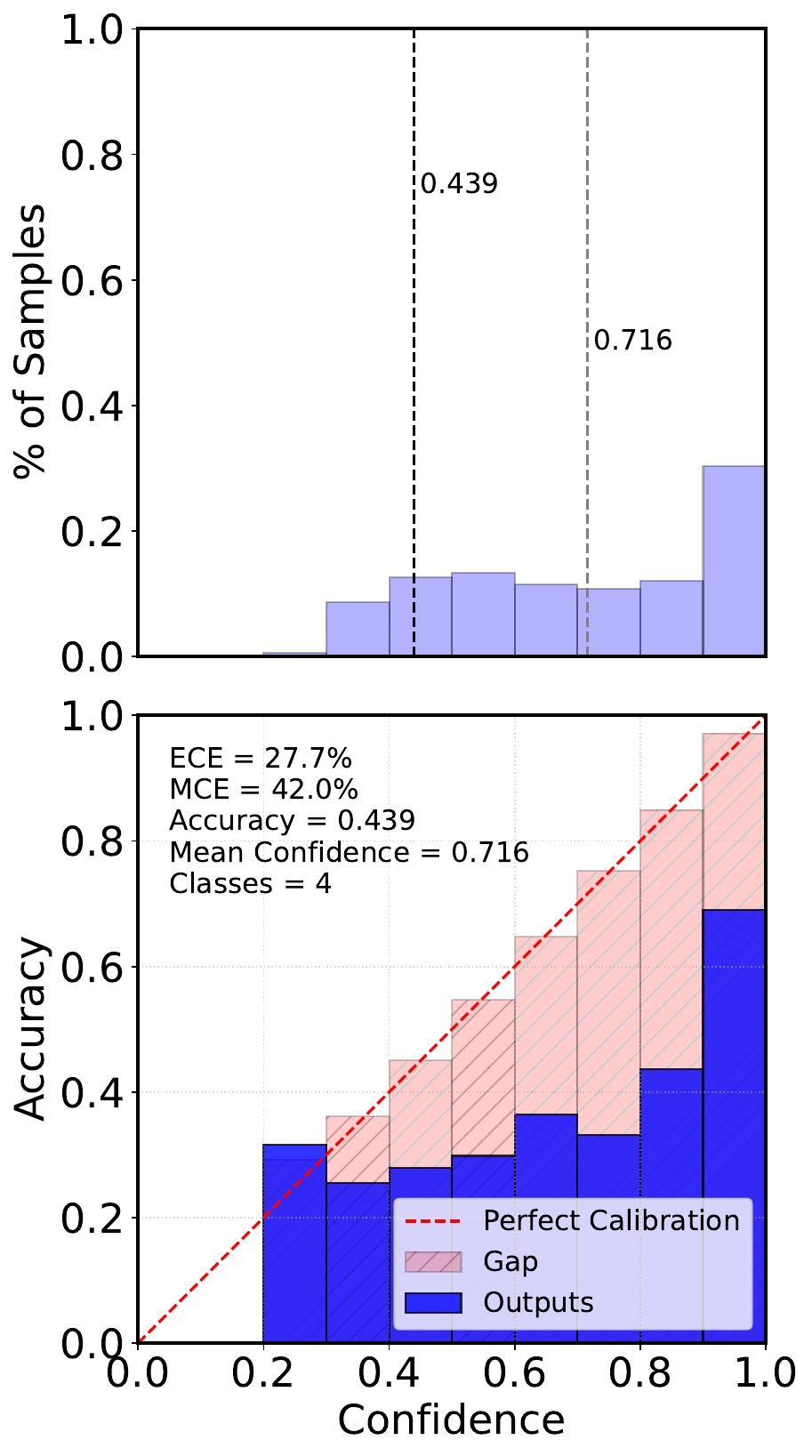}
    \caption{Layer 29}
\end{subfigure}\hfill
\begin{subfigure}[b]{\scalefactor\linewidth}
    \includegraphics[width=\linewidth]{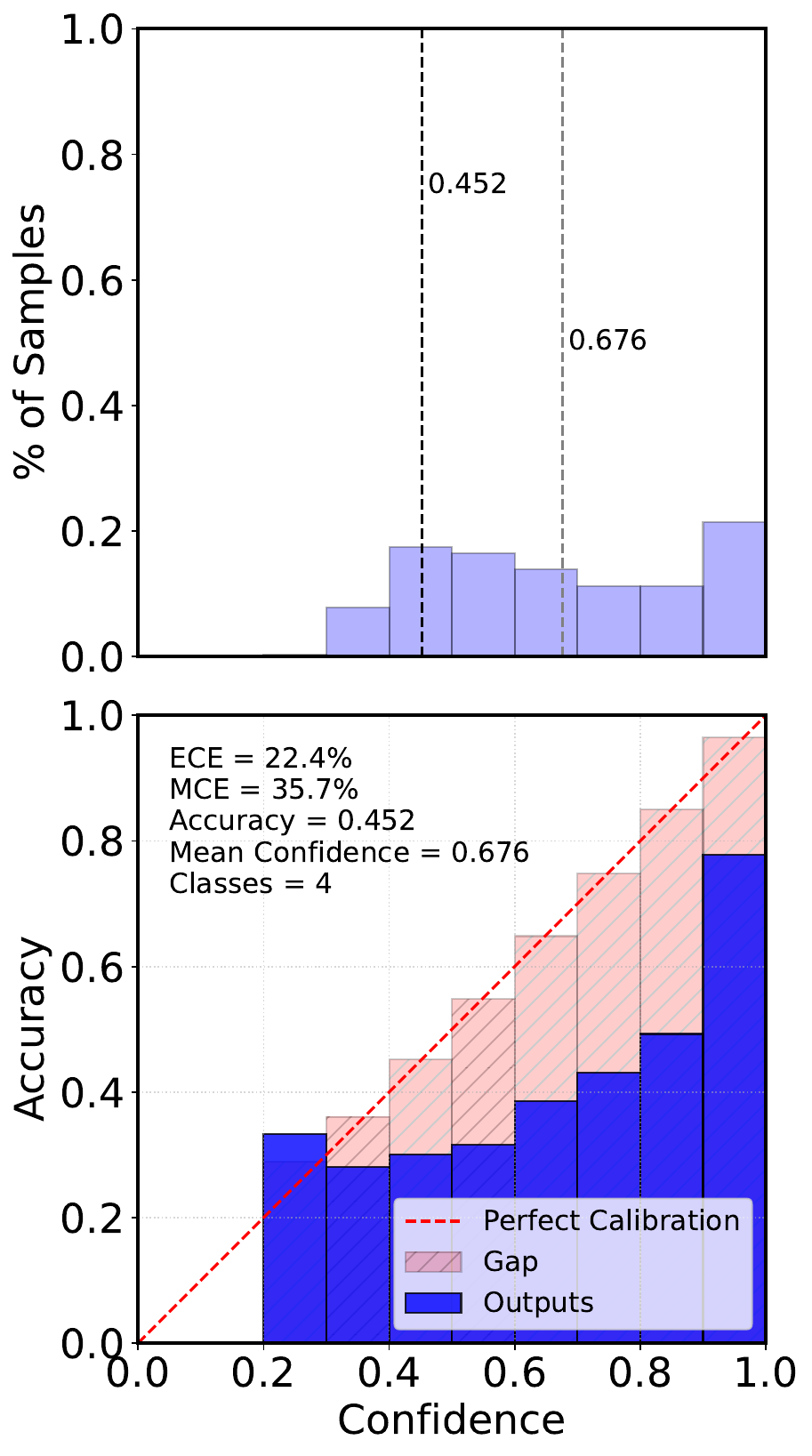}
    \caption{Layer 30}
\end{subfigure}\hfill
\begin{subfigure}[b]{\scalefactor\linewidth}
    \includegraphics[width=\linewidth]{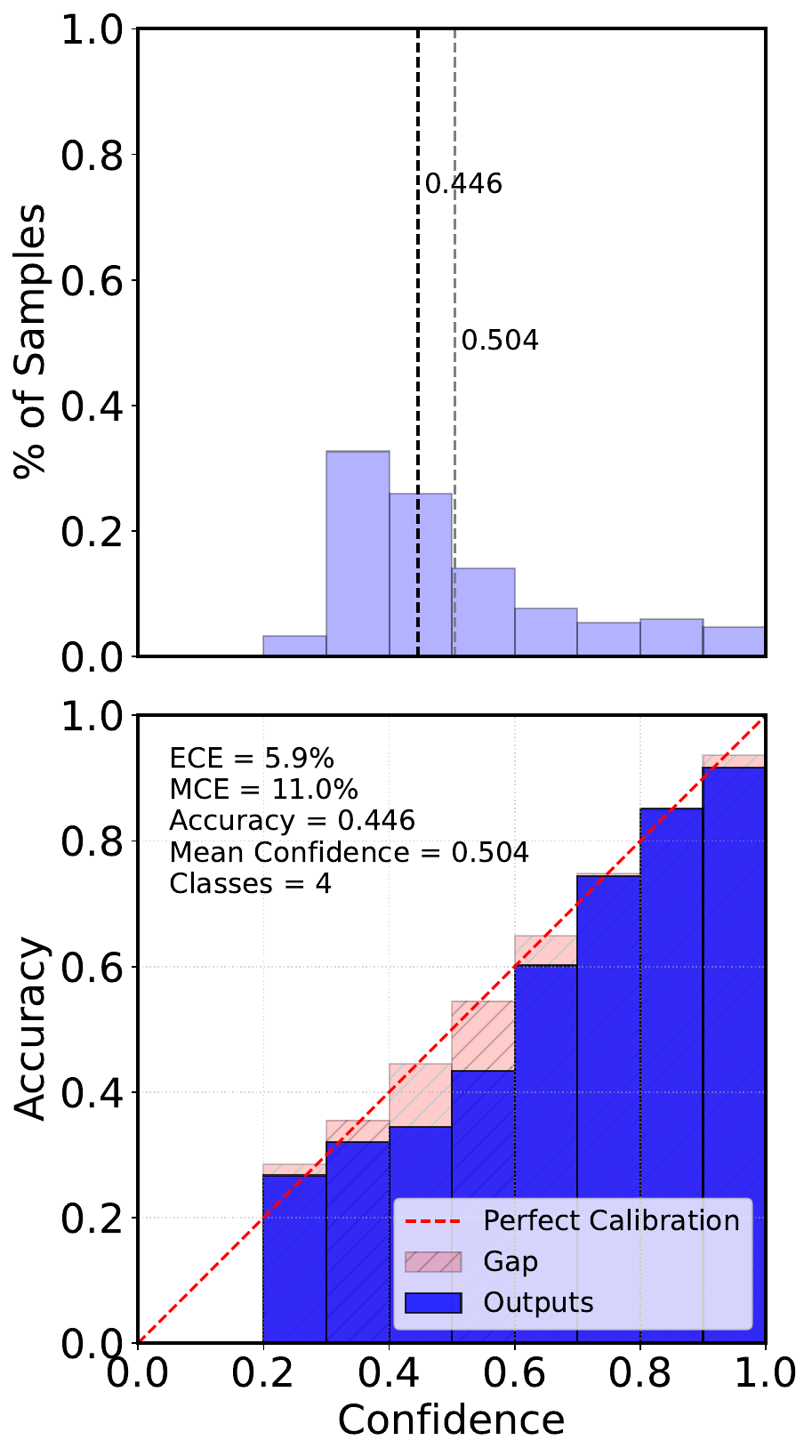}
    \caption{Layer 31}
\end{subfigure}

\caption[Reliability Diagram for Phi-2 on MMLU-STEM]{Performance (Accuracy) and calibration (Reliability diagrams) across the later layers of the phi-2 model on the MMLU STEM dataset. Each subfigure shows the reliability diagram and accuracy metrics for a different transformer layer (20-31).
The Gap in the reliability diagrams reduces in the later layers, with the dashed dark (Accuracy) and light (Mean Confidence) verticle lines coming close in the last layer, showing the improved model calibration. 
}
\label{fig:all_layers_calibration_relialibility_diagrams}
\end{figure*}

\begin{figure*}[t]
\centering
\captionsetup[subfigure]{labelformat=parens}

\newcommand{\scalefactor}{0.45} 

\begin{subfigure}[b]{\scalefactor\linewidth}
    \includegraphics[width=\linewidth]{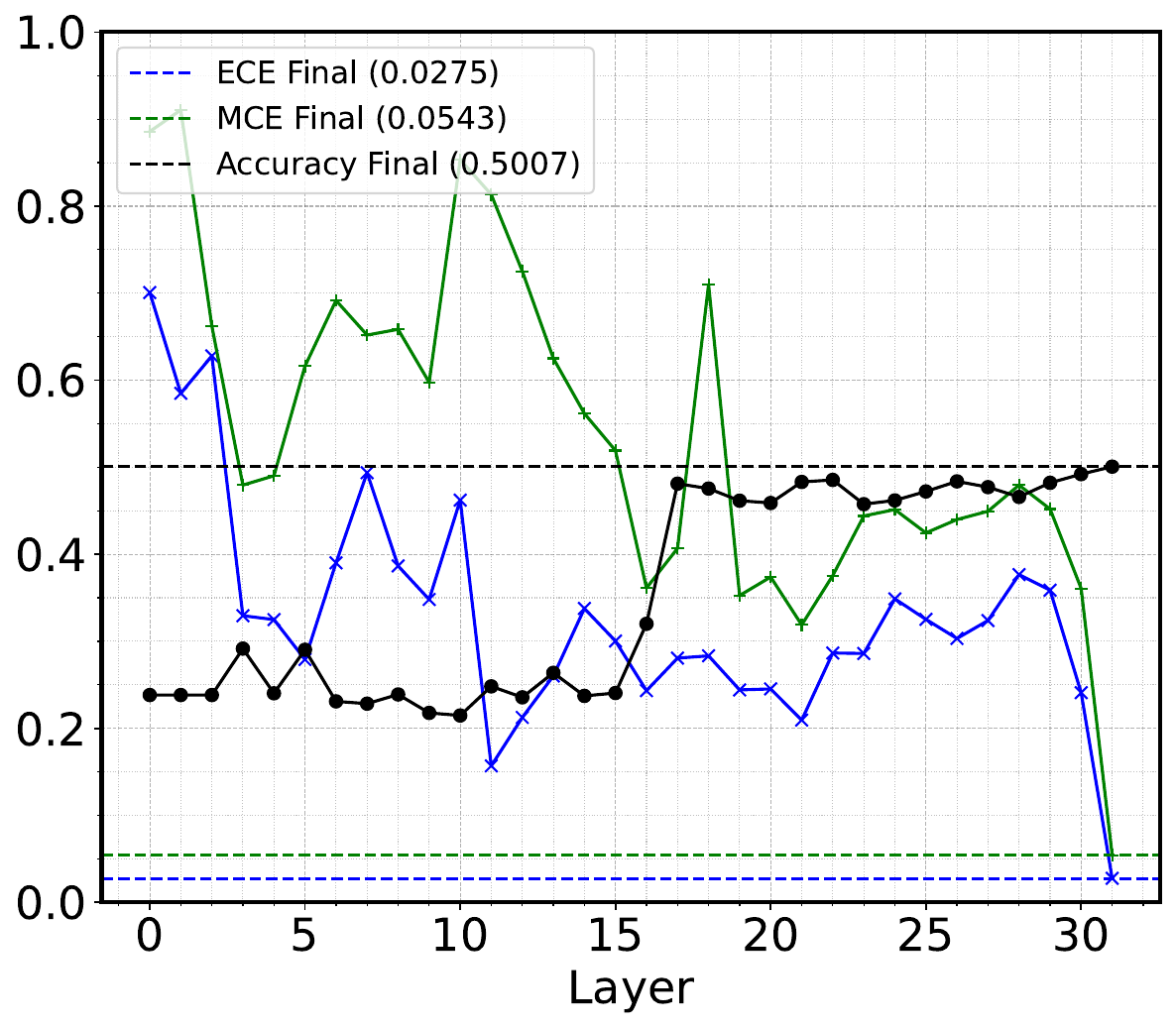}
    \caption{MMLU STEM}
\end{subfigure}\hfill
\begin{subfigure}[b]{\scalefactor\linewidth}
    \includegraphics[width=\linewidth]{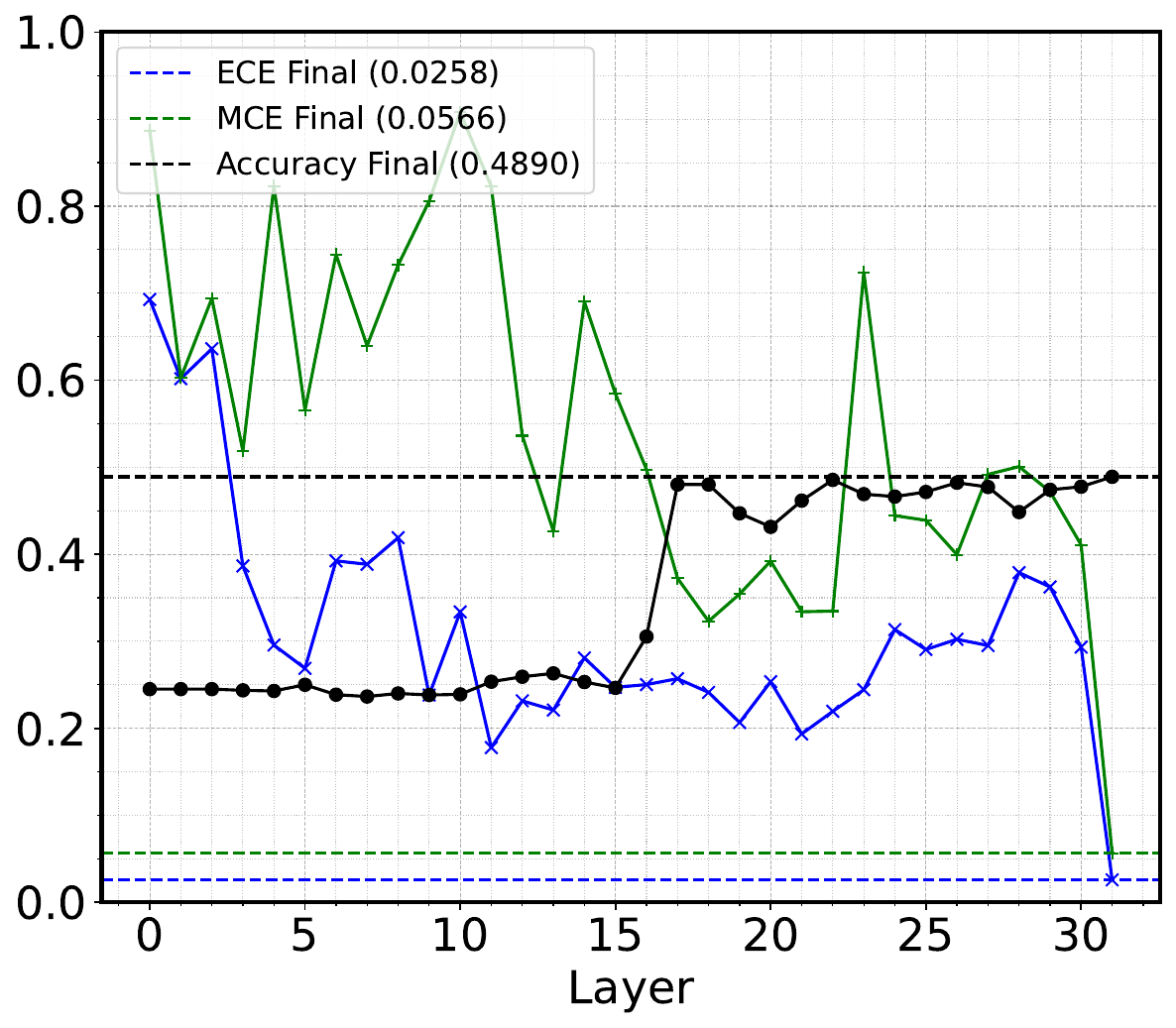}
    \caption{MMLU Humanities}
\end{subfigure}

\begin{subfigure}[b]{\scalefactor\linewidth}
    \includegraphics[width=\linewidth]{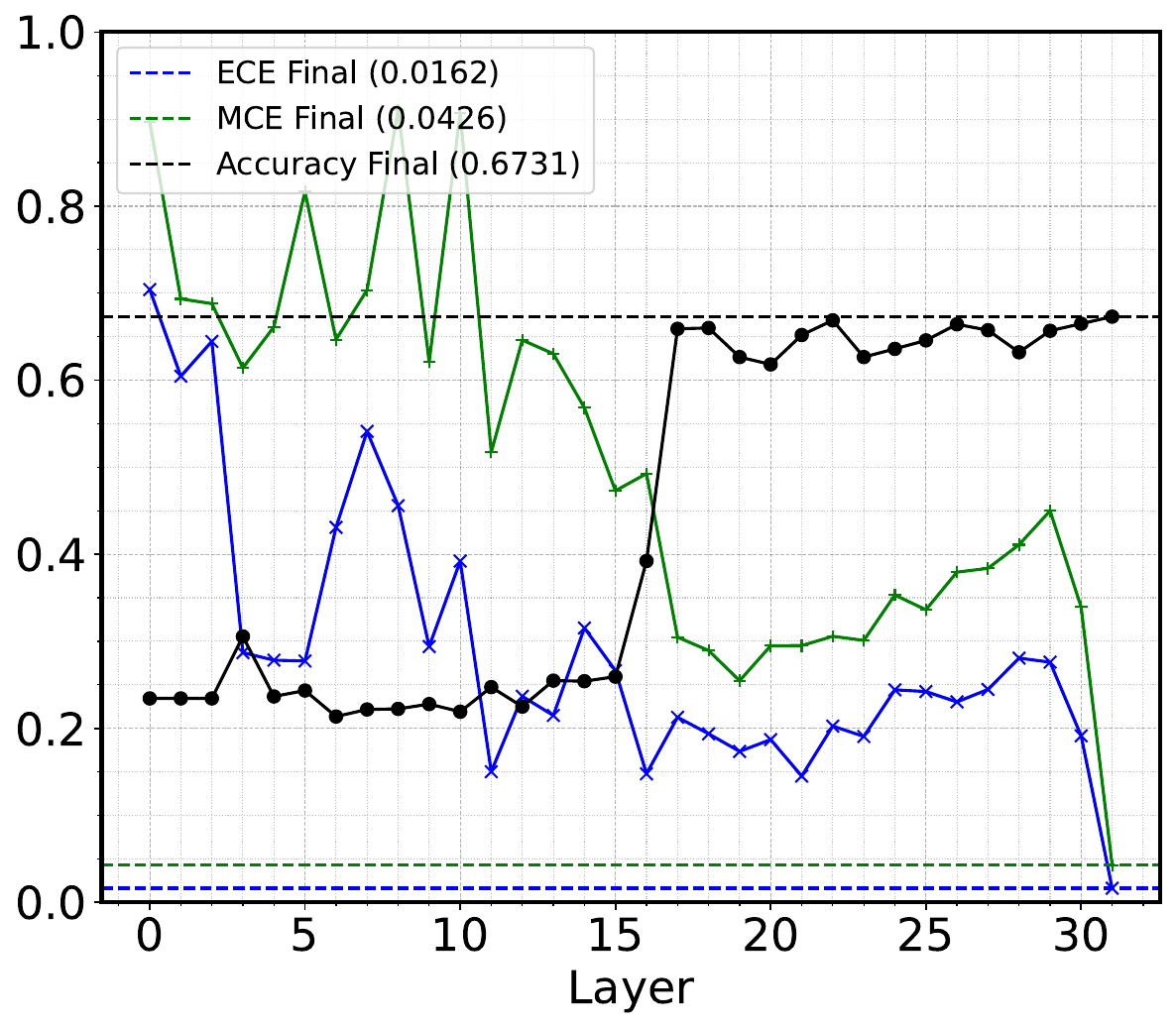}
    \caption{MMLU Social Science}
\end{subfigure}
\hfill
\begin{subfigure}[b]{\scalefactor\linewidth}
    \includegraphics[width=\linewidth]{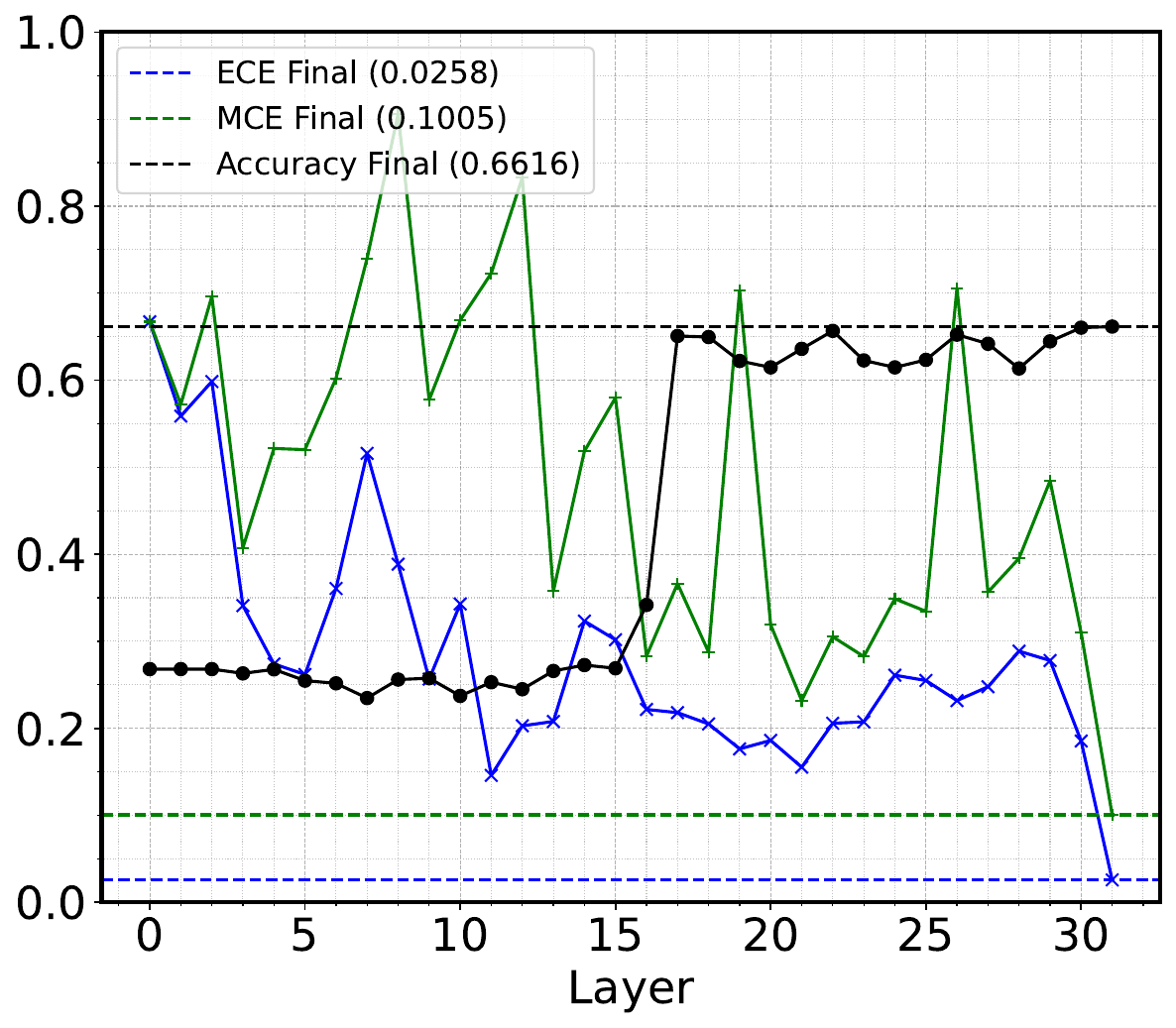}
    \caption{MMLU Others}
\end{subfigure}

\caption[Layer-wise Calibration of Llama-3-8B on MMLU splits]{The figure shows performance (Accuracy) along with model calibration scores (ECE and MCE) of the Llama-3-8B model on the different datasets. We observe that the model performance starts to rise from layer 15 and saturates at layer 17, with minor changes in the 17-31 layers. However, the ECE and MCE scores first rise (layers 25-28) and then decline (layers 28-31), highlighting the model calibration changing in the later layers, with meager changes in the model performance. This denotes that the residual stream in the later layers is affected/modified in such a way that modulates the model calibration with no/minor change in the model performance (black line). The upper/later layers showing the presence of \textbf{\textit{calibration correction phase}}.
}
\label{fig:all_layers_calibration_Llama-3-8B_mmlu_all}
\end{figure*}

\begin{figure*}[t]
\centering
\captionsetup[subfigure]{labelformat=parens}

\newcommand{\scalefactor}{0.45} 

\begin{subfigure}[b]{\scalefactor\linewidth}
    \includegraphics[width=\linewidth]{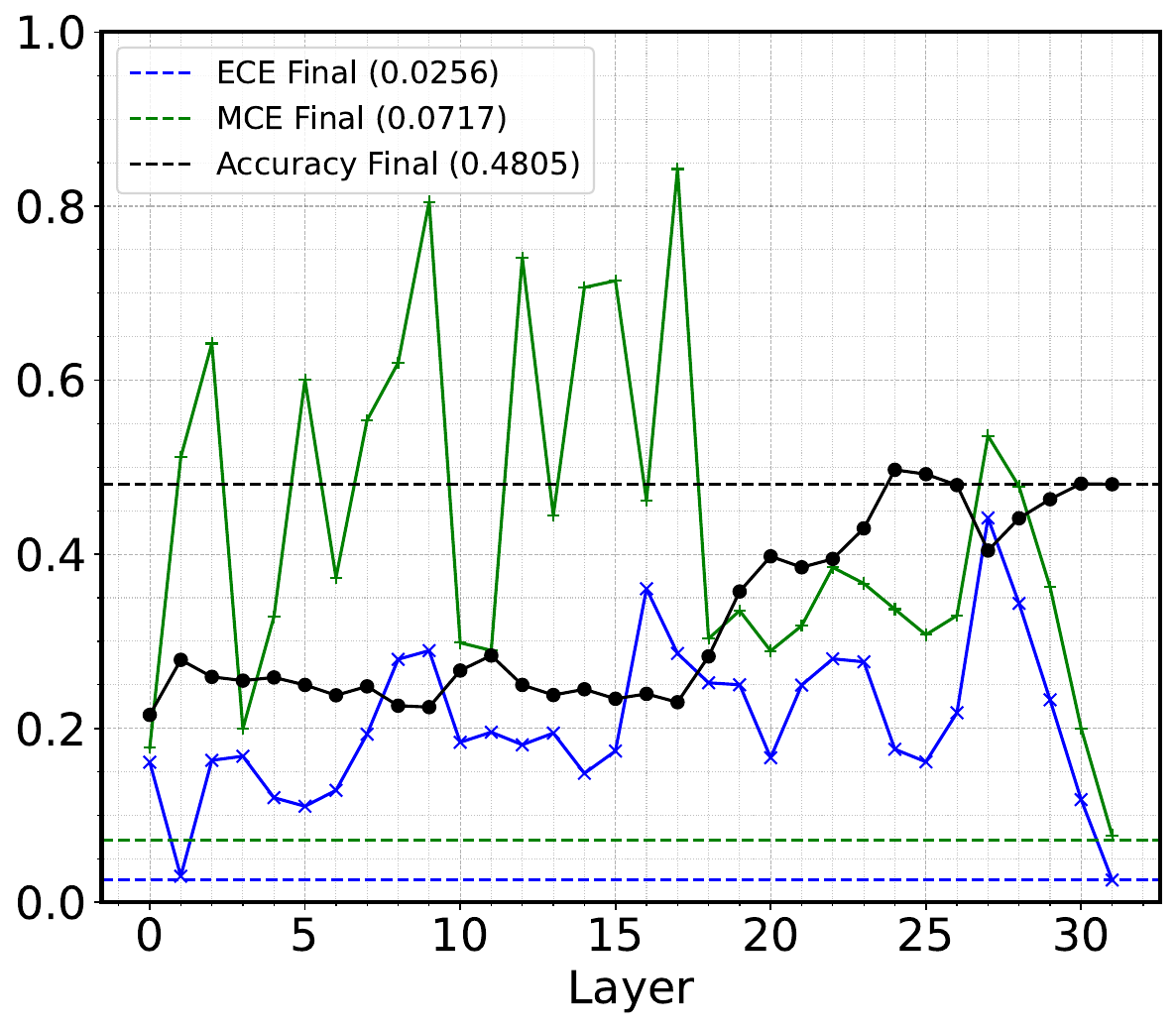}
    \caption{MMLU STEM}
\end{subfigure}\hfill
\begin{subfigure}[b]{\scalefactor\linewidth}
    \includegraphics[width=\linewidth]{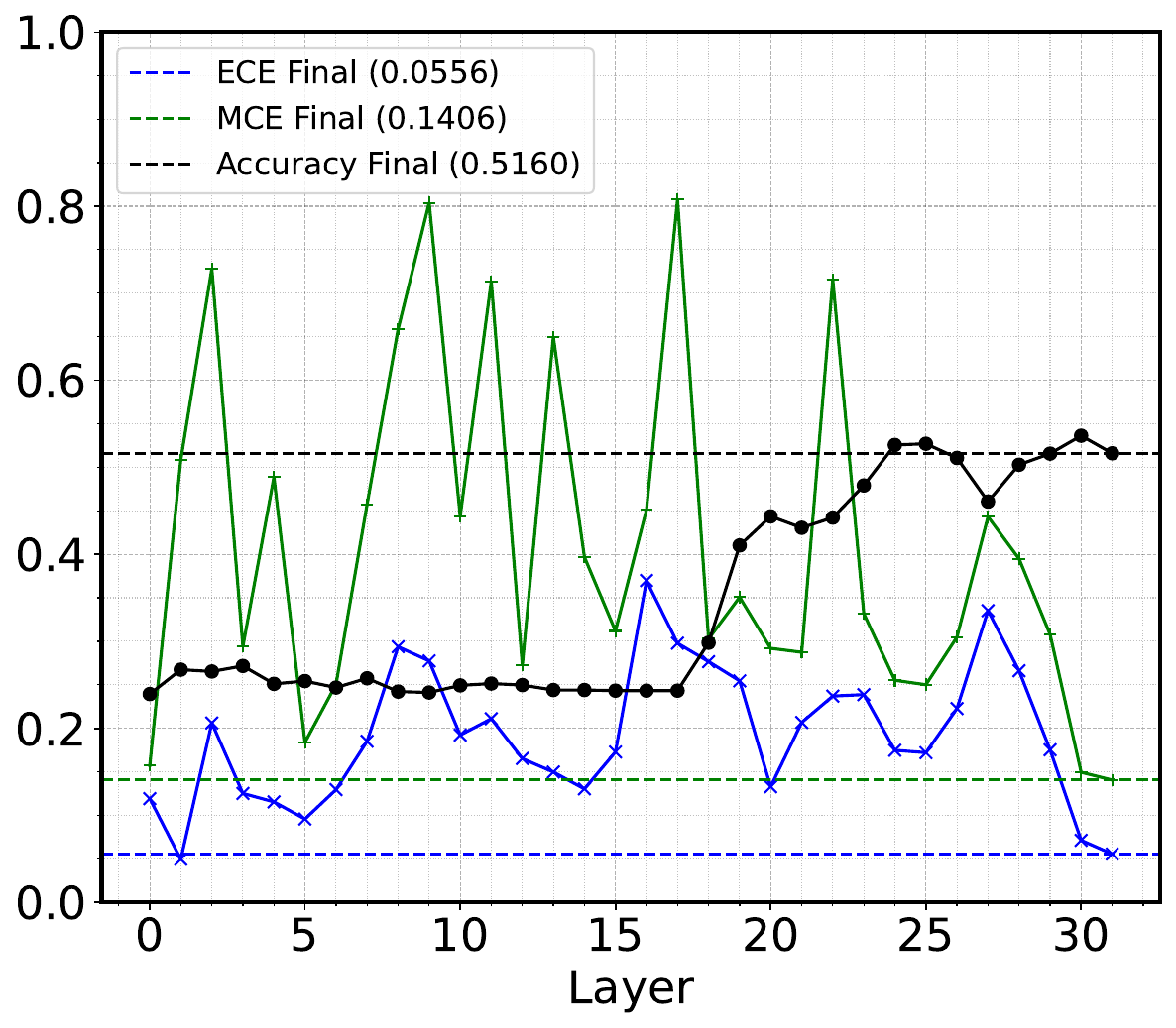}
    \caption{MMLU Humanities}
\end{subfigure}

\begin{subfigure}[b]{\scalefactor\linewidth}
    \includegraphics[width=\linewidth]{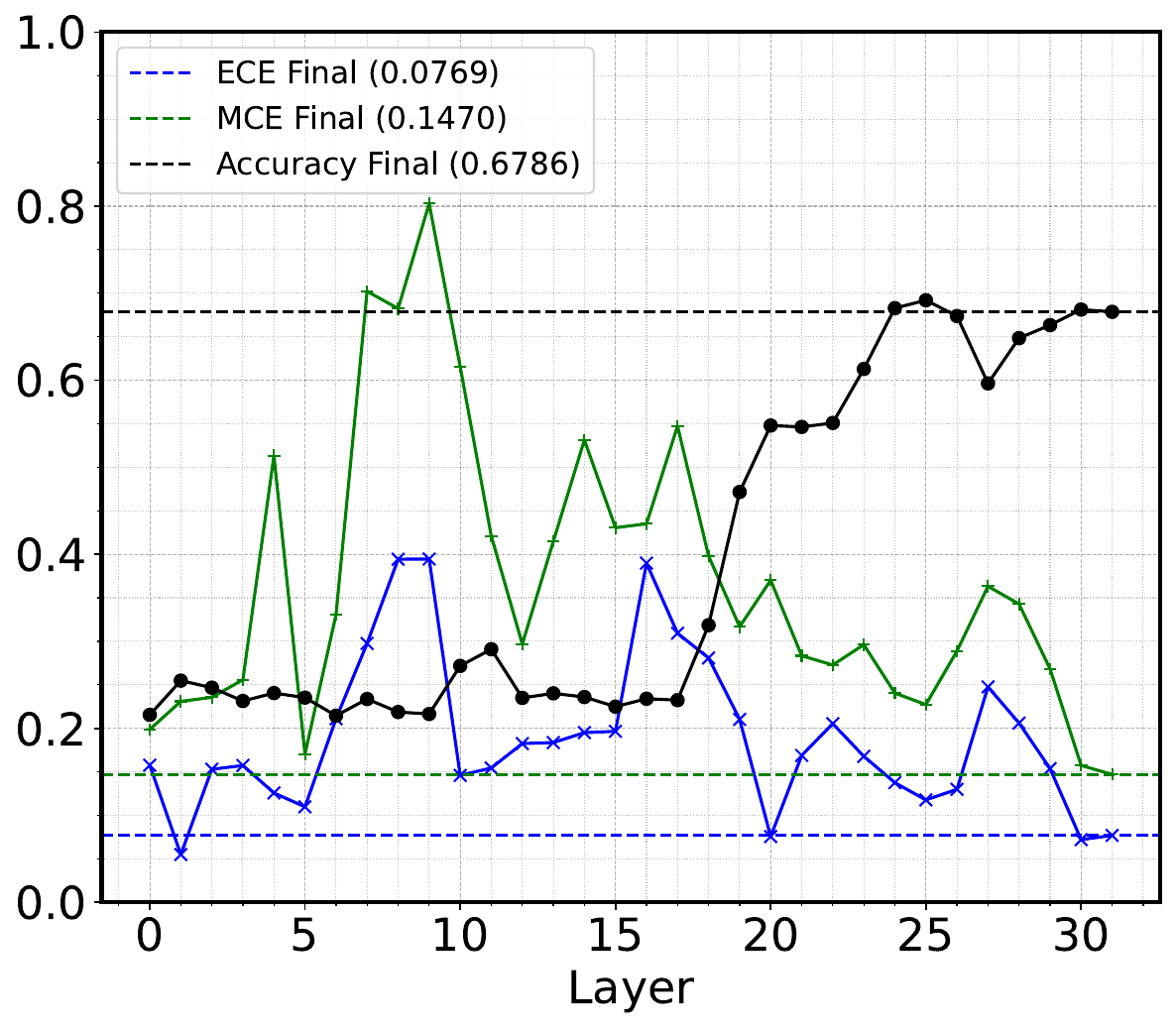}
    \caption{MMLU Social Science}
\end{subfigure}
\hfill
\begin{subfigure}[b]{\scalefactor\linewidth}
    \includegraphics[width=\linewidth]{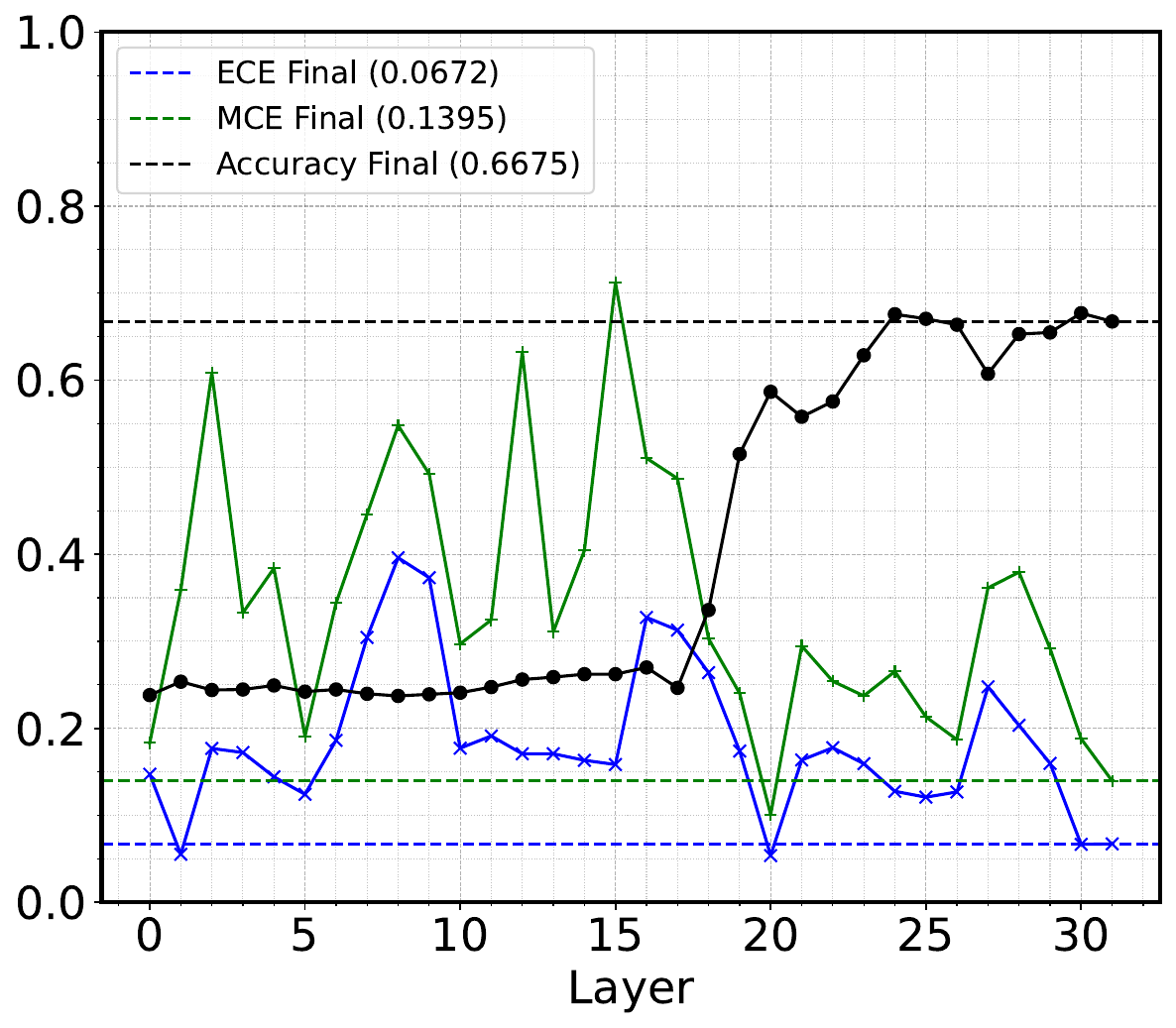}
    \caption{MMLU Others}
\end{subfigure}

\caption[Layer-wise Calibration of Mistral-7B on MMLU splits]{The figure shows performance (Accuracy) along with model calibration scores (ECE and MCE) of the Mistral-7B model on the different datasets. We observe that the model performance starts to rise from layer 16/17 and saturates at layer 24, with minor changes in the 24-31 layers. However, the ECE and MCE scores first rise (layers 24-28) and then decline (layers 28-31), highlighting the model calibration changing in the later layers, with meager changes in the model performance. This denotes that the residual stream in the later layers is affected/modified in such a way that modulates the model calibration with no/minor change in the model performance (black line).
}
\label{fig:all_layers_calibration_mistralai_Mistral-7B-v0.1}
\end{figure*}

\begin{figure*}[t]
\centering
\captionsetup[subfigure]{labelformat=parens}

\newcommand{\scalefactor}{0.45} 

\begin{subfigure}[b]{\scalefactor\linewidth}
    \includegraphics[width=\linewidth]{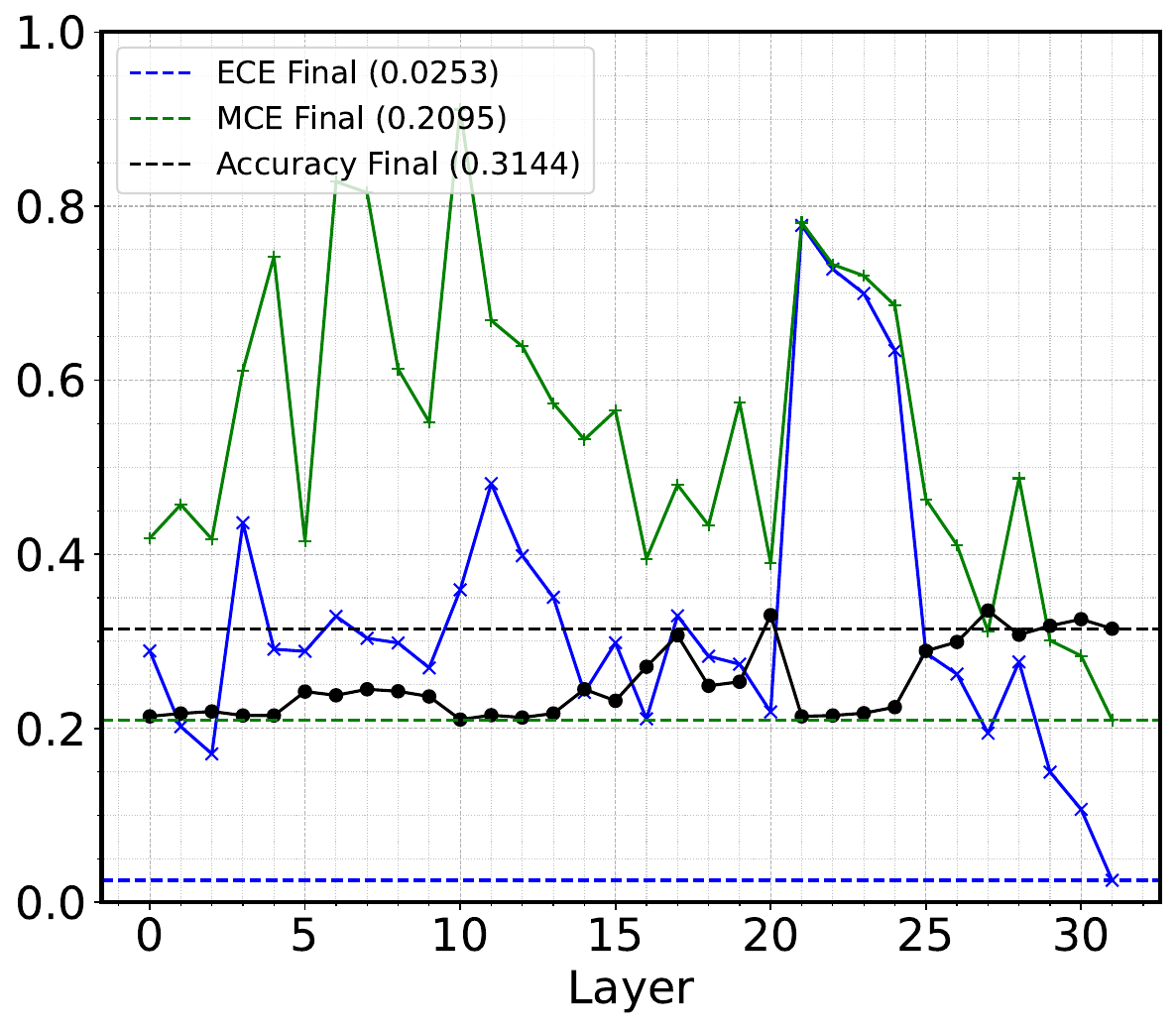}
    \caption{MMLU STEM}
\end{subfigure}\hfill
\begin{subfigure}[b]{\scalefactor\linewidth}
    \includegraphics[width=\linewidth]{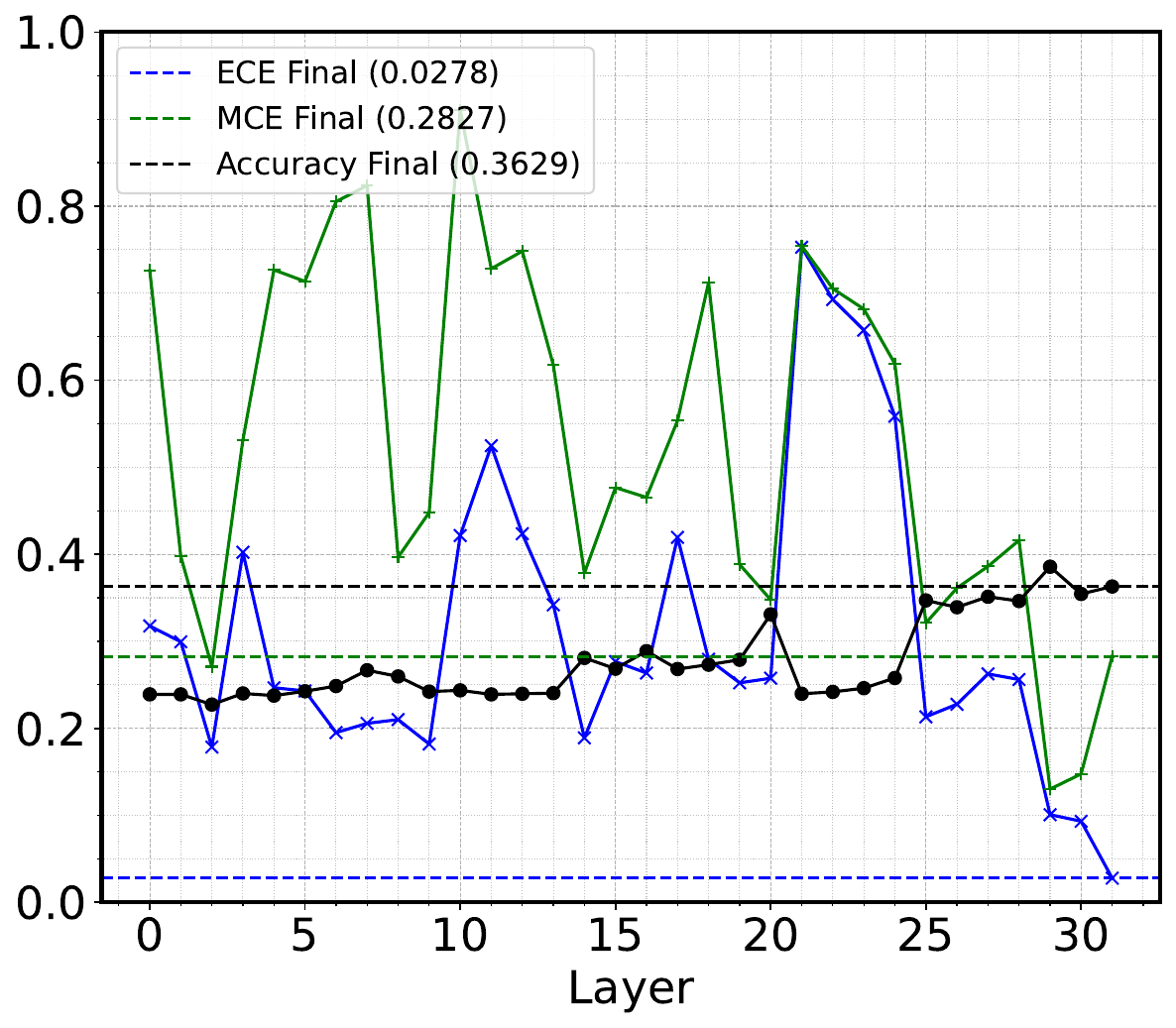}
    \caption{MMLU Humanities}
\end{subfigure}

\begin{subfigure}[b]{\scalefactor\linewidth}
    \includegraphics[width=\linewidth]{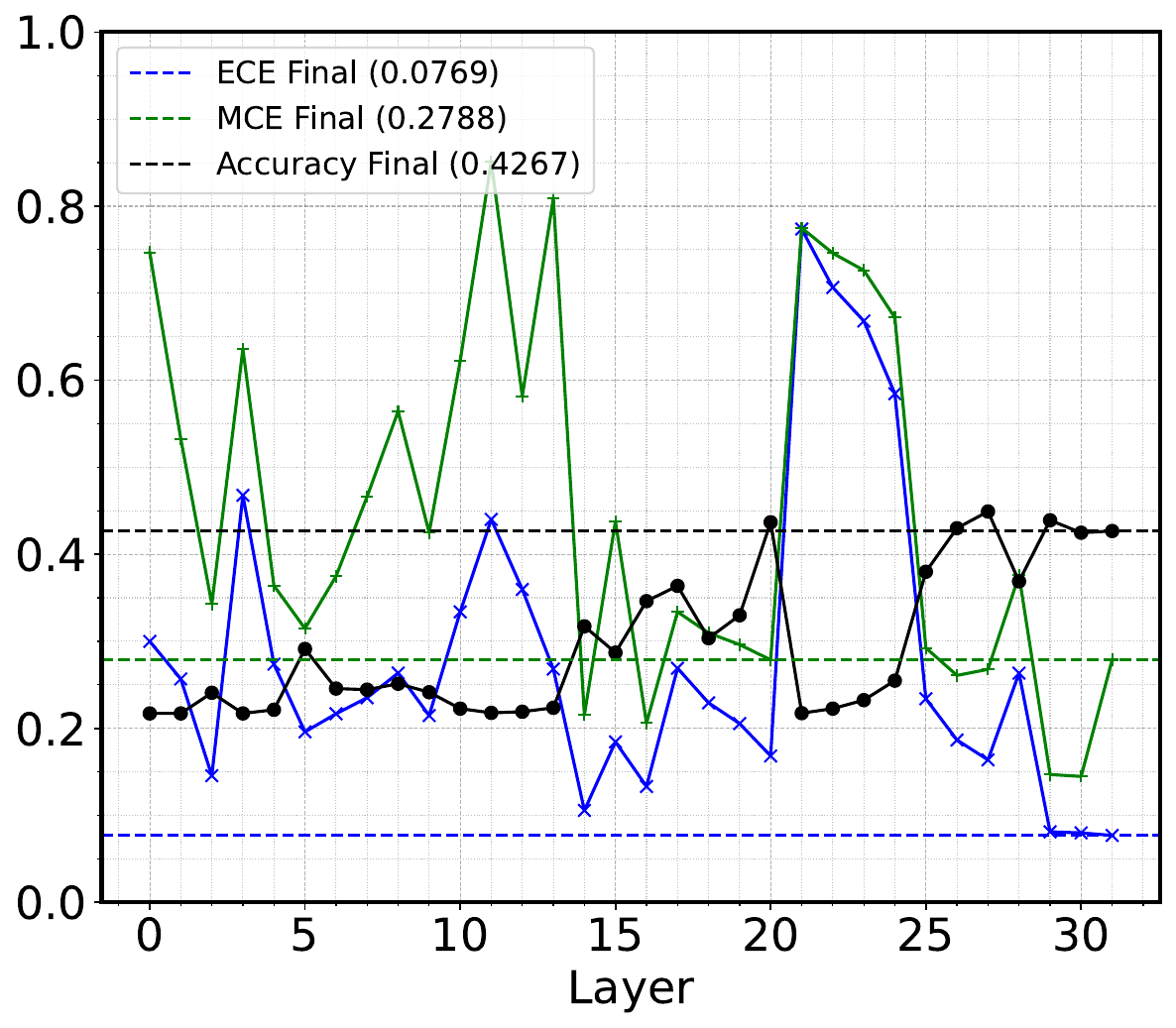}
    \caption{MMLU Social Science}
\end{subfigure}
\hfill
\begin{subfigure}[b]{\scalefactor\linewidth}
    \includegraphics[width=\linewidth]{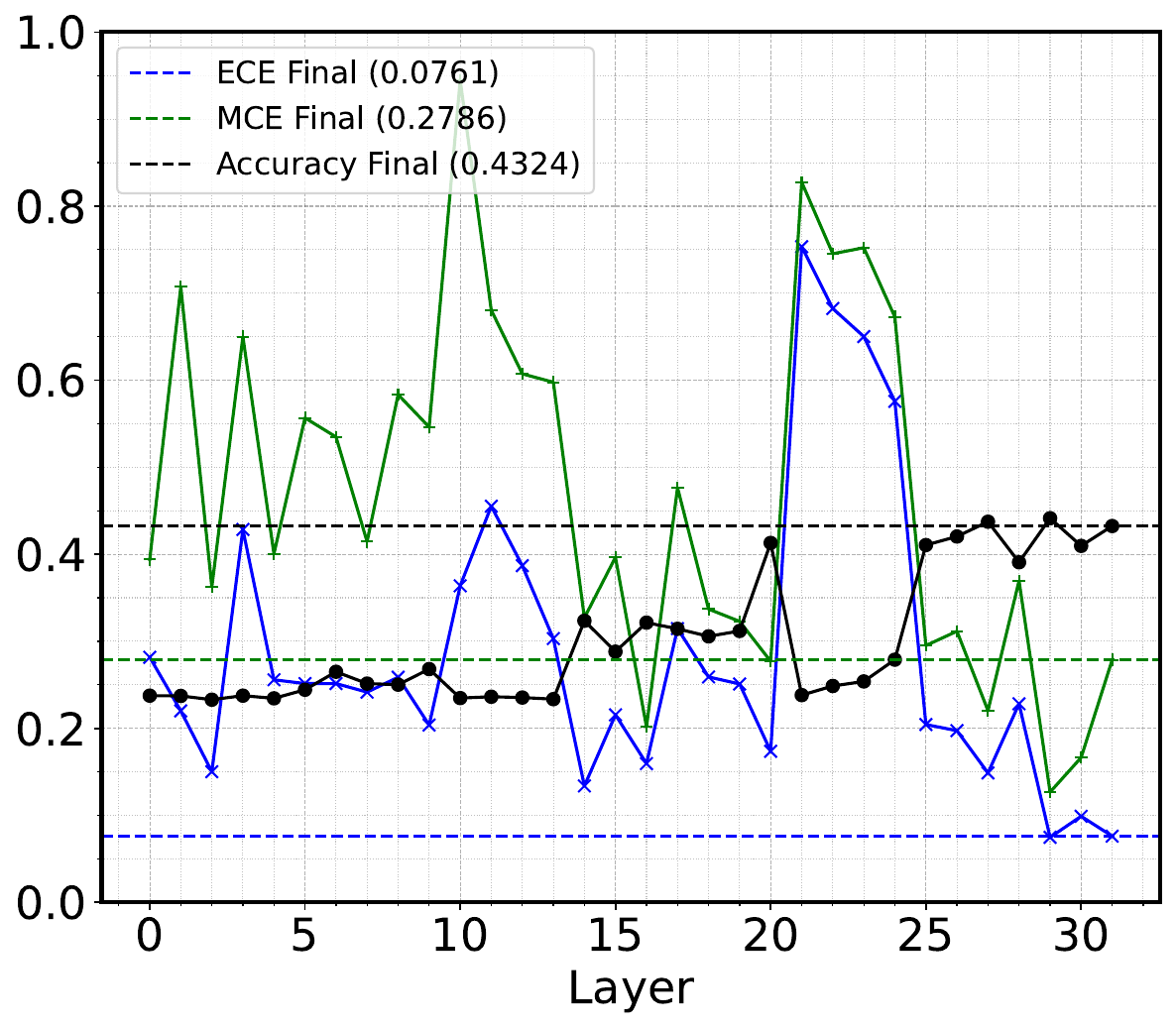}
    \caption{MMLU Others}
\end{subfigure}

\caption[Layer-wise Calibration of Llama-2-7B on MMLU splits]{The figure shows performance (Accuracy) along with model calibration scores (ECE and MCE) of the Llama-2-7B model on the different datasets. We observe that the model performance starts to rise from layer 24 and saturates at layer 27, with minor changes in the 27-31 layers. However, the ECE and MCE scores first rise (layers 25-27) and then decline (layers 28-31), highlighting the model calibration changing in the later layers, with meager changes in the model performance. This denotes that the residual stream in the later layers is affected/modified in such a way that modulates the model calibration with no/minor change in the model performance (black line).
}
\label{fig:all_layers_calibration_meta-llama_Llama-2-7b-hf_mmlu_all}
\end{figure*}

To visualize calibration quality across model depth, we plot layerwise reliability diagrams for Phi-2, Mistral-7B, and LLaMA-2-7B on representative tasks (refer to our codebase for all the reliability diagrams). In Phi-2, we observe a clear overconfidence pattern in middle layers (see Figure \ref{fig:all_layers_calibration_relialibility_diagrams}), followed by a significant correction in later layers, where predicted confidence aligns more closely with empirical accuracy. (also see ECE/MCE patterns in Mistral-7B Figure \ref{fig:all_layers_calibration_mistralai_Mistral-7B-v0.1}, LLaMA-2-7B Figure \ref{fig:all_layers_calibration_meta-llama_Llama-2-7b-hf_mmlu_all}, LLaMA-3-8B Figure \ref{fig:all_layers_calibration_Llama-3-8B_mmlu_all} Phi-2 Figure \ref{fig:all_layers_calibration_phi-2_mmlu_all}, all showing a common \textbf{\textit{calibration correction phase}} in the later/upper layers of the model where the calibration error first increases and then decreases, keeping the prediction accuracy intact.)



\section{Dataset-Level Calibration Trends}

\begin{figure*}[t]
\centering
\captionsetup[subfigure]{labelformat=parens}

\newcommand{\scalefactor}{0.45} 

\begin{subfigure}[b]{\scalefactor\linewidth}
    \includegraphics[width=\linewidth]{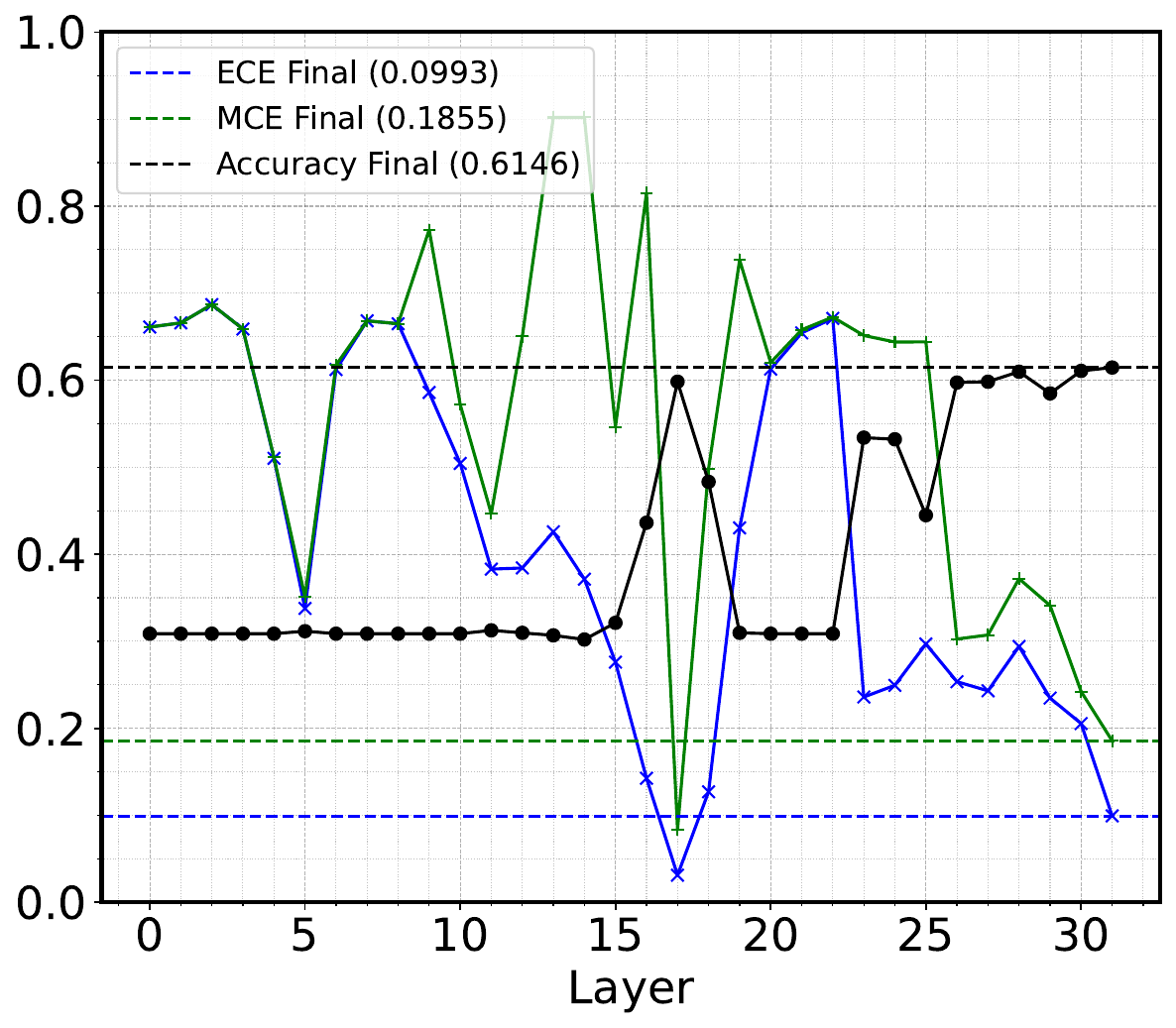}
    \caption{COLA}
\end{subfigure}\hfill
\begin{subfigure}[b]{\scalefactor\linewidth}
    \includegraphics[width=\linewidth]{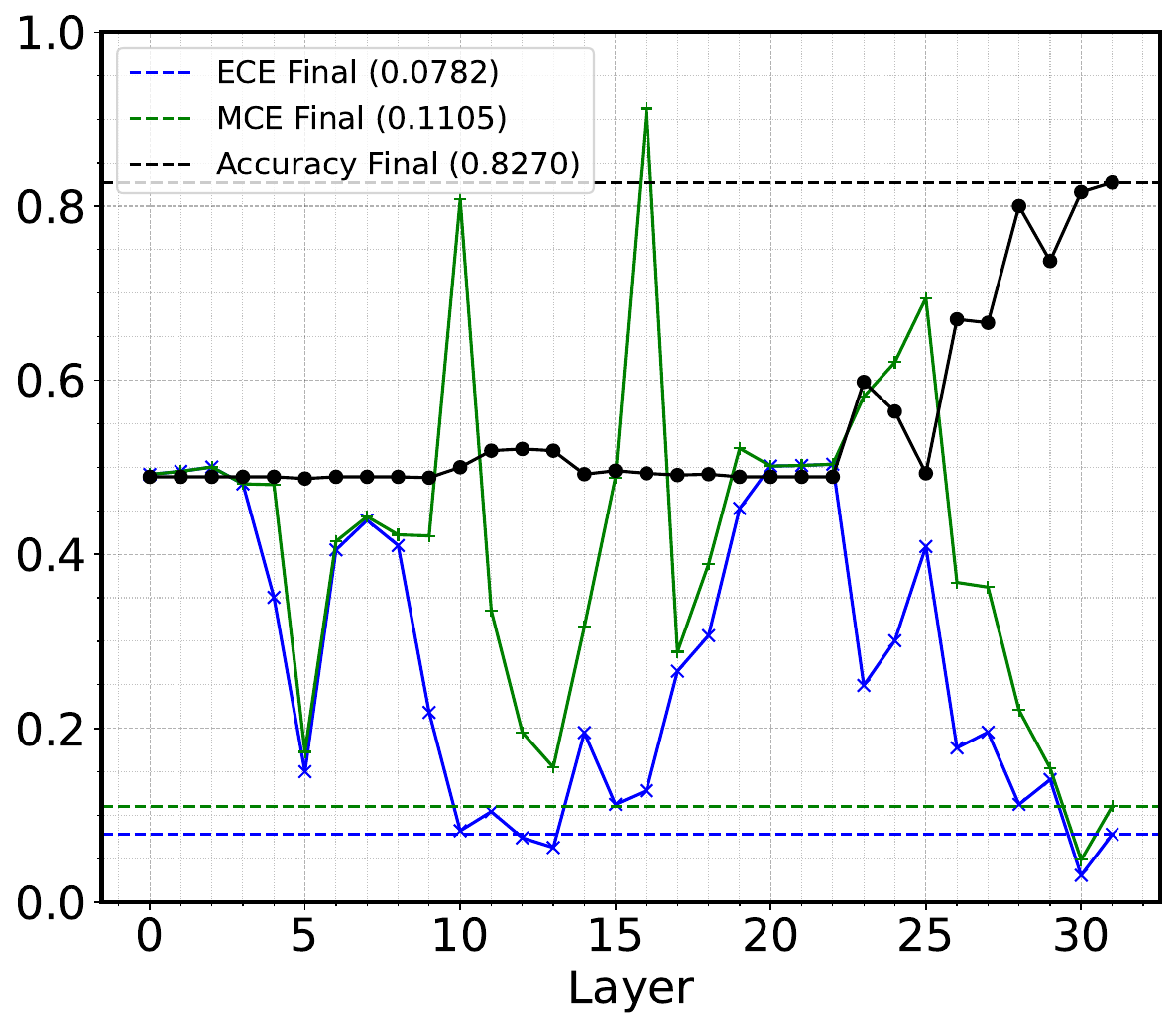}
    \caption{COPA}
\end{subfigure}

\begin{subfigure}[b]{\scalefactor\linewidth}
    \includegraphics[width=\linewidth]{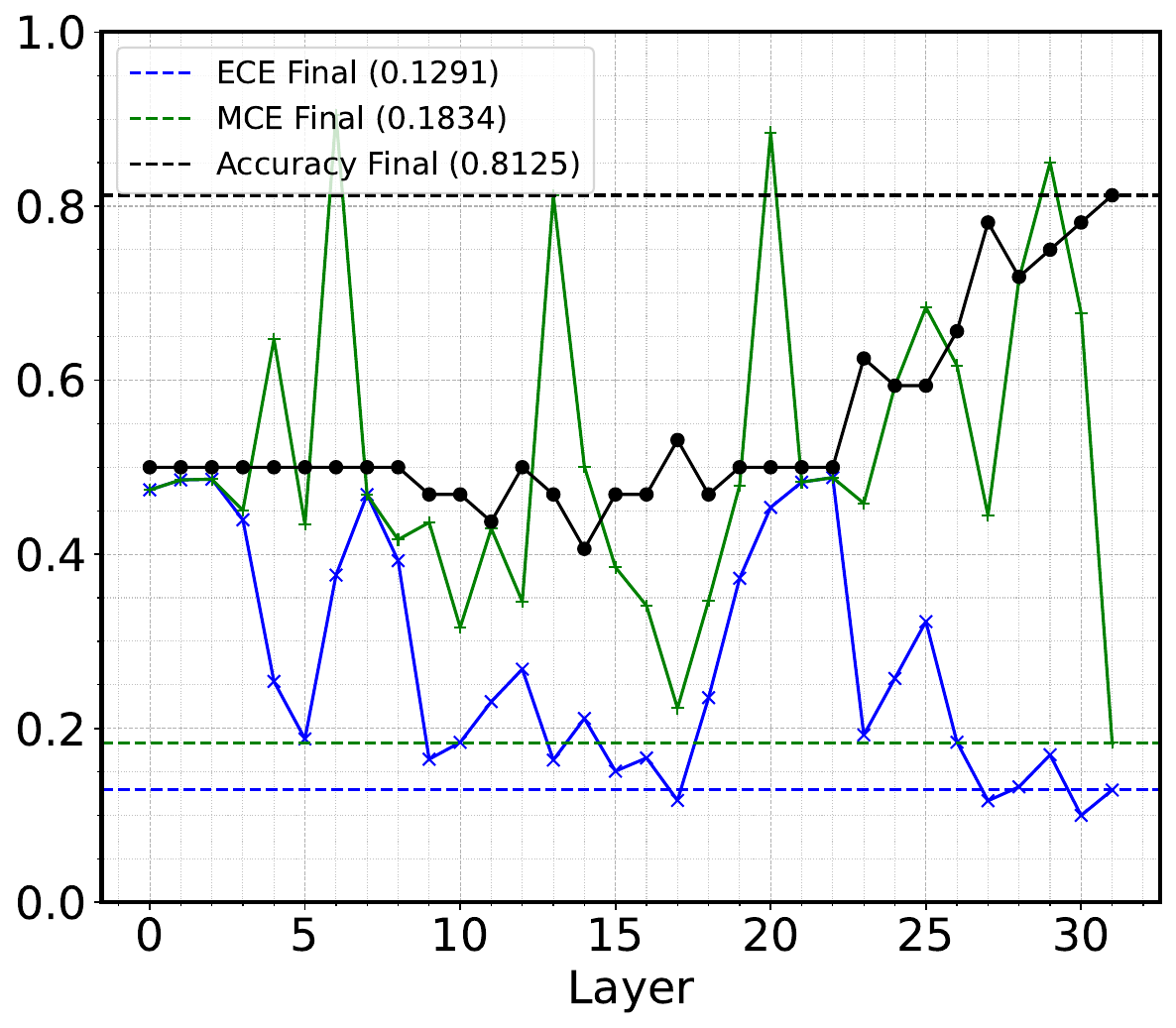}
    \caption{Rotten Tomatoes}
\end{subfigure}
\hfill
\begin{subfigure}[b]{\scalefactor\linewidth}
    \includegraphics[width=\linewidth]{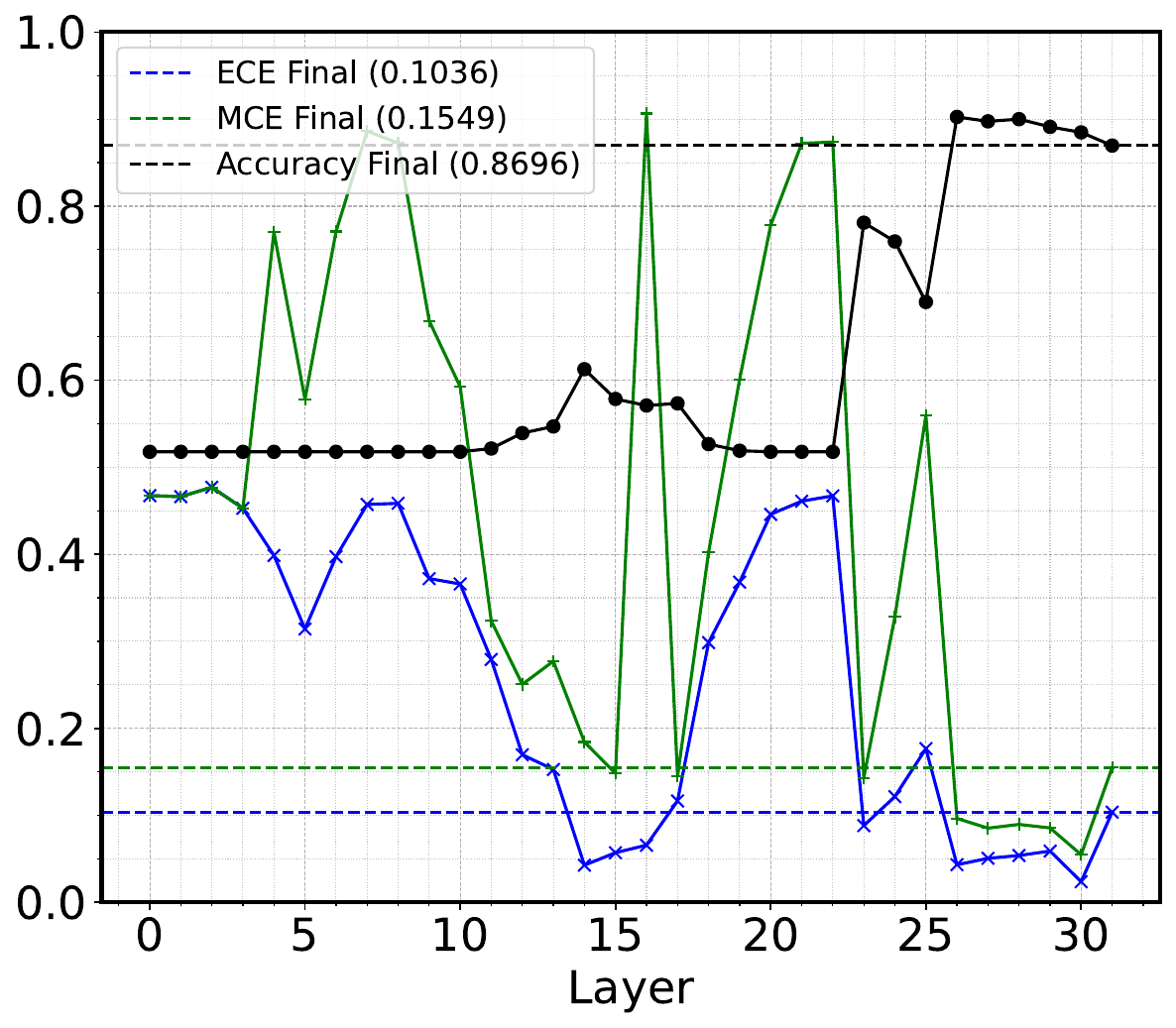}
    \caption{TruthfulQA}
\end{subfigure}

\caption[Layer-wise Calibration of Phi-2 on Other Datasets]{The figure shows performance (Accuracy) along with model calibration scores (ECE and MCE) of the Phi-2 model on the different datasets. We observe a different trend when compared to knowledge acquisition datasets like MMLU, where the accuracy shows a sudden shift. In contrast, here the model shows a gradual change in accuracy, where the model performance starts to rise from layer 22 and gradually increases till layer 28, making it difficult to study the calibration correction phase in particular.
}
\label{fig:all_layers_calibration_phi-2_mmlu_all_other_datasets}
\end{figure*}

\begin{figure*}[t]
\centering
\captionsetup[subfigure]{labelformat=parens}

\newcommand{\scalefactor}{0.45} 

\begin{subfigure}[b]{\scalefactor\linewidth}
    \includegraphics[width=\linewidth]{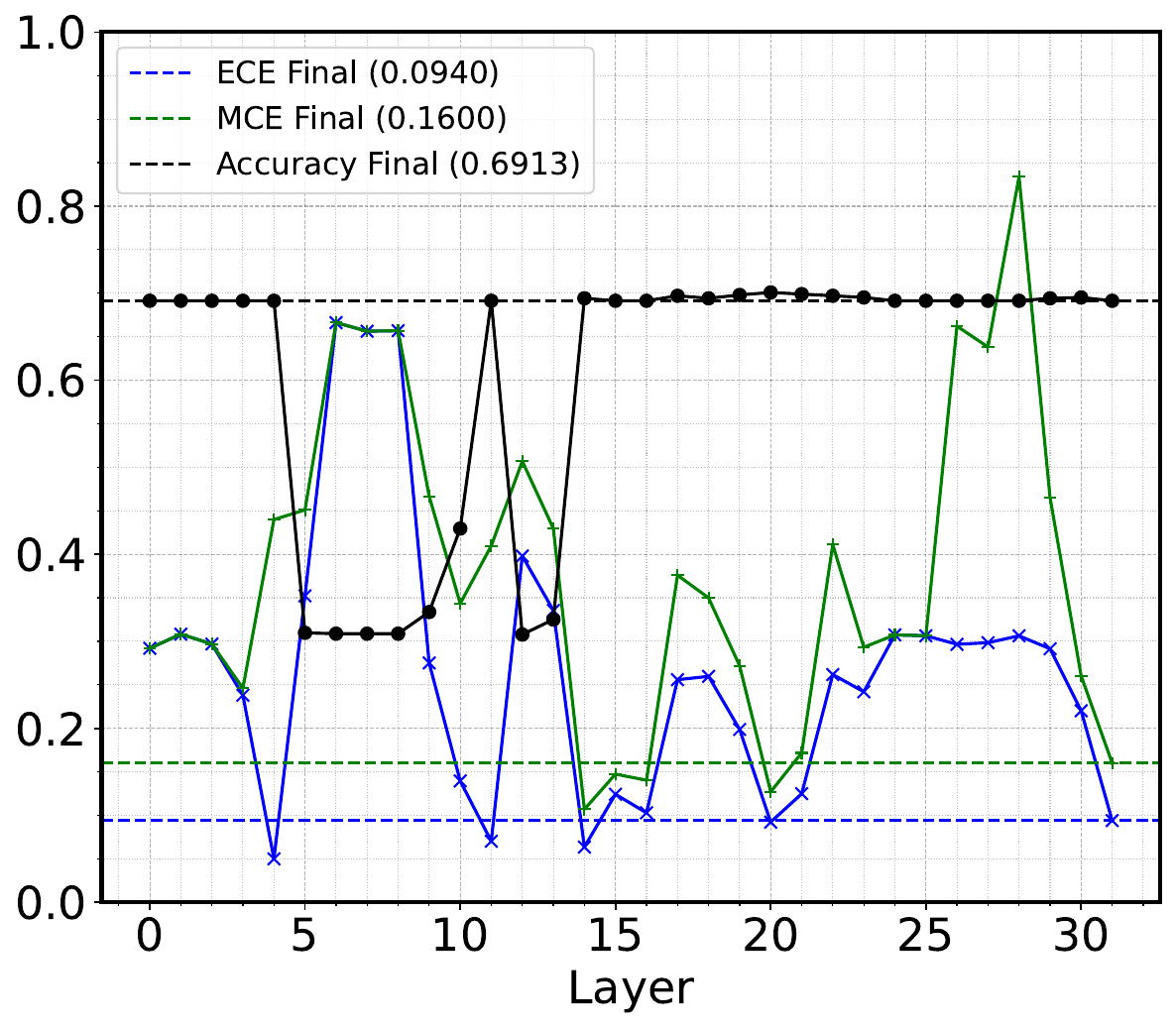}
    \caption{COLA}
\end{subfigure}\hfill
\begin{subfigure}[b]{\scalefactor\linewidth}
    \includegraphics[width=\linewidth]{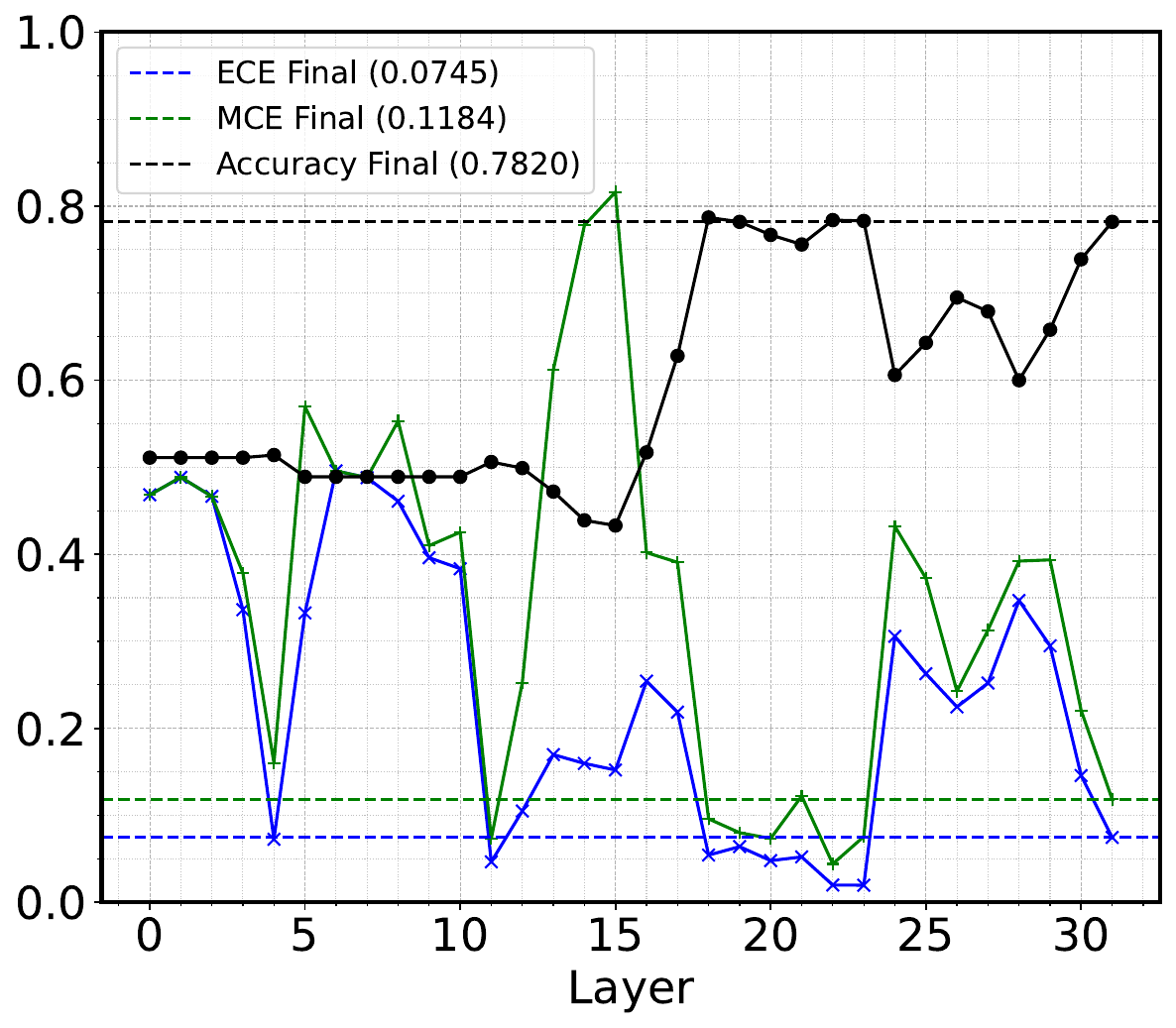}
    \caption{COPA}
\end{subfigure}

\begin{subfigure}[b]{\scalefactor\linewidth}
    \includegraphics[width=\linewidth]{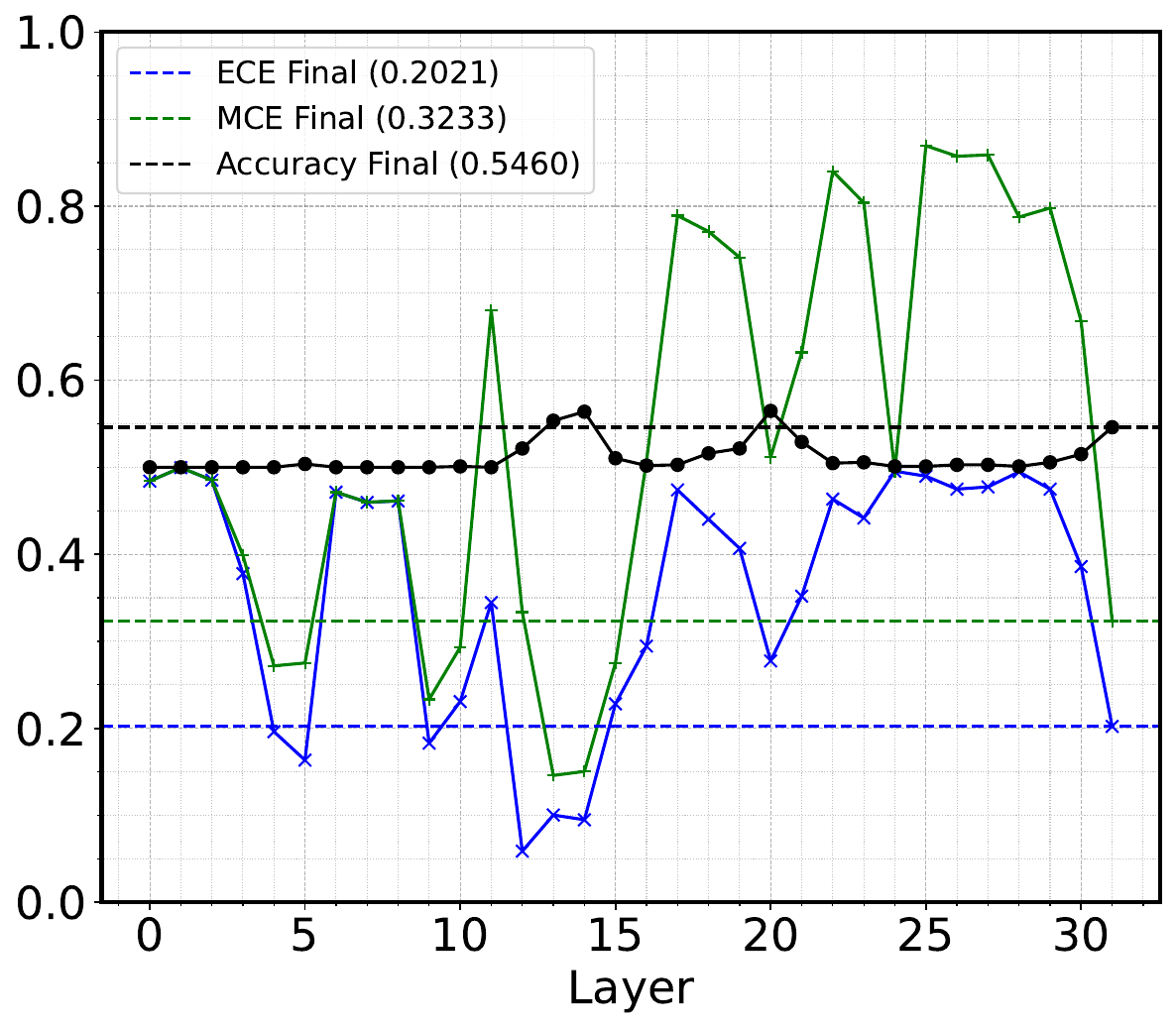}
    \caption{Rotten Tomatoes}
\end{subfigure}
\hfill
\begin{subfigure}[b]{\scalefactor\linewidth}
    \includegraphics[width=\linewidth]{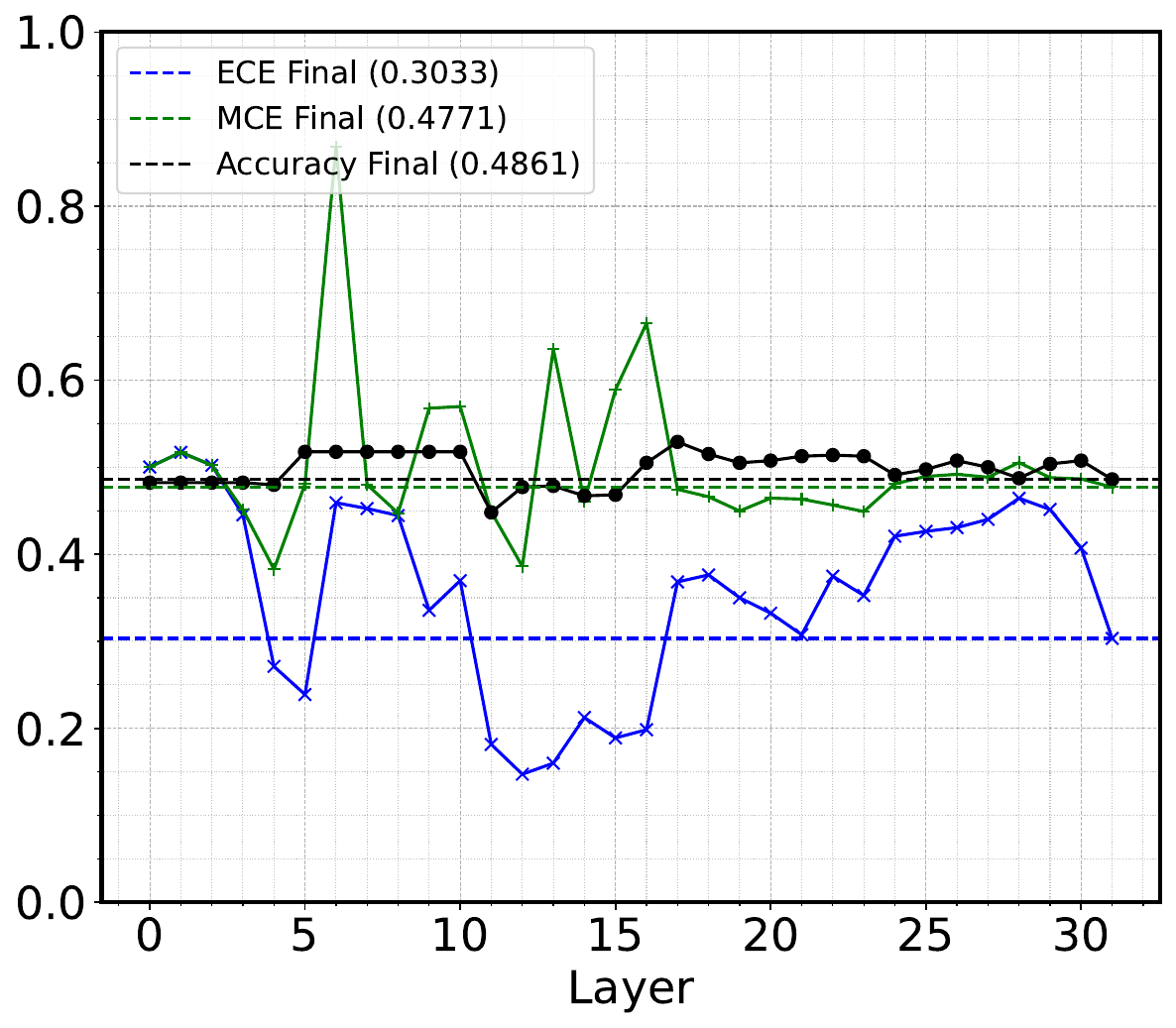}
    \caption{TruthfulQA}
\end{subfigure}

\caption[Layer-wise Calibration of Llama-3-8B on Other Datasets]{The figure shows performance (Accuracy) along with model calibration scores (ECE and MCE) of the Llama-3-8B model on the different datasets. 
We observe that the model performance does not show a significant performance in CoLA (going to near random performance, as per data distribution), with similar meager performance in other datasets like Rotten Tomatoes and TruthfulQA. We only see a performance improvement in the COPA dataset, where again, the calibration correction phase is observed in the later layers. Overall, the poor performance of the model on these datasets makes it difficult to quantify calibration happening across datasets.
}
\label{fig:all_layers_calibration_llama3_mmlu_all_other_datasets}
\end{figure*}

Beyond MMLU, we examine calibration behavior across four additional datasets: CoLA, COPA, Rotten Tomatoes, and TruthfulQA. For each dataset, we compute accuracy, ECE, and MCE across transformer layers. (see Figure \ref{fig:all_layers_calibration_phi-2_mmlu_all_other_datasets} and Figure \ref{fig:all_layers_calibration_llama3_mmlu_all_other_datasets})

In contrast to MMLU, where accuracy stabilizes in later layers, these datasets exhibit gradual performance improvements throughout the model depth. This continuous gain limits our ability to isolate a calibration correction phase, as improvements in calibration may be confounded with accuracy refinement.

On CoLA and COPA, calibration remains noisy across layers, likely due to the small size and binary structure of the tasks.
On Rotten Tomatoes, calibration improves steadily but without a sharp correction pattern.
On TruthfulQA, we observe persistent underconfidence, with predicted probabilities often falling below empirical correctness, especially in earlier layers.

These observations highlight the complexity of measuring calibration when models have not yet saturated in performance.

\section{Interventional Experiments}

To probe the functional role of calibration dynamics, we identify a low-dimensional “calibration direction” in Phi-2’s residual stream using linear probes aligned with calibration error. We perform targeted interventions by adding scaled versions of this direction to intermediate residual representations.

Our experiments reveal that adding this direction at select layers (e.g., layers 22–32, Figure \ref{fig:all_layers_calibration_intervention_main} and Figure \ref{fig:all_layers_calibration_phi-2_mmlu_all_truthfulqa_intervention}) consistently reduces ECE and MCE without degrading accuracy. This confirms that the calibration signal is encoded in the residual stream and can be modulated independently of the model’s final decision.

However, attempts to extract and apply similar directions in Mistral-7B and LLaMA-2-7B were unsuccessful. These models either lacked a distinct calibration direction or showed no calibration improvement upon intervention. This suggests that confidence regulation in Phi-2 is likely facilitated by an architectural or representational property not shared across models.

\begin{figure*}[t]
\centering
\captionsetup[subfigure]{labelformat=parens}

\newcommand{\scalefactor}{0.45} 

\begin{subfigure}[b]{\scalefactor\linewidth}
    \includegraphics[width=\linewidth]{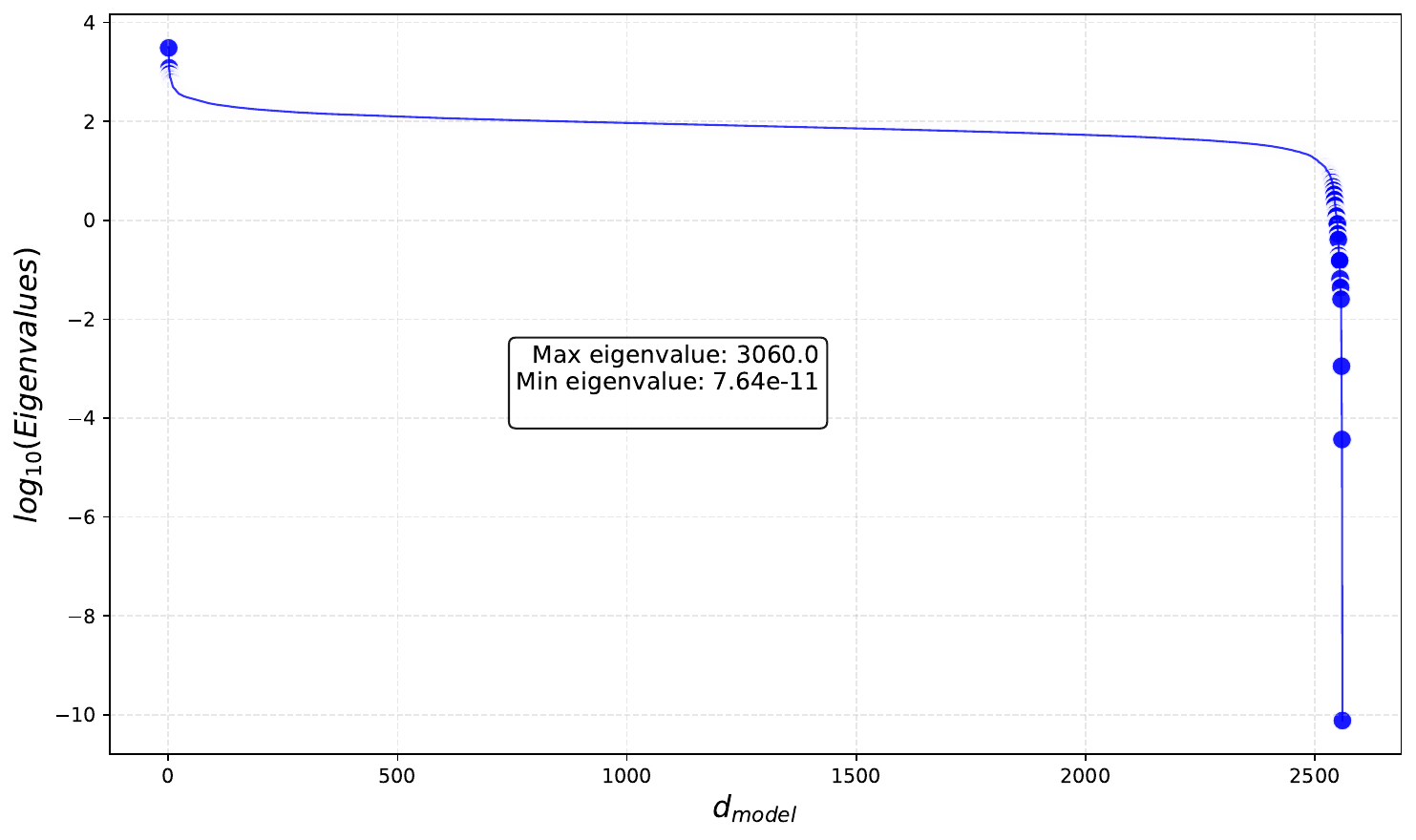}
    \caption{Phi-2}
\end{subfigure}\hfill
\begin{subfigure}[b]{\scalefactor\linewidth}
    \includegraphics[width=\linewidth]{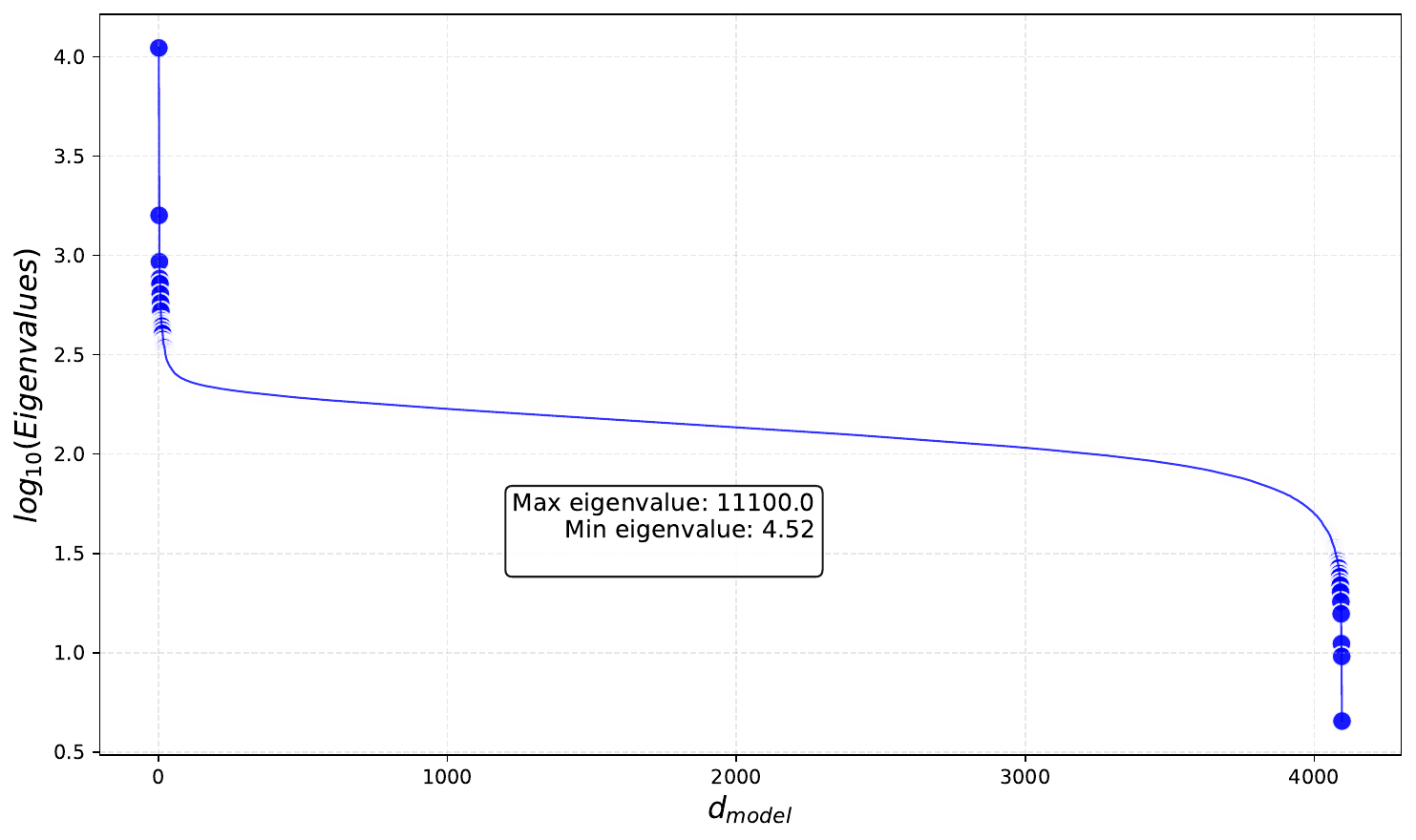}
    \caption{Llama-3-8B}
\end{subfigure}

\caption[Null Space in Umbemedding Matrix for Phi-2 and Llama-3-8B]{The figure shows the eigenvalues (log) of the unembedding matrix of phi-2 and Llama-3-8B. In both models, we observe a sudden decrease in the last 5\% of the singular values, indicating the formation of a null space.
}
\label{fig:eigen_values_unemembedding_matrix}
\end{figure*}

\section{Future Work and Discussion} \label{app:future-work}

In this section, we outline several promising directions to extend and deepen our current findings on calibration mechanisms in LLMs.

\subsection*{Generalizing Calibration Directions}
Our current method for identifying a calibration direction in the residual stream, based on layerwise differences, provides useful insights but may lack generalizability across datasets and models. Future work can explore gradient-based approaches, such as computing the derivative of calibration metrics (e.g., ECE, MCE) with respect to residual activations. This could identify directions more causally linked to calibration. Furthermore, verifying whether such directions align with low-rank or null-space structures in the unembedding matrix (as suggested by \citealp{cancedda2024spectralfiltersdarksignals}) could provide a more principled mechanistic explanation.

\subsection*{Model Internals and Interpretability}
The calibration direction could also serve as a tool for mechanistic interpretability. Specifically, it would be valuable to identify neurons or components (e.g., attention heads or MLP submodules) that significantly contribute to confidence modulation along this direction. Additionally, examining whether calibration behavior propagates through residual stream updates (indirect effects) may reveal compositional mechanisms behind calibration.

\subsection*{Robust Calibration Across Task Types}
Our current findings are most robust in knowledge-centric datasets like MMLU, where prediction accuracy plateaus, enabling clear identification of calibration phases. In contrast, reasoning-focused datasets (e.g., COPA) exhibit gradual accuracy gains across layers, making it harder to isolate calibration-specific dynamics. Understanding how reasoning and knowledge acquisition tasks differentially affect confidence modulation is an important avenue for future research.

\subsection*{Calibration in In-Context Learning Settings}
The current study is limited to zero-shot MCQA prompts. Extending this analysis to in-context learning (ICL) settings, such as few-shot or chain-of-thought prompting, may reveal how calibration dynamics change when more contextual supervision is available. However, this may prove more fruitful for reasoning tasks than factual knowledge tasks, where ICL often yields limited gains.

\subsection*{Cross-Model and Dataset Transferability}
\begin{figure*}[t]
\centering
\captionsetup[subfigure]{labelformat=parens}

\newcommand{\scalefactor}{0.45} 

\begin{subfigure}[b]{\scalefactor\linewidth}
    \includegraphics[width=\linewidth]{./images/upload_results_calibration_direction/phi-2/microsoft_phi-2_ece_mce_accuracy_eta_cross_mmlu_STEM_mmlu_humanities.pdf}
    \caption{MMLU STEM}
\end{subfigure}\hfill
\begin{subfigure}[b]{\scalefactor\linewidth}
    \includegraphics[width=\linewidth]{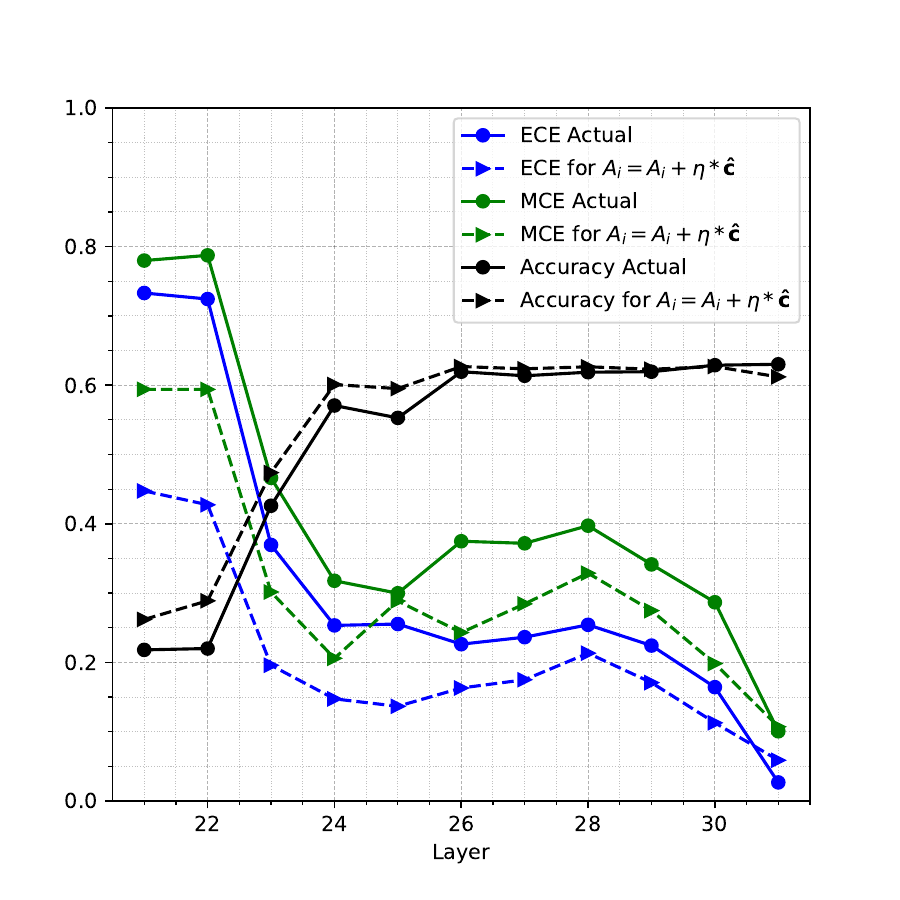}
    \caption{MMLU Social Science}
\end{subfigure}

\begin{subfigure}[b]{\scalefactor\linewidth}
    \includegraphics[width=\linewidth]{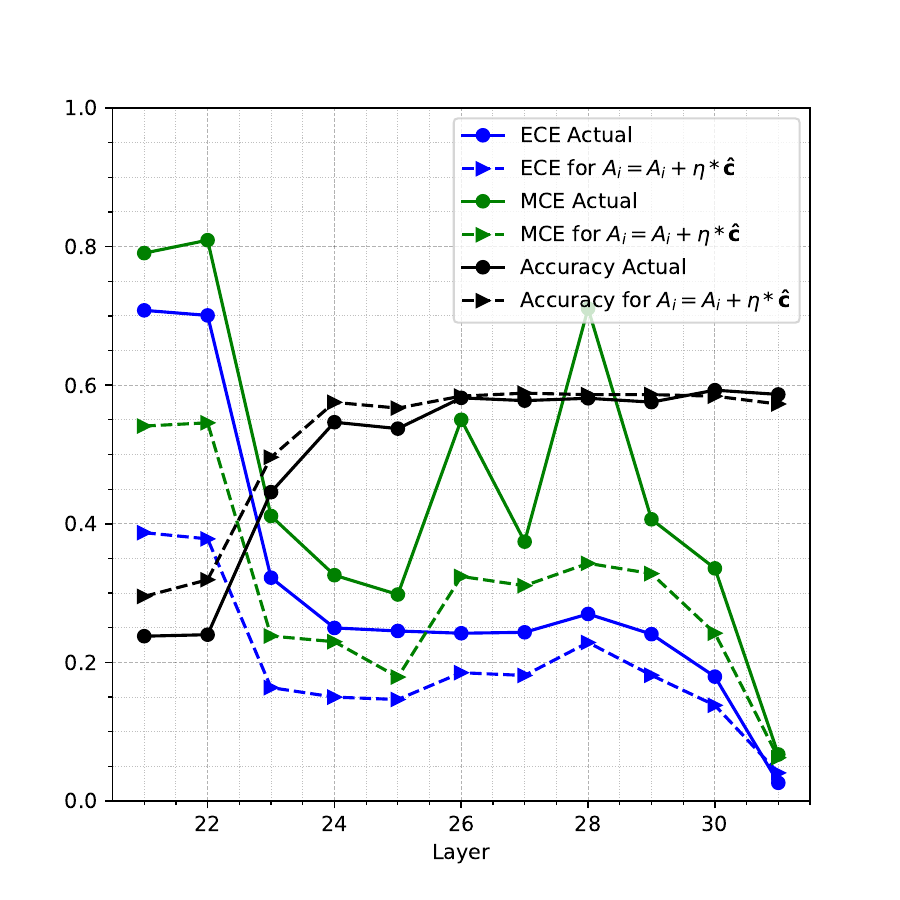}
    \caption{MMLU Others}
\end{subfigure}
\hfill
\begin{subfigure}[b]{\scalefactor\linewidth}
    \includegraphics[width=\linewidth]{./images/upload_results_calibration_direction/phi-2/microsoft_phi-2_ece_mce_accuracy_eta_cross_truthfulqa_mmlu_humanities.pdf}
    \caption{TruthfulQA}
\end{subfigure}

\caption[Calibration Direction, Layer-wise Calibration of Phi-2, generalizing across multiple datasets]{The figure shows performance (Accuracy) along with model calibration scores (ECE and MCE) of the Phi-2 model on the different datasets when the found calibration direction is added to the residual stream. Note that though the calibration direction was found using MMLU Humanities, the found calibration direction generalizes across multiple datasets, including the TruthfulQA, pointing towards a common direction existing in Phi-2 architecture.
}
\label{fig:all_layers_calibration_phi-2_mmlu_all_truthfulqa_intervention}
\end{figure*}

While the confidence correction phase is observed across various models (e.g., Phi-2, LLaMA-3-8B), the calibration direction identified in Phi-2 does not generalize effectively to Mistral or LLaMA-2. 
A more detailed investigation into whether this lack of generalization stems from architecture, training procedure, or representational differences is needed. Moreover, it would be good to explore whether the calibration direction can be made dataset-agnostic by identifying consistent patterns across knowledge-focused datasets like TruthfulQA. Some of our initial findings (see Figure \ref{fig:all_layers_calibration_phi-2_mmlu_all_truthfulqa_intervention}) point towards this direction, more investigations on similar lines would be helpful in formalizing the calibration direction for different models.

\subsection*{Recent Developments and Positioning}
Several recent works provide alternative strategies for calibration. These include post-hoc methods based solely on generated outputs \citep{ulmer-etal-2024-calibrating}, reward-based adjustments in RLHF \citep{leng2025taming}, and sample-consistency-based calibrations \citep{Lyu_Shridhar_Malaviya_Zhang_Elazar_Tandon_Apidianaki_Sachan_Callison-Burch_2025}. In contrast, our approach contributes a mechanistic perspective, highlighting internal residual stream dynamics and structured directions that actively regulate model confidence during forward computation. This complements post-hoc techniques by providing more grounded explanations for how and when calibration emerges inside large-scale models.

In summary, our findings open several compelling directions for advancing both the interpretability and reliability of LLMs, especially in applications where calibrated uncertainty estimates are crucial. Exploring these avenues can help move toward principled architectures and training objectives that foster better-calibrated models by design.

\begin{figure*}[t]
\centering
\captionsetup[subfigure]{labelformat=parens}

\newcommand{\scalefactor}{0.45} 

\begin{subfigure}[b]{\scalefactor\linewidth}
    \includegraphics[width=\linewidth]{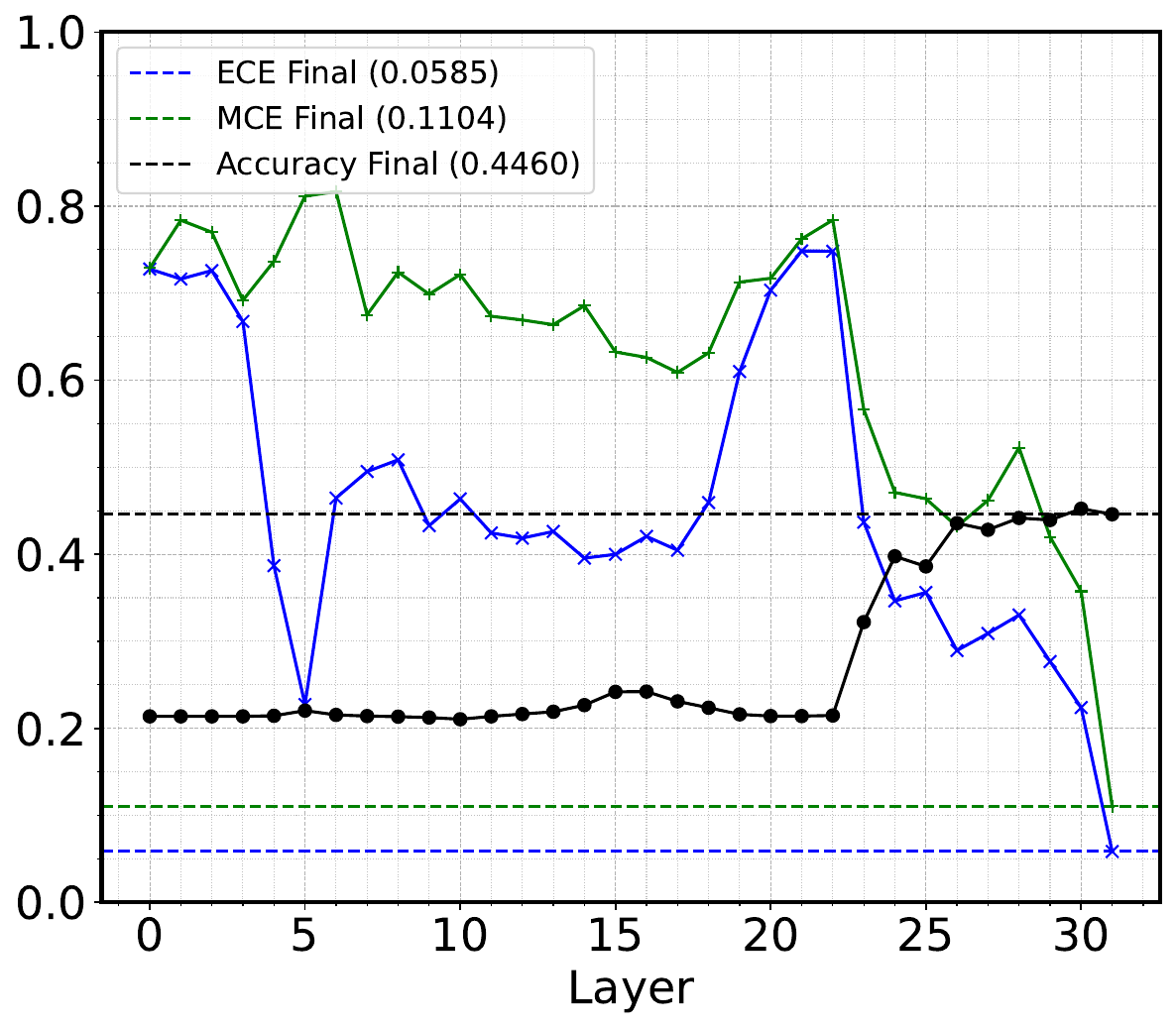}
    \caption{MMLU STEM}
\end{subfigure}\hfill
\begin{subfigure}[b]{\scalefactor\linewidth}
    \includegraphics[width=\linewidth]{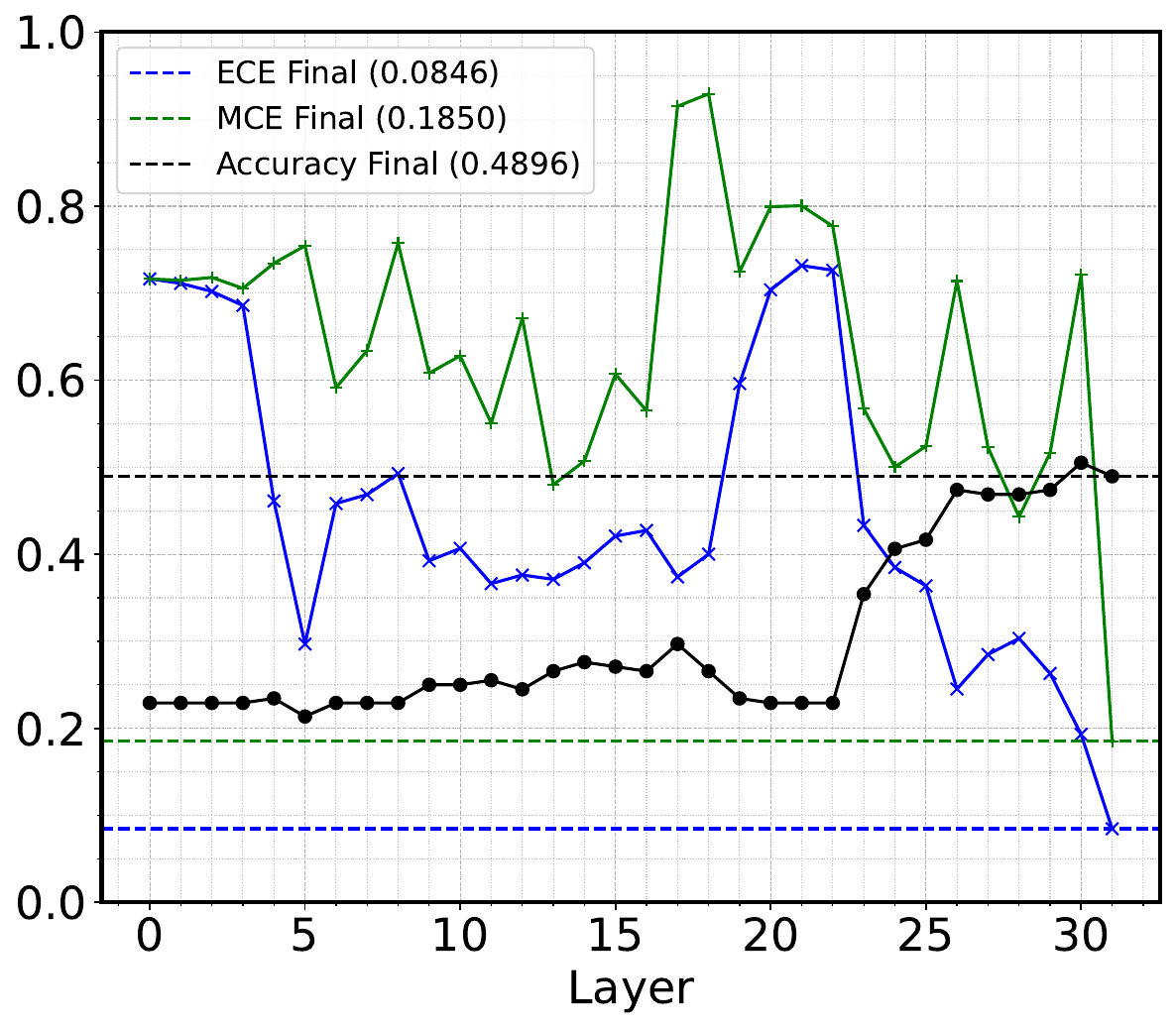}
    \caption{MMLU Humanities}
\end{subfigure}

\begin{subfigure}[b]{\scalefactor\linewidth}
    \includegraphics[width=\linewidth]{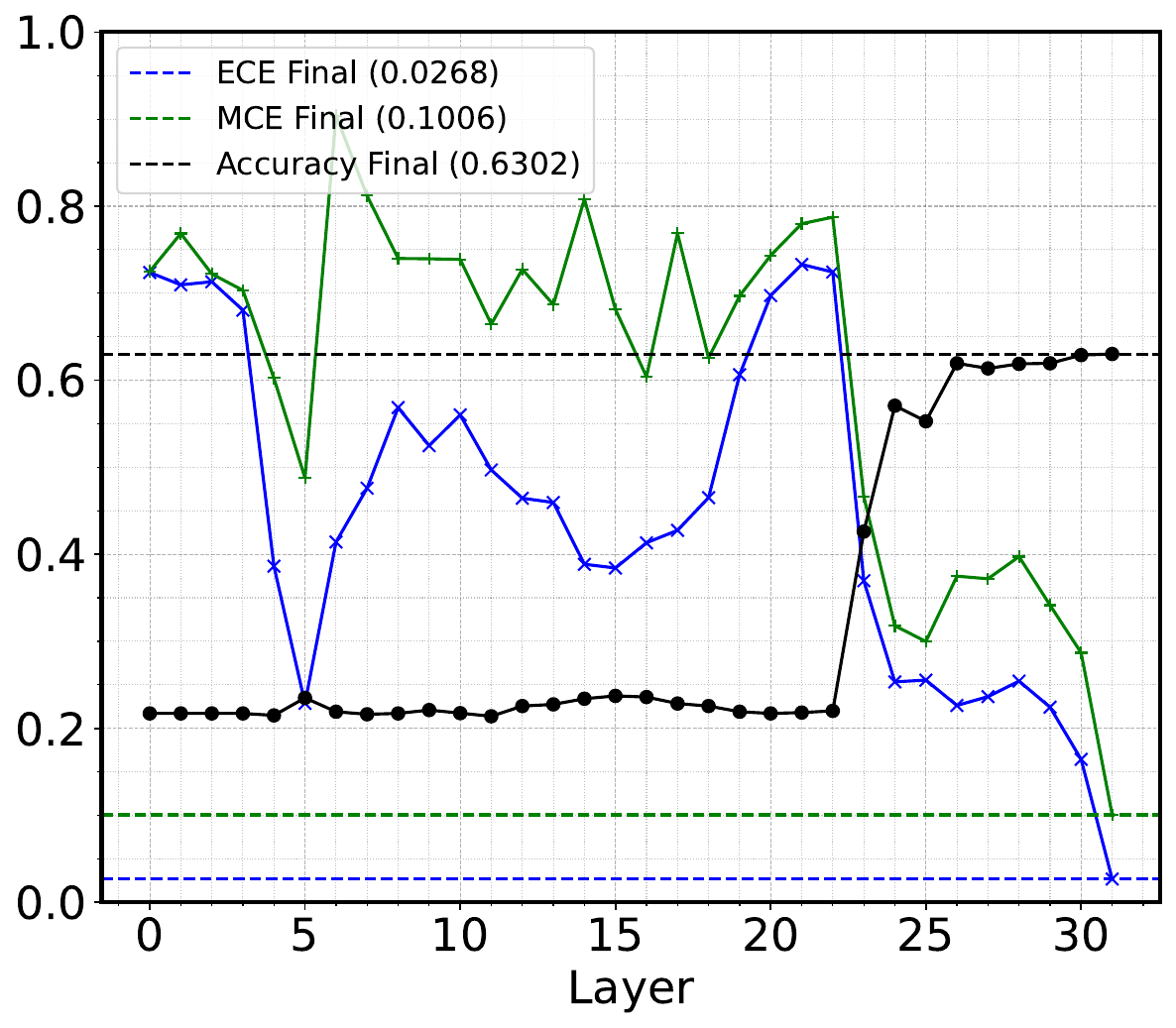}
    \caption{MMLU Social Science}
\end{subfigure}
\hfill
\begin{subfigure}[b]{\scalefactor\linewidth}
    \includegraphics[width=\linewidth]{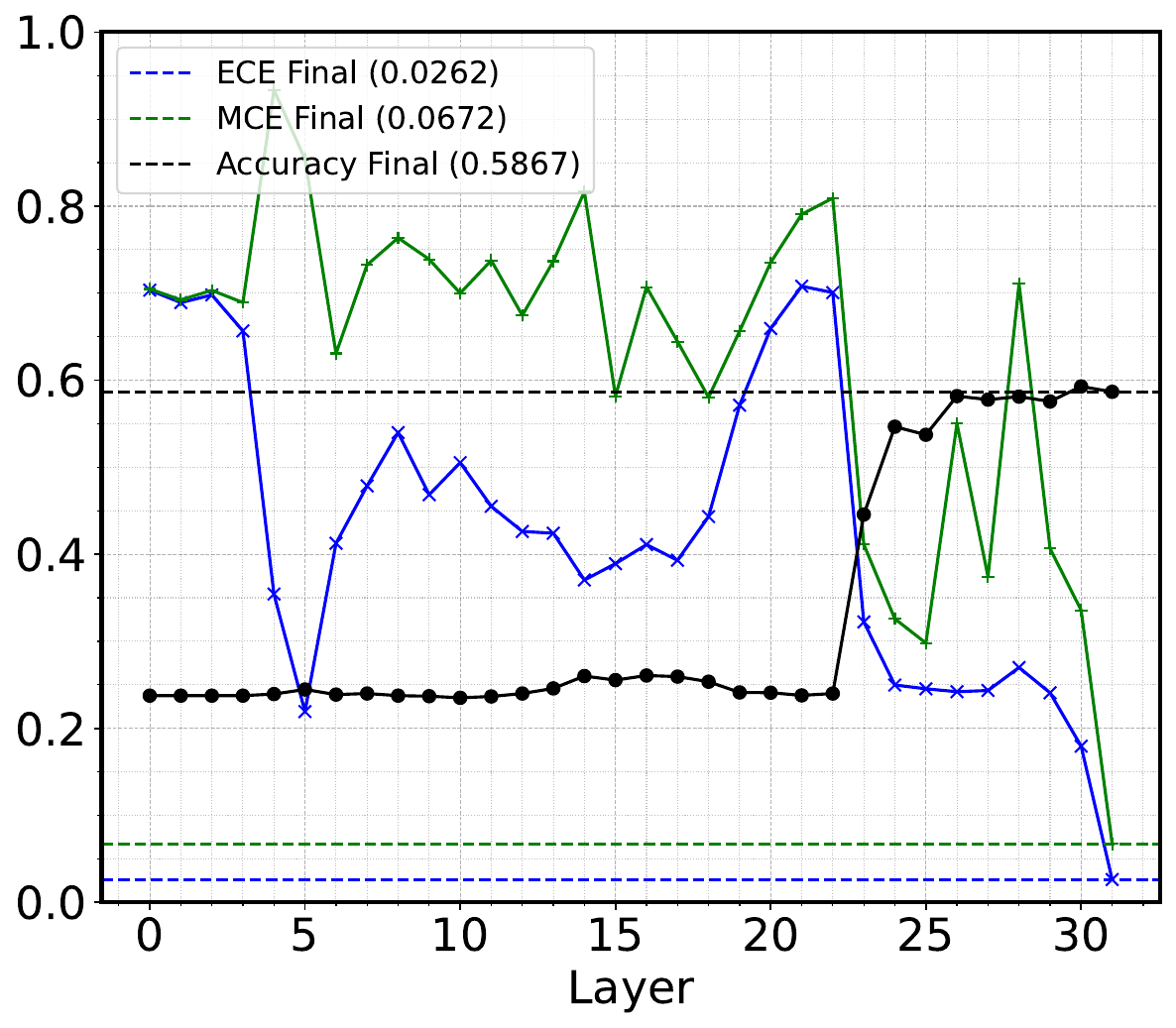}
    \caption{MMLU Others}
\end{subfigure}

\caption[Layer-wise Calibration of Phi-2 on MMLU splits]{The figure shows performance (Accuracy) along with model calibration scores (ECE and MCE) of the Phi-2 model on the different datasets. We observe that the model performance starts to rise from layer 22 and saturates at layer 25/26, with minor changes in the 26-31 layers. However, the ECE and MCE scores first rise (layers 26-28) and then decline (layers 29-31), highlighting the model calibration changing in the later layers, with meager changes in the model performance. This denotes that the residual stream in the later layers is affected/modified in such a way that modulates the model calibration with no/minor change in the model performance (black line). The upper/later layers showing the presence of \textbf{\textit{calibration correction phase}}. Similar trends are found for other models (Llama-3-8B Figure \ref{fig:all_layers_calibration_Llama-3-8B_mmlu_all}, Mistral-7B Figure \ref{fig:all_layers_calibration_mistralai_Mistral-7B-v0.1}, and Llama-2-7B Figure \ref{fig:all_layers_calibration_meta-llama_Llama-2-7b-hf_mmlu_all}).
}
\label{fig:all_layers_calibration_phi-2_mmlu_al}
\end{figure*}

\begin{figure*}[t]
\centering
 \includegraphics[width=0.62\linewidth]{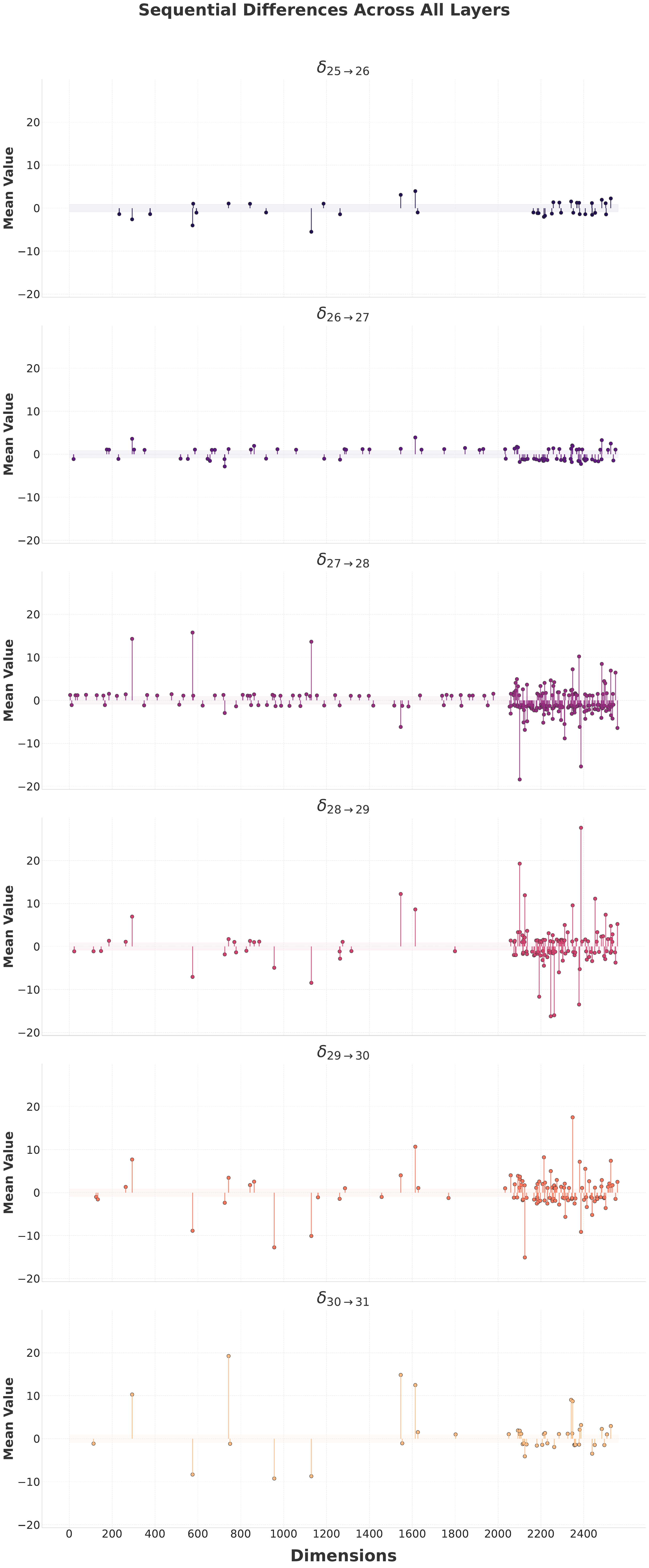}
 \caption[Consecutive Difference in Residual Stream for Phi-2]{{The figure shows the difference in residual stream, showing higher changes in the later dimensions, which further help compute the calibration direction $\mathbf{\hat{c}}$. We use the last three layers of the phi-2 model to compute the calibration direction (as described in the main paper), which shows generalization across multiple datasets (see Figure \ref{fig:all_layers_calibration_phi-2_mmlu_all_truthfulqa_intervention} where the direction computed using MMLU humanities generalizes for other datasets.)
  }}
  \label{fig:phi-2_layer_difference}
\end{figure*}

\end{document}